\pdfoutput=1

\documentclass[11pt]{article}

\usepackage[]{EMNLP2022}

\usepackage{times}
\usepackage{latexsym}

\usepackage[T1]{fontenc}

\usepackage[utf8]{inputenc}

\usepackage{microtype}

\usepackage{caption}

\usepackage{graphicx}
\usepackage{enumitem}

\newcommand{\thickhline}{\noalign{\hrule height 2pt}}

\usepackage{xparse}
\NewDocumentCommand{\heng}
{ mO{} }{\textcolor{red}{\textsuperscript{\textit{Heng}}\textsf{\textbf{\small[#1]}}}}

\NewDocumentCommand{\carl}
{ mO{} }{\textcolor{blue}{\textsuperscript{\textit{Carl}}\textsf{\textbf{\small[#1]}}}}

\NewDocumentCommand{\tuan}
{ mO{} }{\textcolor{orange}{\textsuperscript{\textit{Tuan}}\textsf{\textbf{\small[#1]}}}}

\NewDocumentCommand{\garrett}
{ mO{} }{\textcolor{orange}{\textsuperscript{\textit{Garrett}}\textsf{\textbf{\small[#1]}}}}

\NewDocumentCommand{\kevin}
{ mO{} }{\textcolor{purple}{\textsuperscript{\textit{Kevin}}\textsf{\textbf{\small[#1]}}}}

\title{Translation between Molecules and Natural Language}%

\renewcommand\footnotemark{}
\author{Carl Edwards\textsuperscript{1}*\thanks{* indicates equal contributions.}, Tuan Lai\textsuperscript{1,2}*, Kevin Ros\textsuperscript{1}, Garrett Honke\textsuperscript{2}, Kyunghyun Cho\textsuperscript{3,4}, Heng Ji\textsuperscript{1} \\
     \textsuperscript{1}University of Illinois Urbana-Champaign \\
     \textsuperscript{2}X, the Moonshot Factory \\
     \textsuperscript{3}New York University, \textsuperscript{4} Genentech   \\
     \texttt{\{cne2, tuanml2, kjros2, hengji\}@illinois.edu} \\
     \texttt{ghonk@google.com}, \texttt{kyunghyun.cho@nyu.edu}}

\begin{document}
\maketitle

\begin{abstract}

We present \textbf{MolT5} -- a self-supervised learning framework for pretraining models on a vast amount of unlabeled natural language text and molecule strings. \textbf{MolT5} allows for new, useful, and challenging analogs of traditional vision-language tasks, such as molecule captioning and text-based de novo molecule generation (altogether: translation between molecules and language), which we explore for the first time. Since  \textbf{MolT5} pretrains models on single-modal data, it helps overcome the chemistry domain shortcoming of data scarcity. Furthermore, we consider several metrics, including a new cross-modal embedding-based metric, %
to evaluate the tasks of molecule captioning and text-based molecule generation.
Our results show that \textbf{MolT5}-based models are able to generate outputs, both molecules and captions, which in many cases are high quality\footnote{All resources are publicly available at \href{https://github.com/blender-nlp/MolT5}{github.com/blender-nlp/MolT5}}.
\end{abstract}

\section{Introduction}

Imagine a future where a doctor can write a few sentences describing a specialized drug for treating a patient and then receive the exact structure of the desired drug. Although this seems like science fiction now, with progress in integrating natural language and molecules, it might well be possible in the future.
Historically, drug creation has commonly been done by humans who design and build individual molecules. In fact, bringing a new drug to market can cost over a billion dollars and take over ten years \cite{gaudelet2021utilizing}. Recently, there has been considerable interest in using new deep learning tools to facilitate in silico drug design-- a field often called cheminformatics \cite{10.1093/bib/bby061}. Yet, many of these experiments still focus on molecules and their low-level properties such as logP (the octanol-water partition coefficient) \cite{bagal2021molgpt}. In the future, we foresee a need for a higher-level control over molecule design, which can easily be facilitated by natural language. 

\begin{figure}
\centering
\includegraphics[width=\columnwidth]{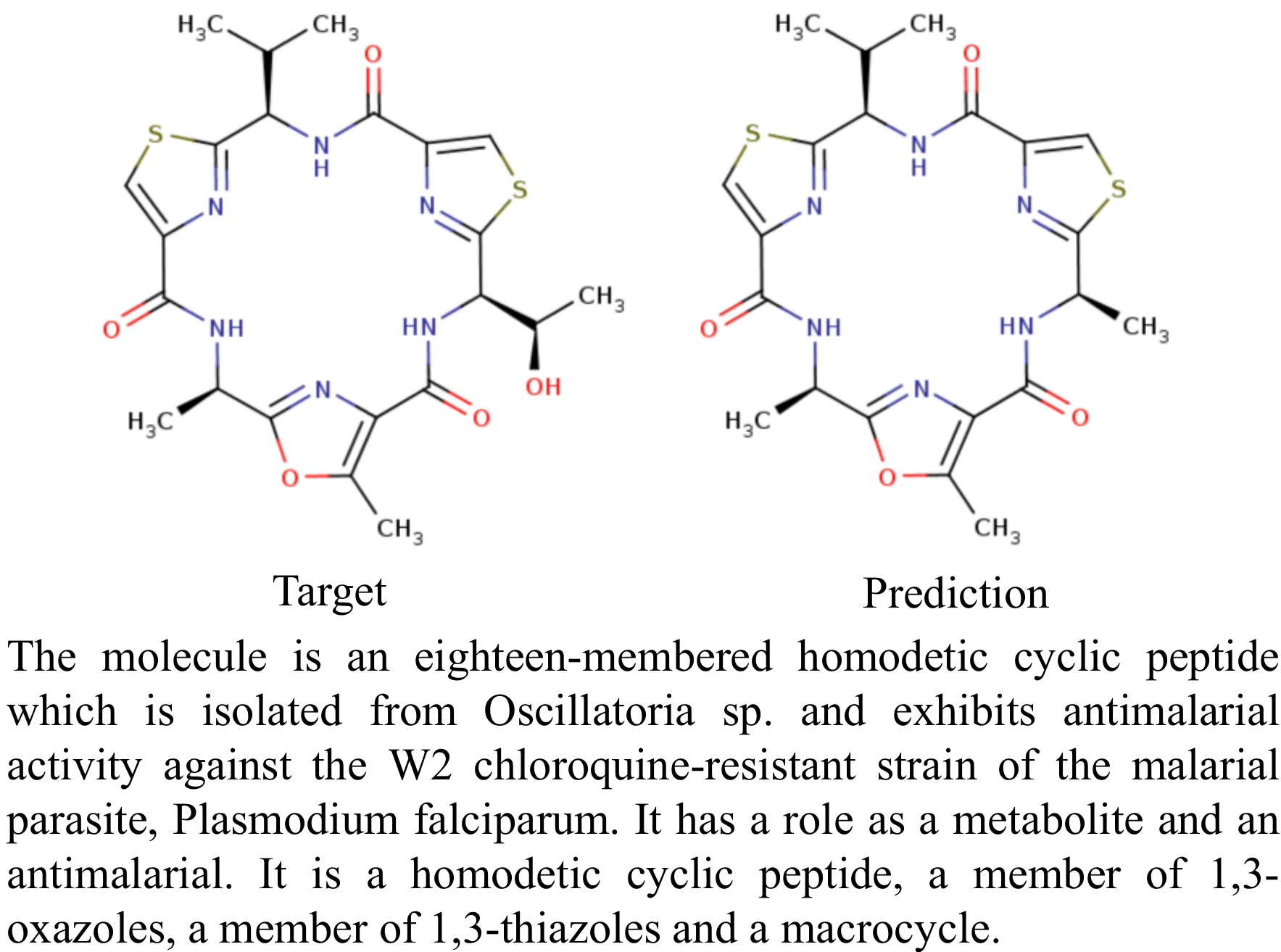}
\caption{An example output from our model for the molecule generation task. The left is the ground truth, and the right is a molecule generated from the given natural language caption. %
}
\label{fig:gen_task}
\end{figure}

\begin{figure*}[!ht]
\centering
\includegraphics[width=0.95\textwidth]{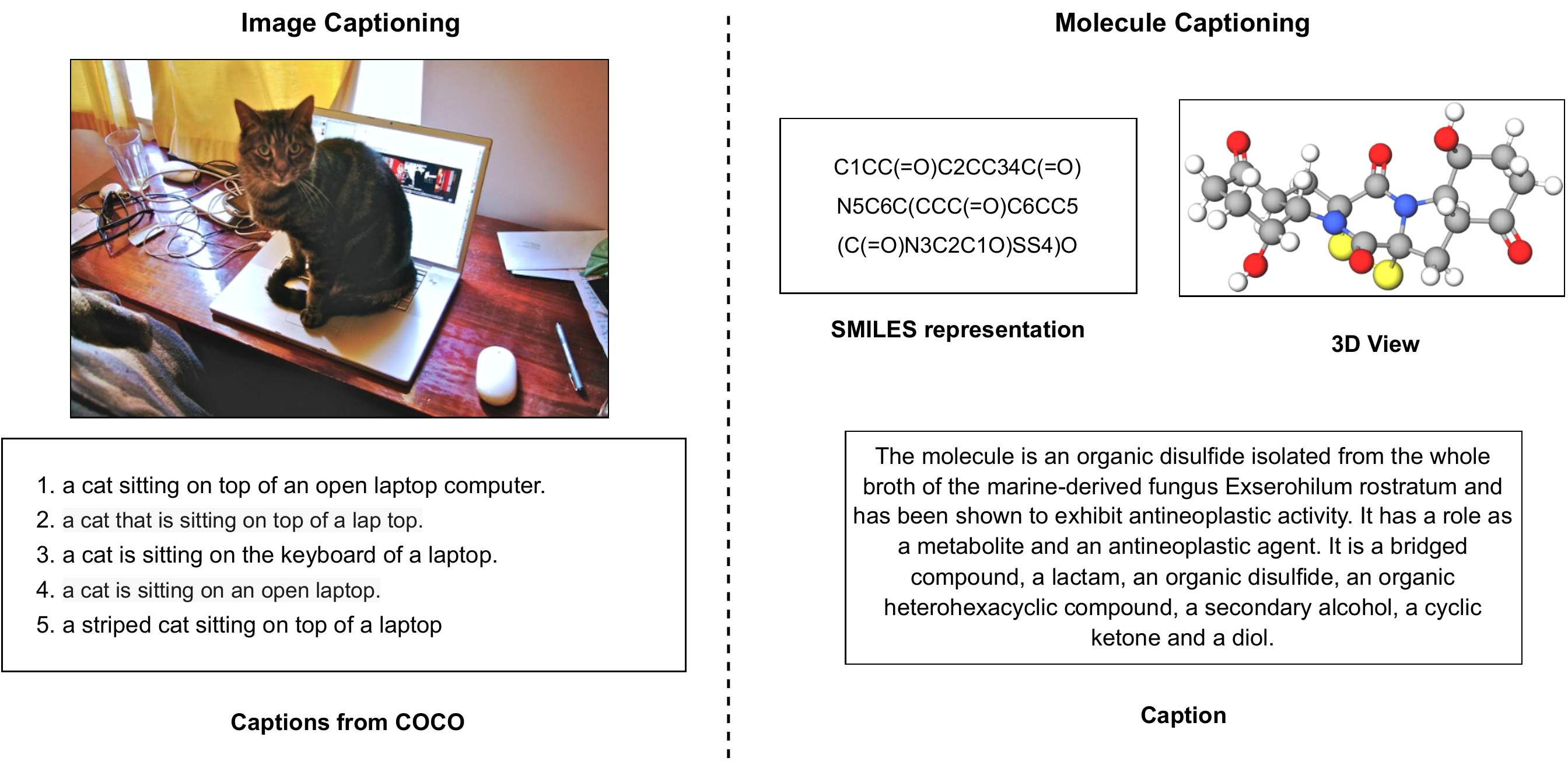}
\caption{An example of both the image captioning task \cite{Chen2015MicrosoftCC} and molecule captioning. Molecule captioning is considerably more difficult because of the increased linguistic variety in possible captions.} %
\label{fig:task}
\end{figure*}

In this work, we pursue an ambitious goal of translating between molecules and language by proposing two new tasks: molecule captioning and text-guided de novo molecule generation. In molecule captioning, we take a molecule (e.g., as a SMILES string) and generate a caption that describes it (Figure \ref{fig:task}). In text-guided molecule generation, the task is to create a molecule that matches a given natural language description (Figure \ref{fig:gen_task}). These new tasks would help to accelerate research in multiple scientific domains by enabling chemistry domain experts to generate new molecules and better understand them using natural language.

While our proposed molecule-language tasks share some similarities with vision-language tasks, they have several inherent difficulties that separate them from existing vision-language analogs: 1) creating annotations for molecules requires significant domain expertise, 2) thus, it is significantly more difficult to acquire large numbers of molecule-description pairs, 3) the same molecule can have many functions and thus be described in very different ways, 
which causes 4) existing evaluation measures based on reference descriptions, such as BLEU, to fail to adequately evaluate these tasks.

To address the issue of data scarcity (i.e., difficulties 1 and 2), we propose  a new self-supervised learning framework named MolT5 (\textbf{\underline{Mol}}ecular \textbf{\underline{T5}}) that is inspired by the recent progress in pretraining multilingual models \cite{devlin2019bert,Liu2020MultilingualDP}. MolT5 first pretrains a model on a vast amount of unlabeled natural language text and molecule strings using a simple denoising objective. After that, the pretrained model is finetuned on limited gold standard annotations. Furthermore, to adequately evaluate models for molecule captioning or generation, we consider various kinds of metrics and also adopt a new metric based on Text2Mol \cite{edwards2021text2mol}. We repurpose this retrieval model for assessing the similarity between the ground truth molecule/description and the generated description/molecule, respectively.

To the best of our knowledge, there is no work yet on molecule captioning or text-guided molecule generation. The closest existing work to molecule captioning falls within the scope of image captioning  \cite{Vinyals2015ShowAT}. However, molecule captioning is arguably much more challenging due to the increased linguistic variety in possible captions (Figure~\ref{fig:task}). A molecule could be described with an IUPAC name, with one of many different synthetic routes from known precursor molecules, in terms of the properties (e.g. carcinogenic or lipophilic), with the applications of the molecule (e.g. a dye, an antipneumonic, or an antifungal), or in terms of its functional groups (e.g. ``substituted by hydroxy groups at positions 5 and 7 and a methyl group at position 8''), among other methods.

In summary, our main contributions are:
\begin{enumerate}[nolistsep]
    \item We propose two new tasks: 1) molecule captioning, where a description is generated for a given molecule, and 2) text-based de novo molecule generation, where a molecule is generated to match a given text description.%
    \item We consider multiple evaluation metrics for these new tasks, and we adopt a new cross-modal retrieval similarity metric based on Text2Mol \cite{edwards2021text2mol}.
    \item We propose \textbf{MolT5}: a self-supervised learning framework for jointly training a model on molecule string representations and natural language text, which can then be finetuned on a cross-modal task.
\end{enumerate}

\section{Tasks}

With the ambitious goal of bi-directional translation between molecules and language, we propose two new novel tasks: molecule captioning (Section \ref{sec:task_molecule_captioning}) and text-based molecule generation (Section \ref{sec:task_molecule_generation}). 

\subsection{Molecule Captioning} \label{sec:task_molecule_captioning}

For any given molecule, the goal of molecule captioning is to describe the molecule and what it does. An example is shown in Figure \ref{fig:task}.
Molecules are often represented as SMILES strings \cite{weininger1988smiles, weininger1989smiles}, a linearization of the molecular graph which can be interpreted as a language for molecules. Thus, this task can be considered an exotic translation task, and sequence to sequence models serve as excellent baselines. %

\subsection{Text-Based de Novo Molecule Generation}\label{sec:task_molecule_generation}

The goal of the de novo molecule generation task is to train a model which can generate a variety of possible new molecules. %
Existing work tends to focus on evaluating the model coverage of the chemical space~\cite{polykovskiy2020molecular}. Instead, we propose generating molecules based on a natural language description of the desired molecule--this is essentially swapping the input and output for the captioning task. An example of this task is shown in Figure \ref{fig:gen_task}. Recent work, such as DALL$\cdot$E \cite{ramesh2021zero,ramesh2022hierarchical}, which generates images from text, has shown the ability to seamlessly integrate multiple properties, such as chairs and avocados, in an image. This points towards similar applications in the molecule generation domain via the usage of natural language.

\section{Evaluation Metrics}

\subsection{Text2Mol Metric}

Since we are considering new cross-modal tasks between molecules and text, we also introduce a new cross-modal evaluation metric. This is based on Text2Mol \cite{edwards2021text2mol}, which aims to train a retrieval model to rank molecules given their text descriptions. Since the ranking function uses cosine similarity between embeddings, a trained model can be repurposed for evaluating the similarity between the ground truth molecule/description and the generated description/molecule (respectively). To this end, we first train a base multi-layer perceptron (MLP) model from Text2Mol. This model is then used to generate similarities of the candidate molecule-description pairs, which can be compared to the average similarity of the ground truth molecule-description pairs. We also note that negative molecule-description pairs have an average similarity of roughly zero.

\subsection{Evaluating Molecule Captioning}

Traditionally, captioning tasks have been evaluated by natural language generation metrics such as BLEU \cite{papineni2002bleu}, ROUGE \cite{lin2004rouge}, and METEOR \cite{banerjee2005meteor}. %
Unlike captioning tasks such as COCO \cite{Chen2015MicrosoftCC}, which has several captions per image, in our task we only have one reference caption. This makes these metrics less effective, especially because there are many non-overlapping ways to describe a molecule. Nevertheless, for comparison, we still report these scores (e.g., aggregated sentence-level METEOR scores).

\subsection{Evaluating Text-Based de Novo Molecule Generation}

Considerable interest has grown in applying deep generative models to de novo molecule generation. Because of this, a number of metrics have been proposed, such as novelty and scaffold similarity \cite{polykovskiy2020molecular}. However, many of these metrics do not apply to our problem-- we want our generated molecule to match the input text instead of being generally diverse. Instead, we consider metrics which measure the distance of the generated molecule to either the ground truth molecule or the ground truth description, such as our proposed Text2Mol-based metric. 

We employ three fingerprint metrics: MACCS FTS, RDK FTS, and Morgan FTS, %
where FTS stands for fingerprint Tanimoto similarity \cite{tanimoto1958elementary}. MACCS \cite{durant2002reoptimization}, RDK \cite{schneider2015get}, and Morgan \cite{rogers2010extended} are each fingerprinting methods for molecules. The fingerprints of two molecules are compared using Tanimoto similarity (also known as Jaccard index), and the average similarity over the evaluation dataset is reported. See \cite{campos2021img2smi} for more details. We also report exact SMILES %
string matches, Levenshtein distance \cite{miller2009levenshtein}, and SMILES BLEU scores. %

\citet{preuer2018frechet} propose Fréchet ChemNet Distance (FCD), which is inspired by the Fréchet Inception Distance (FID) \cite{heusel2017gans}. FCD is based on the penultimate layer of a network called ``ChemNet'', which was trained to predict the activity of drug molecules. Thus, FCD takes into account chemical and biological information about molecules in order to compare them. This allows molecules to be compared based on the latent information required to predict useful properties rather than a string-based metric.

\begin{figure*}[ht]
\centering
\includegraphics[width=\textwidth]{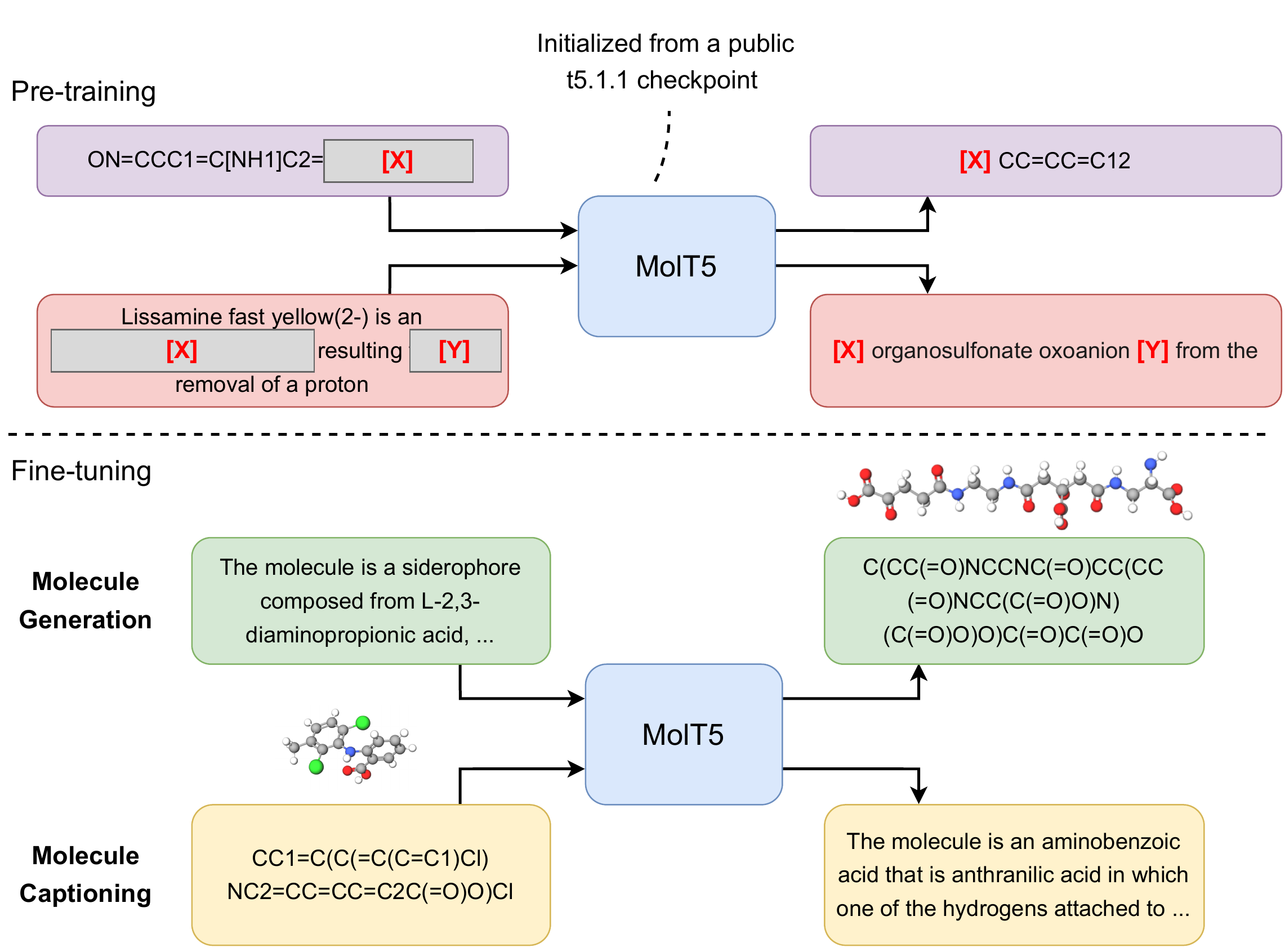}
\caption{A diagram of our framework. We first pre-train MolT5 on a large amount of data of both SMILES string and natural language using the ``replace corrupted spans'' objective \cite{raffel2020exploring}. After the pre-training stage, MolT5 can be easily fine-tuned for either the task of molecule captioning or generation (or both).}
\label{fig:t5_task}
\end{figure*}

In the case of models which use SMILES strings, generated molecules can be syntactically invalid. Therefore, we also report validity as the percent of molecules which can be processed by RDKIT \cite{Landrum2021RDKit2021_03_2} as in \cite{polykovskiy2020molecular}.

\section{MolT5 -- Multimodal Text-Molecule Representation Model} \label{sec:molt5}

We can crawl a massive amount of text from the Internet. For example, \newcite{raffel2020exploring} built a Common Crawl-based dataset that contains over 700 GB of reasonably clean and natural English text. On the other hand, over a billion molecules are also available from public databases such as ZINC-15 \cite{Sterling2015ZINC1}. Inspired by the progress in large-scale pretraining \cite{ramesh2021zero}, we propose a new self-supervised learning framework named \textbf{MolT5} (\textbf{\underline{Mol}}ecular \textbf{\underline{T5}}) to leverage the vast amount of unlabeled natural language text and molecule strings.

Figure \ref{fig:t5_task} shows an overview of MolT5. We first initialize an encoder-decoder Transformer model \cite{vaswani2017attention} using one of the public checkpoints of T5.1.1\footnote{\url{https://tinyurl.com/t511-ckpts}}, an improved version of T5 \cite{raffel2020exploring}. After that, we pretrain the model using the ``replace corrupted spans'' objective \cite{raffel2020exploring}. More specifically, during each pretraining step, we sample a minibatch comprising both natural language sequences and SMILES sequences. For each sequence, some words in the sequence are randomly chosen for corruption. Each consecutive span of corrupted tokens is replaced by a sentinel token (shown as [X] and [Y] in Figure \ref{fig:t5_task}). Then the task is to predict the dropped-out spans.\footnote{For more explanation of the pretraining task, we refer the readers to the original T5 paper \cite{raffel2020exploring}.}

Molecules (e.g. represented as SMILES strings) can be thought of as a language with a very unique grammar. Then, intuitively, our pretraining stage essentially trains a single language model on two monolingual corpora from two different languages, and there is no explicit alignment between the two corpora. This approach is similar to how some multilingual language models such as mBERT \cite{devlin2019bert} and mBART \cite{Liu2020MultilingualDP} were pretrained. As models such as mBERT demonstrate excellent cross-lingual capabilities \cite{piresetal2019multilingual}, we also expect models pretrained using MolT5 to be useful for text-molecule translation tasks.

After the pretraining process, we can finetune the pretrained model for either molecule captioning or generation (depicted by the bottom half of Figure \ref{fig:t5_task}). In molecule generation, the input is a description, and the output is the SMILES representation of the target molecule. On the other hand, in molecule captioning, the input is the SMILES string of some molecule, and the output is a caption describing the input molecule.

\section{Experiments and Results}

\subsection{Data}

\paragraph{Pretraining Data} As described in Section \ref{sec:molt5}, the pretraining stage of MolT5 requires two monolingual corpora: one consisting of natural language text and the other consisting of molecule representations. We use the ``Colossal Clean Crawled Corpus'' (C4) \cite{raffel2020exploring} as the pretraining dataset for the textual modality. For the molecular modality, we directly utilize the 100 million SMILES strings used in Chemformer \cite{irwindimitriadishebjerrum2021}. As these strings were selected from the ZINC-15 dataset \cite{sterling2015zinc}, we refer to this pretraining dataset as ZINC from this point.

\paragraph{Finetuning and Evaluation Data} We use ChEBI-20 \cite{edwards2021text2mol} as our gold standard dataset for finetuning and evaluation. It consists of 33,010 molecule-description pairs, which are separated into 80/10/10\% train/validation/test splits.
We use ChEBI-20 to finetune MolT5-based models and to train baseline models. Many captions in ChEBI-20 contain a name for the molecule at the start of the string (e.g., ``Rostratin D is an organic disulfide isolated from ...''). To force the models to focus on the semantics of the description, we replace the molecule's name with "The molecule is [...]" (e.g., ``The molecule is an organic disulfide isolated from ...''). 

\subsection{Baselines}\label{sec:baselines}
Any sequence-to-sequence model is applicable to our new tasks (i.e., molecule captioning and generation). We implement the following baselines:
\begin{enumerate}
    \item \textbf{RNN-GRU} \cite{cho2014learning}. We implement a 4-layer GRU recurrent neural network. The encoder is bidirectional.
    \item \textbf{Transformer} \cite{vaswani2017attention}. We train a vanilla Transformer model consisting of six encoder and decoder layers. %
    \item \textbf{T5} \cite{raffel2020exploring}. We experiment with three public T5.1.1 checkpoints\footnote{\url{https://tinyurl.com/t511-ckpts}}: small, base, and large. We finetune each checkpoint for molecule captioning or molecule generation using the t5x framework \cite{roberts2022t5x}.%
\end{enumerate}

We train the baseline models on ChEBI-20 using SMILES representations for the molecules. Molecule captioning and generation are trained with molecules as input/output and text as output/input. More information about the baselines and the hyperparameters is in the appendix. %

\subsection{Pretraining Process}
We first initialize an encoder-decoder Transformer model using a public checkpoint of T5.1.1 (either \textit{t5.1.1.small}, \textit{t5.1.1.base}, or \textit{t5.1.1.large}). We then pretrain the model on the combined dataset of C4 and ZINC (i.e., C4+ZINC) for 1 million steps. Each step uses a batch size of 256 evenly split between text and molecule sequences. After this, we finetune the pretrained model on ChEBI-20 for either molecule captioning or generation. The number of finetuning steps is 50,000.

\begin{table*}[ht!]
\resizebox{\textwidth}{!}{
\centering
\tiny
\begin{tabular}{ c|c|c|c|c|c|c|c }

\multicolumn{1}{c}{\textbf{Model}} & \multicolumn{1}{c}{BLEU-2} & \multicolumn{1}{c}{BLEU-4} & \multicolumn{1}{c}{ROUGE-1} & \multicolumn{1}{c}{ROUGE-2} & \multicolumn{1}{c}{ROUGE-L} & \multicolumn{1}{c}{METEOR} & \multicolumn{1}{c}{Text2Mol} \\
\thickhline
Ground Truth& &  &  &  &  &  & 0.609 \\\hline
RNN & 0.251 & 0.176 & 0.450 & 0.278 & 0.394 & 0.363 & 0.426 \\
Transformer & 0.061 & 0.027 & 0.204 & 0.087 & 0.186 & 0.114 & 0.057 \\\hline
T5-Small & 0.501 & 0.415 & 0.602 & 0.446 & 0.545 & 0.532 & 0.526 \\
MolT5-Small & 0.519 & 0.436 & 0.620 & 0.469 & 0.563 & 0.551 & 0.540 \\\hline
T5-Base & 0.511 & 0.423 & 0.607 & 0.451 & 0.550 & 0.539 & 0.523 \\
MolT5-Base & 0.540 & 0.457 & 0.634 & 0.485 & 0.578 & 0.569 & 0.547 \\\hline
T5-Large & 0.558 & 0.467 & 0.630 & 0.478 & 0.569 & 0.586 & 0.563 \\
MolT5-Large & \textbf{0.594} & \textbf{0.508} & \textbf{0.654} & \textbf{0.510} & \textbf{0.594} & \textbf{0.614} & \textbf{0.582} \\
\end{tabular}
}
\caption{Molecule captioning results on the test split of CheBI-20. Rouge scores are F1 values.}
\label{tab:results_captioning}
\end{table*}

\begin{figure*}[t]
\centering
\includegraphics[width=\textwidth]{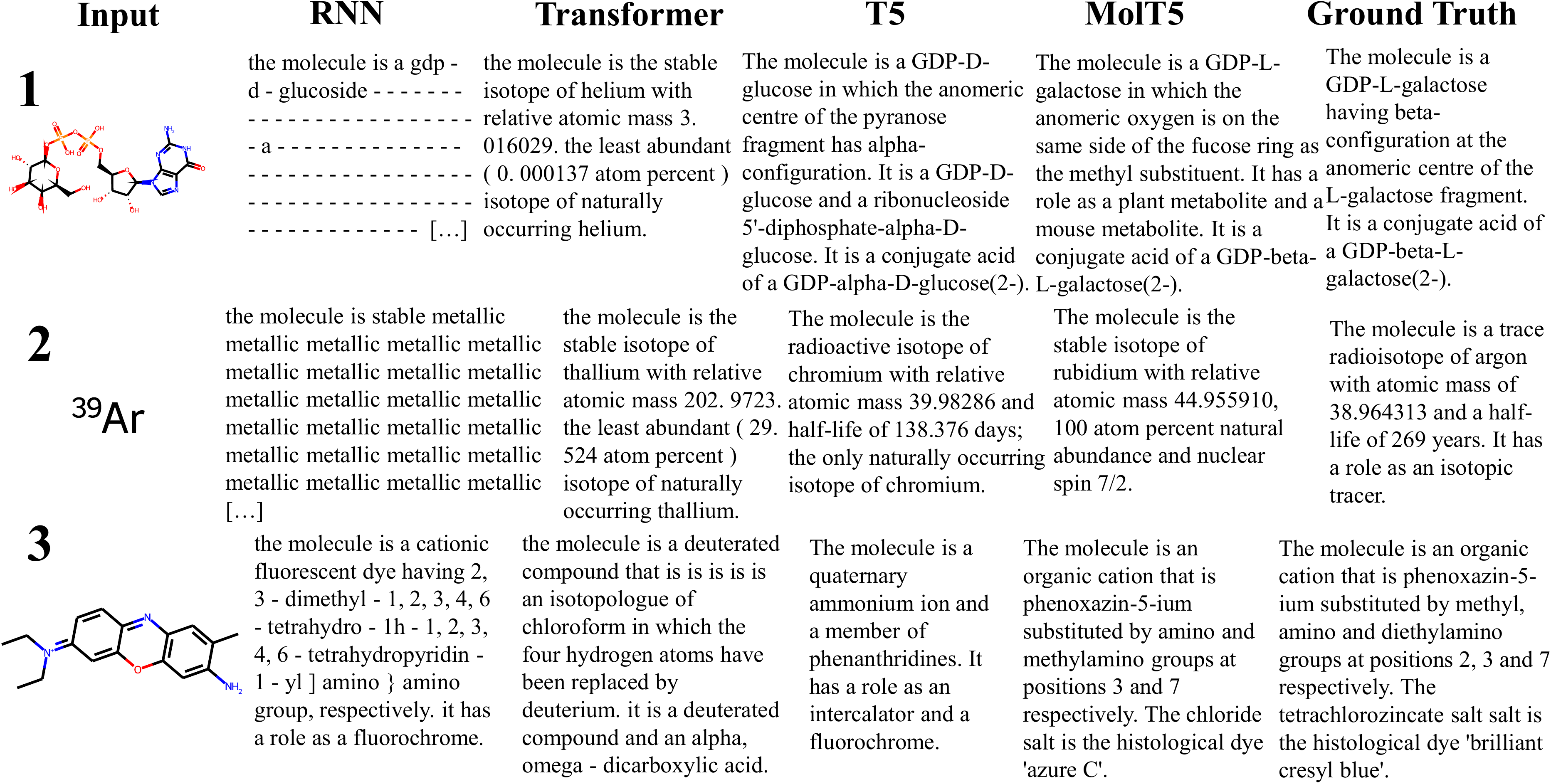}
\caption{Example captions generated by different models.}
\label{fig:qual_caption}
\end{figure*}

\subsection{Molecule Captioning}

Table \ref{tab:results_captioning} shows the overall molecule captioning results. The pretrained models, either T5 or MolT5, are considerably better at generating realistic language to describe a molecule than the RNN and Transformer baselines. The RNN is more capable of extracting relevant properties from molecules than the Transformer, but it generally produces ungrammatical outputs. On the other hand, the Transformer produces grammatical outputs, but they tend to repeat the same properties, such as carcinogenic, regardless of whether they apply. For this reason, the Text2Mol scores are much lower for the Transformer model, since its outputs match the given molecule much less frequently. We speculate that the ChEBI-20 dataset is too small to effectively train a Transformer without large-scale pretraining. We find that our additional pretraining of MolT5 results in a reasonable increase over T5 in captioning performance on both the traditional NLG metrics and our Text2Mol metric for each model size. Finally, we refer the reader to Section \ref{appendix:stats_sign} in the appendix for information about the statistical significance of our results.

Several examples of different models' outputs are shown in Figure \ref{fig:qual_caption} and Appendix Figure \ref{fig:qual1_cap_appendix}. In (1), MolT5's description matches best, identifying the molecule as a ``GDP-L-galactose''. MolT5 is usually able to recognize what general class of molecule it is looking at (e.g. cyclohexanone, maleate salt, etc.). In general, all models often look for the closest compound they know and base their caption on that. The argon atom, example (2) with SMILES `[39Ar]', is not present in the training dataset bonded to any other atoms (likely because it is an inert noble gas). All models recognize that (2) is a single atom, but they are unable to describe it.
In (3), the models try to caption a histological dye. MolT5 captions the molecule as an azure histological dye, which is very close to the ground truth ``brilliant cresyl blue'', while T5 does not.

\begin{table*}[ht!]
\resizebox{\textwidth}{!}{
\centering
\begin{tabular}{ c|c|c|c|c|c|c|c|c|c }

\multicolumn{1}{c}{\textbf{Model}} & \multicolumn{1}{c}{BLEU$\uparrow$} & \multicolumn{1}{c}{Exact$\uparrow$} & \multicolumn{1}{c}{Levenshtein$\downarrow$} & \multicolumn{1}{c}{MACCS FTS$\uparrow$} & \multicolumn{1}{c}{RDK FTS$\uparrow$} & \multicolumn{1}{c}{Morgan FTS$\uparrow$} & \multicolumn{1}{c}{FCD$\downarrow$} & \multicolumn{1}{c}{Text2Mol$\uparrow$} & \multicolumn{1}{c}{Validity$\uparrow$} \\
\thickhline
Ground Truth & 1.000 & 1.000 & 0.0 & 1.000 & 1.000 & 1.000 & 0.0 & 0.609 & 1.0 \\\hline
RNN & 0.652 & 0.005 & 38.09 & 0.591 & 0.400 & 0.362 & 4.55 & 0.409 & 0.542 \\
Transformer & 0.499 & 0.000 & 57.66 & 0.480 & 0.320 & 0.217 & 11.32 & 0.277 & \textbf{0.906} \\\hline
T5-Small & 0.741 & 0.064 & 27.703 & 0.704 & 0.578  & 0.525  & 2.89  & 0.479 & 0.608 \\
MolT5-Small & 0.755 & 0.079 & 25.988 & 0.703 & 0.568 & 0.517 & 2.49 & 0.482 & 0.721 \\\hline
T5-Base & 0.762 & 0.069 & 24.950 & 0.731 &  0.605 & 0.545 & 2.48 & 0.499 & 0.660  \\
MolT5-Base & 0.769 & 0.081 & 24.458 & 0.721 &  0.588 & 0.529 & 2.18 & 0.496 & 0.772 \\\hline
T5-Large & 0.854 & 0.279 & 16.721 & 0.823 &  0.731 & 0.670 & 1.22 & 0.552 & 0.902 \\
MolT5-Large & \textbf{0.854} & \textbf{0.311} & \textbf{16.071} & \textbf{0.834} & \textbf{0.746} & \textbf{0.684} & \textbf{1.20} & \textbf{0.554} & 0.905 \\
\end{tabular}
}
\caption{Molecule generation results on the test split of CheBI-20. Except for BLEU, Exact, Levenshtein, and Validity, other metrics are computed using only syntactically valid molecules, as in \cite{campos2021img2smi}.
}
\label{tab:results_generation}
\end{table*}

\begin{figure*}[t]
\centering
\includegraphics[width=\textwidth]{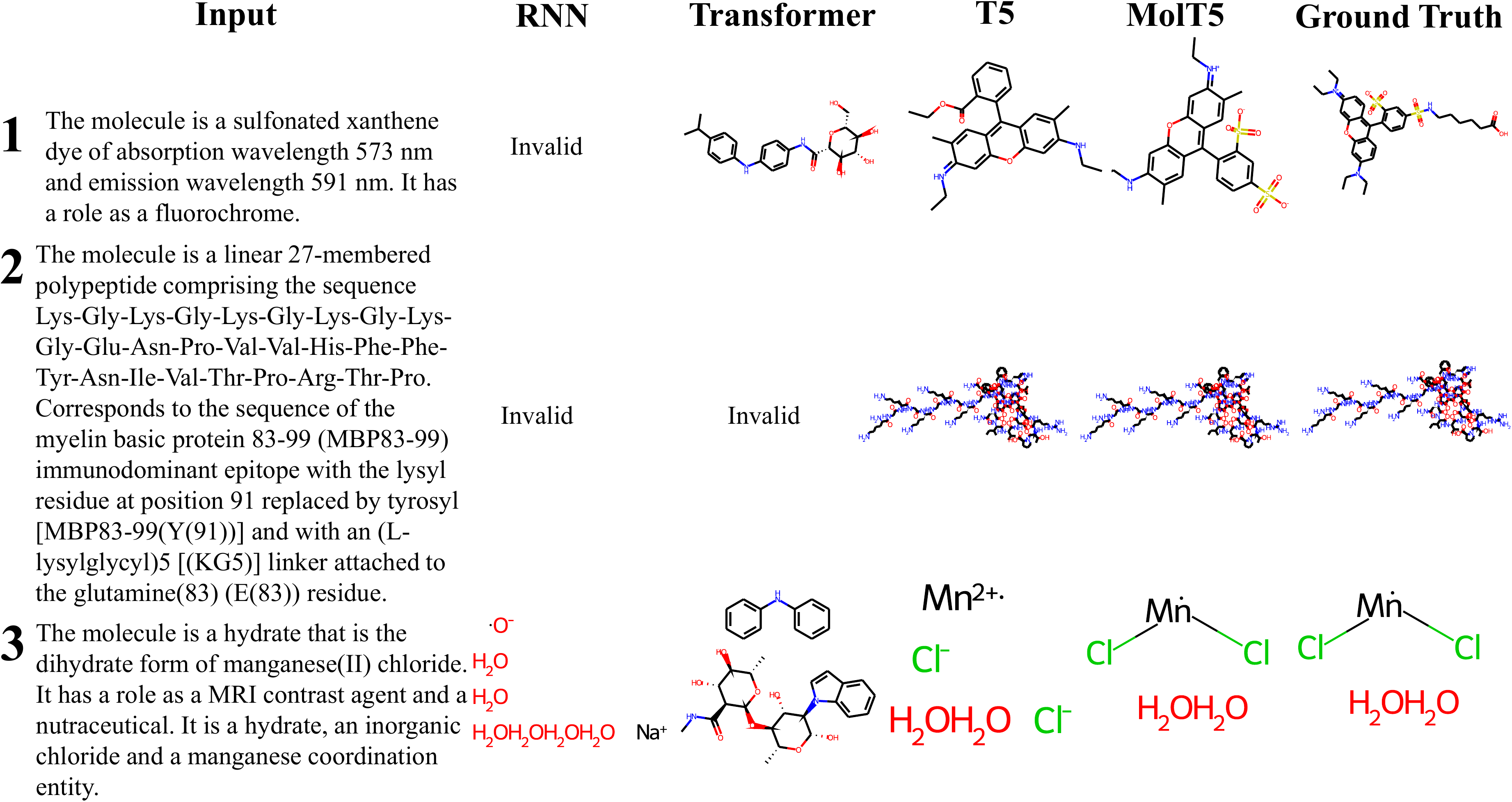}
\caption{Examples of molecules generated by different models.}
\label{fig:qual}
\end{figure*}

\subsection{Text-Based de novo Molecule Generation}

In the molecule generation task, the pretrained models also perform much better than the RNN and Transformer (Table \ref{tab:results_generation}). Although it is well known that scaling model size and pretraining data leads to significant performance increases \cite{kaplan2020scaling}, it was still surprising to see the results. For example, a default T5 model, which was only pretrained on text data, is capable of generating molecules which are much closer to the ground truth than the RNN and which are often valid. This trend also persists as language model size scales, since T5-large with 770M parameters outperforms the specifically pretrained MolT5-small with 60M parameters. Still, the pretraining in MolT5 slightly improves some molecule generation results, with especially large gains in validity. Finally, Section \ref{appendix:stats_sign} in the appendix has information about the statistical significance of our results.

We show results for the models in Figure \ref{fig:qual} and also in Figures \ref{fig:qual2}, \ref{fig:qual3}, and \ref{fig:qual4} in Appendix \ref{appendix:examples}, which we number by input description. Compared to T5, MolT5 is better able to understand instructions for manipulating molecules, as shown in examples (3, 4, 6, 7, 16, 18, 21). In many cases, MolT5 obtains exact matches with the ground truth (2, 3, 4, 6, 7, 8, 10, 12, 17, 20, 21). (3) is an interesting case, since it shows that MolT5 can understand crystalline solids like hydrates. (2) is another interesting example; it is the longest SMILES string, at 474 characters, which MolT5 is able to generate an exact match for. MolT5 understands peptides and can produce them from descriptions (2,15,17). It also shows this ability for saccharides (6, 21) and enzymes (8,20). MolT5 is able to understand rare atoms such as Ruthenium (5). However, in this case it still misses the atom's charge. Some example descriptions, such as (1), lack details so the molecules generated by MolT5 may be interesting to investigate.

\subsection{Probing the Model}

We conduct probing tests on the model for certain input properties, which are shown in Appendix \ref{appendix:probing}. Often, the model will generate molecules that it knows matches the input description from the finetuning data. It also creates solutions from these as well by adding various ions (e.g. ".[Na+]"). In some cases, it generates molecules not appearing in finetuning data (sometimes successfully sometimes not). For example, given the input ``The molecule is a corticosteroid.'', the first molecule generated is a well known corticosteroid called corticosterone. The fifth molecule generated is not present in the PubChem database. Based on a structure similarity search, it is most closely related to the androgenic steroid Fluoxymesterone and the corticosteroid Hydrocortisone.

\section{Related Work}

\subsection{Multimedia Representation}

Much recent work on multimedia representations falls into training large vision-language models \citep{su2019vl,lu2019vilbert,chen2020uniter}.
CLIP \citep{radford2021learning} trains a zero-shot image classifier by using natural language labels which can be easily extended.  A modification of CLIP's contrastive loss function, which follows
\cite{sohn2016improved}, is applied by Text2Mol \cite{edwards2021text2mol} for cross-modal retrieval between molecule and text pairs. %
\citet{edwards2021text2mol} also released the ChEBI-20 dataset of molecule-description pairs, which is used for training and evaluation in this paper. \citet{vallbioassayclr} leverage a contrastive loss between bioassay descriptions and molecules to predict activity between the two. \citet{sun2021fine} uses cross-modal attention with molecule structures to improve chemical entity typing. \citet{zeng2022deep} pretrain a language model to learn a joint representation between molecules and biomedical text via entity linking which they use for tasks such as relation extraction, molecule property prediction, and cross-modal retrieval like Text2Mol. Unlike our work, they do not explore generating text nor molecules. %
\newcite{vaucher2020automated} create a dataset of chemical equations and associated action sequences in natural language. \citet{vaucher2021inferring} then leverage this dataset to train a BART model which can plan chemical reaction steps. Their natural language generation is constrained to the specific reaction steps in their dataset-- the main purpose of their model is to create the steps for a reaction rather than describing molecules.

\subsection{Image Captioning and Text-Guided Image Generation}

Image captioning has been studied extensively \cite{pan2004automatic,Lu18imagecaption,hossain2019comprehensive, stefanini2021show}. Many recent studies tend to pretrain Transformer-based models on massive text-image corpora \cite{li2020oscar,Hu2022ScalingUV}.
Work has also been done in the biomedical domain \cite{pavlopoulos2019survey}, a close cousin of the chemistry domain, where tasks tend to be focused on diagnosis of various image types such as x-rays \cite{demner2016preparing}. %

The reverse problem, text-guided image generation, has proven considerably more challenging \cite{khan2021transformers}. Several attempts have used GAN-based methods \cite{reed2016generative,zhang2017stackgan,xu2018attngan}. Recent work has shown remarkable results. DALL$\cdot$ E \cite{ramesh2021zero,ramesh2022hierarchical} can seamlessly fuse multiple concepts together to generate a realistic image.

\subsection{Molecule Representation}
Molecule representation has been a long-standing problem in the field of cheminformatics. Traditionally, fingerprinting methods have been a preferred technique to featurize molecule structural representations \cite{rogers2010extended, cereto2015molecular}. 
These approaches do not allow representations to be learned from data.
In recent years, advances in machine learning and NLP have been applied to this problem. A popular input for these algorithms has been SMILES strings \cite{weininger1988smiles, weininger1989smiles}, which are a computer-readable linearization of molecule graphs. \citet{jaeger2018mol2vec} use the Morgan fingerprinting algorithm to convert each molecule into a `sentence' of its substructures, to  which it applies the Word2vec algorithm \cite{mikolov2013efficient, mikolov2013distributed}. \citet{duvenaud2015convolutional} use neural methods to learn fingerprints. Other advances such as BERT \citep{devlin2019bert} have also been applied to the domain, such as MolBERT \citep{fabian2020molecular} and ChemBERTa \citep{chithrananda2020chemberta}, which use SMILES strings as inputs to pretrain a BERT-esque model. 
Work has been done to use the molecule graph structure and known reactions for learning representations \cite{molecule2022}. \newcite{schwaller2021mapping} trains a BERT model to learn representations of chemical reactions. \newcite{schwaller2021extraction} leverages unsupervised representation learning with Transformers to extract an organic chemistry grammar. Unlike existing work, MolT5's molecule representations allow for translation between molecules and natural language.%

There has been particular interest in training generative models for de novo molecule discovery. %
\newcite{bagal2021molgpt} apply a GPT-style decoder for this task. \citet{lu2022unified} apply a T5 model to SMILES strings for multitask reaction prediction problems. MegaMolBART\footnote{\url{https://tinyurl.com/megamolbart}} trains a BART model on 500M SMILES strings from the ZINC-15 dataset \cite{sterling2015zinc}

\section{Conclusions and Future Work}

In this work, we propose \textbf{MolT5}, a self-supervised learning framework for pretraining models on a vast amount of unlabeled text and molecule strings. Furthermore, we propose two new tasks: molecule captioning and text-guided molecule generation, for which we explore various evaluation methods. Together, these tasks allow for translation between natural language and molecules. Using \textbf{MolT5}, we are able to obtain high scores for both tasks. %

\section{Broader Impacts}

Our proposed model and tasks will have the following broader impacts. 1) It will help to democratize molecular AI, allowing chemistry experts to take advantage of new AI technologies for discovering new life-changing drugs by interacting in the natural language, because it is most natural for humans to provide explanations and requirements in natural language. 2) Text-based molecule generation enables the ability to generate molecules with specific functions (such as taste) rather than properties, enabling the next generation of chemistry where custom molecules are used for each application. Specifically-designed molecular solutions have the potential to revolutionize fields such as medicine and material science. 3) Our models, whose weights we will release, will allow further research in the NLP community on the applications of multimodal text-molecule models.

\subsection{Risks}

MolT5, like other large language models, can potentially be abused. First, there may be biases learned by the model due to its large-scale training data. These biases may affect what type of molecules are generated when the model is prompted about certain diseases. Thus, any molecules discovered by usage of MoLT5 should strictly evaluated by standard clinical processes before being considered for medicinal use. Another risk is that the model may be used to discover potentially dangerous molecules instead of beneficial ones. It is difficult to predict what exact molecules may be discovered via usage of our work. However, while there is this unfortunate potential for misuse of the technology, 
knowledge of dangerous molecule's existence and structure is generally not harmful due to the requisite technical knowledge and laboratory resources required to synthesize them in any meaningful quantity. Overall, we believe these downsides are outweighed by the benefits to the research and pharmaceutical communities.

\section{Limitations}

Since this work focuses on a new application for large language models, many of the same limitations apply here. Namely, the model is trained on a large dataset collected from the Internet, so it may contain unintended biases. One limitation of our model is using SMILES strings -- recent work \cite{krenn2020self} proposes a string representation with validity guarantees. In practice, we found this to work poorly with pretrained T5 checkpoints (which were important from a computational perspective). We also note that some compounds in ChEBI-20 can cause validity problems in the default SELFIES implementation. We leave further investigation of this to future work. Finally, we stress that MolT5 was created for research purposes and generated molecules should not be used for medical purposes without careful evaluation by standard clinical testing first.

\section*{Acknowledgement}
We would like to thank Martin Burke for his helpful discussion. This research is based upon work supported by the Molecule Maker Lab Institute: an AI research institute program supported by NSF under award No. 2019897 and No. 2034562. The views and conclusions contained herein are those of the authors and should not be interpreted as necessarily representing the official policies, either expressed or implied, of the U.S. Government. The U.S. Government is authorized to reproduce and distribute reprints for governmental purposes notwithstanding any copyright annotation therein.

\clearpage

\bibliography{anthology,custom,emnlp}
\bibliographystyle{acl_natbib}

\clearpage
\appendix

\section{Baselines and Hyperparameters}
\label{appendix:baseline}

Any sequence-to-sequence model is applicable to our new tasks (i.e., molecule captioning and generation). We implement the following baselines:
\begin{enumerate}
    \item \textbf{RNN-GRU} \cite{cho2014learning}. We implement a 4-layer GRU recurrent neural network with a hidden size of 512. We use a learning rate of 1e-4 and a batch size of 128 for molecule generation. For caption generation, a batch size of 116 is used. The number of training epochs is 50. Additionally, the encoder is bidirectional. For training, teacher forcing is used 50\% of the time, and gradient clipping to 50 is applied.
    \item \textbf{Transformer} \cite{vaswani2017attention}. We train a vanilla Transformer model consisting of six encoder and decoder layers. The number of training epochs is 40, the batch size is 16, and the learning rate is 1e-4. We use a linear decay with a warmup of 400 steps. %
    \item \textbf{T5} \cite{raffel2020exploring}. We experiment with three public T5.1.1 checkpoints\footnote{\url{https://tinyurl.com/t511-ckpts}}: small, base, and large. We finetune each checkpoint for molecule captioning or molecule generation using the open-sourced t5x framework \cite{roberts2022t5x}. The number of training steps is set to be 50,000. The dropout rate is set to be 0.0 for the small and base models, and it is set to be 0.1 for the large model. For other hyperparameters, we use the default values provided by the t5x framework.
\end{enumerate}

We train the baseline models on the ChEBI-20 dataset using SMILES representations for the molecules. Molecule captioning and generation are trained with molecules as input/output and text as output/input. Sequences are limited to 512 tokens for input and output. During inference, a beam decoder with a beam size of 5 is used.

On the RNN and vanilla Transformer models, we use a character-split vocabulary for SMILES. For the text vocabulary, we use SciBERT's 31,090-token vocabulary \cite{beltagy2019scibert}.

\section{Reproducibility Checklist}

The programs, trained models, and resources will be made publicly available. For training the RNN and Transformer baselines, we use NVIDIA Tesla V100 GPUs. For pretraining and finetuning T5-related models, we use TPUs.

When testing on a MacBook Pro that has no access to GPUs, the average inference time of our MolT5-Base molecule generation model is 2.24 seconds/query. The average inference time of our large MolT5-Base molecule captioning model is 9.86 seconds/query.

\section{Decoding with Huggingface Model}

For ease of adoption, we converted our original models trained using the t5x framework \cite{roberts2022t5x} to HuggingFace-based models \cite{Wolf2019HuggingFacesTS}. We will release the converted models on HuggingFace (HF) Hub. Due to implementation differences, the HF-based models produce slightly different outputs from the original models. Therefore, we also report the numbers of the HF-based models in Table \ref{tab:results_captioning_hf} and Table \ref{tab:results_generation_hf}.

\begin{table*}[ht!]
\resizebox{\textwidth}{!}{
\centering
\tiny
\begin{tabular}{ c|c|c|c|c|c|c|c }

\multicolumn{1}{c}{\textbf{Model}} & \multicolumn{1}{c}{BLEU-2} & \multicolumn{1}{c}{BLEU-4} & \multicolumn{1}{c}{ROUGE-1} & \multicolumn{1}{c}{ROUGE-2} & \multicolumn{1}{c}{ROUGE-L} & \multicolumn{1}{c}{METEOR} & \multicolumn{1}{c}{Text2Mol} \\
\thickhline
Ground Truth& &  &  &  &  &  & 0.609 \\\hline
RNN & 0.251 & 0.176 & 0.450 & 0.278 & 0.394 & 0.363 & 0.426 \\
Transformer & 0.061 & 0.027 & 0.204 & 0.087 & 0.186 & 0.114 & 0.057 \\\hline
T5-Small & 0.515 & 0.424 & 0.613 & 0.459 & 0.568 & 0.538 & 0.527 \\
MolT5-Small & 0.532 & 0.445 & 0.627 & 0.477 & 0.583 & 0.557 & 0.543 \\\hline
T5-Base & 0.522 & 0.432 & 0.616 & 0.461 & 0.572 & 0.545 & 0.524 \\
MolT5-Base & 0.551 & 0.464 & 0.637 & 0.489 & 0.594 & 0.574 & 0.549 \\\hline
T5-Large & 0.555 & 0.464 & 0.632 & 0.482 & 0.585 & 0.588 & 0.564 \\
MolT5-Large & \textbf{0.588} & \textbf{0.502} & \textbf{0.650} & \textbf{0.507} & \textbf{0.604} & \textbf{0.614} & \textbf{0.582} \\
\end{tabular}
}
\caption{HuggingFace model molecule captioning results for the different baseline models on the test split of CheBI-20. Rouge scores are F1 values.}
\label{tab:results_captioning_hf}
\end{table*}

\begin{table*}[ht!]
\resizebox{\textwidth}{!}{
\centering
\begin{tabular}{ c|c|c|c|c|c|c|c|c|c }

\multicolumn{1}{c}{\textbf{Model}} & \multicolumn{1}{c}{BLEU$\uparrow$} & \multicolumn{1}{c}{Exact$\uparrow$} & \multicolumn{1}{c}{Levenshtein$\downarrow$} & \multicolumn{1}{c}{MACCS FTS$\uparrow$} & \multicolumn{1}{c}{RDK FTS$\uparrow$} & \multicolumn{1}{c}{Morgan FTS$\uparrow$} & \multicolumn{1}{c}{FCD$\downarrow$} & \multicolumn{1}{c}{Text2Mol$\uparrow$} & \multicolumn{1}{c}{Validity$\uparrow$} \\
\thickhline
Ground Truth & 1.000 & 1.000 & 0.0 & 1.000 & 1.000 & 1.000 & 0.0 & 0.609 & 1.0 \\\hline
RNN & 0.652 & 0.005 & 38.09 & 0.591 & 0.400 & 0.362 & 4.55 & 0.409 & 0.542 \\
Transformer & 0.499 & 0.000 & 57.66 & 0.480 & 0.320 & 0.217 & 11.32 & 0.277 & 0.906 \\\hline
T5-Small & 0.740 & 0.061 & 30.05 & 0.798 & 0.681  & 0.623  & 1.77 & 0.541 & 0.597 \\
MolT5-Small & 0.749 & 0.082 & 28.816 & 0.780 & 0.654 & 0.601 & 1.35 & 0.535 & 0.725 \\
MolT5-Small-HV & 0.613 & 0.075 & 30.458 & 0.699 & 0.547 & 0.482 & 1.44 & 0.479 & 0.983 \\\hline
T5-Base & 0.769 & 0.067 & 27.112 & 0.816 &  0.701 & 0.637 & 1.44 & 0.554 &  0.654  \\
MolT5-Base & 0.783 & 0.082 & 24.846 & 0.788 &  0.661 & 0.602 & 1.16 & 0.544 & 0.787 \\
MolT5-Base-HV & 0.661 & 0.073 & 28.276 & 0.721 & 0.579 & 0.509 & 1.38 & 0.501 & 0.979\\\hline
T5-Large &  0.856 & 0.285 & 16.845 & 0.877 &  0.794 & 0.732 & 0.40 & 0.587 & 0.959 \\
MolT5-Large & \textbf{0.858} & \textbf{0.318} & \textbf{15.957} & \textbf{0.890} & \textbf{0.813} & \textbf{0.750} & \textbf{0.38} & \textbf{0.590} & 0.958 \\
MolT5-Large-HV & 0.810 & 0.314 & 16.758 & 0.872 & 0.786 & 0.722 & 0.44 & 0.582 & \textbf{0.996} \\
\end{tabular}
}
\caption{HuggingFace model de novo molecule generation results for the different baseline models on the test split of CheBI-20. MolT5-Small-HV, MolT5-Base-HV, and MolT5-Large-HV are models that use a high-validity decoding process--see Appendix \ref{appendix:high_valid}.
}
\label{tab:results_generation_hf}
\end{table*}

\section{High Validity Molecule Generation}
\label{appendix:high_valid}

To increase the validity score of the molecule generation models, we consider a high-validity decoding strategy. We use diverse beam search \cite{vijayakumar2016diverse} with a beam width and beam group of 30 and a diversity penalty of 0.5. Then, we use RDKit \cite{Landrum2021RDKit2021_03_2} to select the first valid beam. On rare occasions, the beam size exceeds memory limitations, so we iteratively reduce the beam size by 5 for that input and try again. In Table \ref{tab:results_generation_hf}, MolT5-Small-HV, MolT5-Base-HV, and MolT5-Large-HV denote models that use this decoding process.

\section{Ablations}
\label{appendix:ablations}
We perform ablations on MolT5-Small pretraining. For molecule captioning (Table \ref{tab:ablation_results_captioning}), pretraining on both C4 and ZINC is clearly more beneficial than pretraining only on C4 or only on ZINC.

For molecule generation, at first glance, pretraining on C4+ZINC seems not to outperform pretraining only on C4 (Table \ref{tab:ablation_results_generation}). However, note that except for BLEU, Exact, Levenshtein, and Validity, other metrics in Table \ref{tab:ablation_results_generation} are computed using only syntactically valid molecules. Table \ref{tab:normalized_ablation_results_generation} shows the normalized molecule generation results. After normalization, we see that pretraining on C4+ZINC outperforms pretraining only on C4 or only on ZINC according to most metrics. Finally, pretraining only on ZINC increases the validity score substantially. However, this leads to decreased similarity of the generated molecules to the ground truths.

\begin{table*}[ht!]
\resizebox{\textwidth}{!}{
\centering
\tiny
\begin{tabular}{ c|c|c|c|c|c|c|c }

\multicolumn{1}{c}{\textbf{Pretraining}} & \multicolumn{1}{c}{BLEU-2} & \multicolumn{1}{c}{BLEU-4} & \multicolumn{1}{c}{ROUGE-1} & \multicolumn{1}{c}{ROUGE-2} & \multicolumn{1}{c}{ROUGE-L} & \multicolumn{1}{c}{METEOR} & \multicolumn{1}{c}{Text2Mol} \\
\thickhline
Ground Truth &  &  &  &  &  &  & 0.609 \\\hline
C4-Only & 0.523 & 0.433 & 0.616 & 0.463 & 0.571 & 0.545 & 0.530 \\
ZINC-Only & 0.519 & 0.434 & 0.619 & 0.466 & 0.573 & 0.548 & 0.538 \\
C4+ZINC & 0.532 & 0.445 & 0.627 & 0.477 & 0.583 & 0.557 & 0.543 \\
\end{tabular}
}
\caption{Pretraining ablation results of molecule captioning for MolT5-Small on the test split of CheBI-20. Rouge scores are F1 values.}
\label{tab:ablation_results_captioning}
\end{table*}

\begin{table*}[ht!]
\centering
\resizebox{\textwidth}{!}{
\begin{tabular}{ c|c|c|c|c|c|c|c|c|c }

\multicolumn{1}{c}{\textbf{Pretraining}} & \multicolumn{1}{c}{BLEU$\uparrow$} & \multicolumn{1}{c}{Exact$\uparrow$} & \multicolumn{1}{c}{Levenshtein$\downarrow$} & \multicolumn{1}{c}{MACCS FTS$\uparrow$} & \multicolumn{1}{c}{RDK FTS$\uparrow$} & \multicolumn{1}{c}{Morgan FTS$\uparrow$} & \multicolumn{1}{c}{FCD$\downarrow$} & \multicolumn{1}{c}{Text2Mol$\uparrow$} & \multicolumn{1}{c}{Validity$\uparrow$} \\
\thickhline
Ground Truth &  &  &  &  &  &  & 0.0 & 0.609 & 1.0 \\\hline
C4-Only & 0.771 & 0.081 & 26.84 & 0.811 & 0.697 & 0.641  & 2.99 & 0.555 & 0.635 \\
ZINC-Only & 0.716 & 0.063 & 32.953 & 0.701 & 0.576 & 0.524 & 2.75 & 0.463 & 0.807 \\
C4+ZINC & 0.749 & 0.082 & 28.816 & 0.78 & 0.654 & 0.601 & 2.60 & 0.535 & 0.725 \\
\end{tabular}
}
\caption{Pretraining ablation results of molecule generation for MolT5-Small on the test split of CheBI-20. 
}
\label{tab:ablation_results_generation}
\end{table*}

\begin{table*}[ht!]
\centering
\resizebox{\textwidth}{!}{
\begin{tabular}{ c|c|c|c|c|c|c|c|c|c }

\multicolumn{1}{c}{\textbf{Pretraining}} & \multicolumn{1}{c}{BLEU$\uparrow$} & \multicolumn{1}{c}{Exact$\uparrow$} & \multicolumn{1}{c}{Levenshtein$\downarrow$} & \multicolumn{1}{c}{MACCS FTS$\uparrow$} & \multicolumn{1}{c}{RDK FTS$\uparrow$} & \multicolumn{1}{c}{Morgan FTS$\uparrow$} & \multicolumn{1}{c}{FCD$\downarrow$} & \multicolumn{1}{c}{Text2Mol$\uparrow$} & \multicolumn{1}{c}{Validity$\uparrow$} \\
\thickhline
Ground Truth &  &  &  &  &  &  & 0.0 & 0.609 & 1.0 \\\hline
C4-Only & 0.771 & 0.081 & 26.84 & 0.51499 & 0.44259 & 0.40704  & 4.71 & 0.35243 & 0.635 \\
ZINC-Only & 0.716 & 0.063 & 32.953 & 0.56571  & 0.46483 & 0.42287 & 3.41 & 0.37364 & 0.807 \\
C4+ZINC & 0.749 & 0.082 & 28.816 & 0.5655 & 0.47415 & 0.43572 & 3.59 &  0.38788 & 0.725 \\
\end{tabular}
}
\caption{Normalized pretraining ablation results of molecule generation for MolT5-Small on the test split of CheBI-20. Molecule-based results (FTS, FCD, Text2Mol) are normalized by multiplying by validity (for scores where higher is better) or dividing by validity (for scores where lower is better).
}
\label{tab:normalized_ablation_results_generation}
\end{table*}

\section{More Examples}
\label{appendix:examples}
\clearpage

\begin{figure*}
\centering
\includegraphics[width=\textwidth]{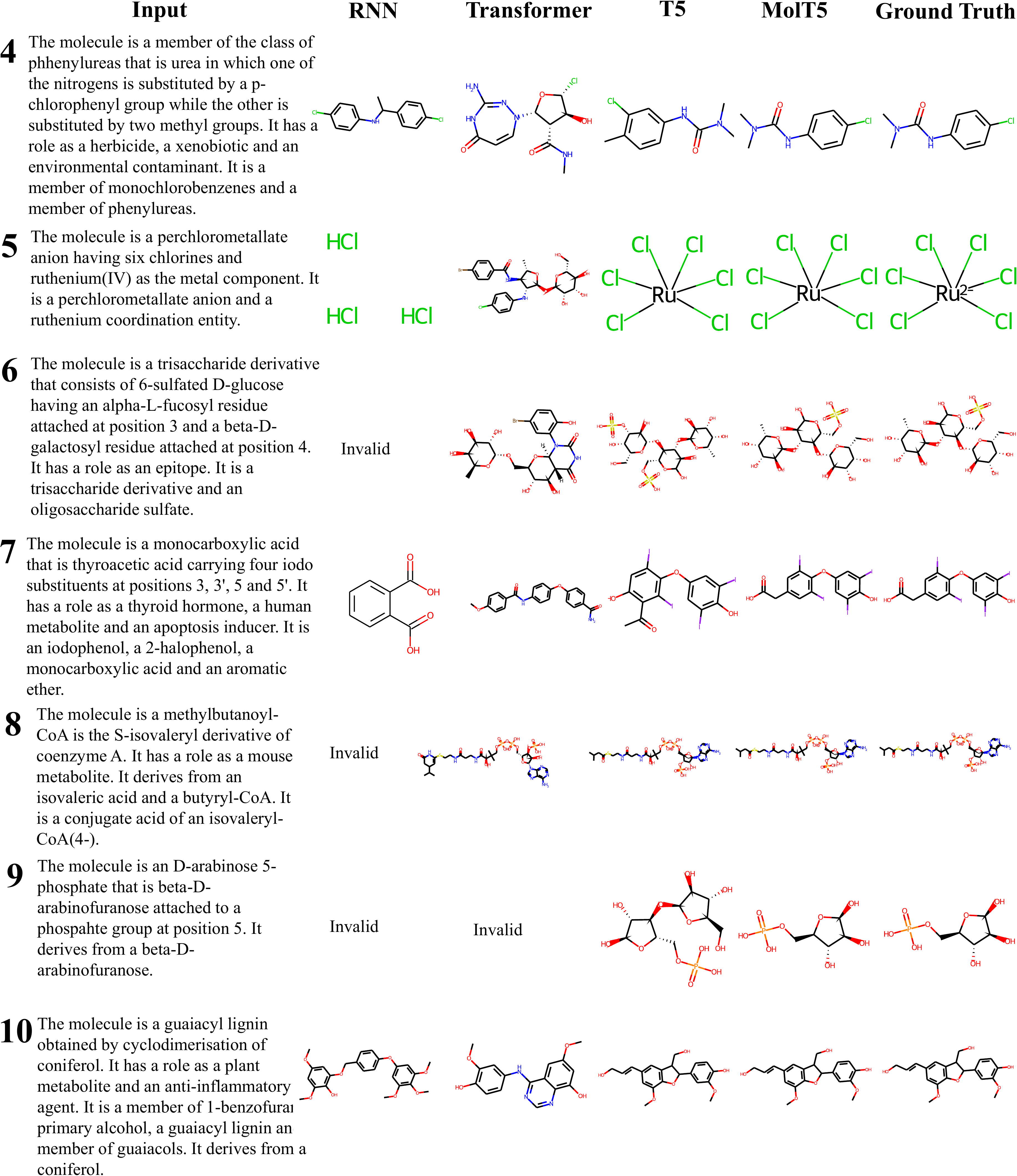}
\caption{More examples of interesting molecules generated by different models.}
\label{fig:qual2}
\end{figure*}

\begin{figure*}
\centering
\includegraphics[width=\textwidth]{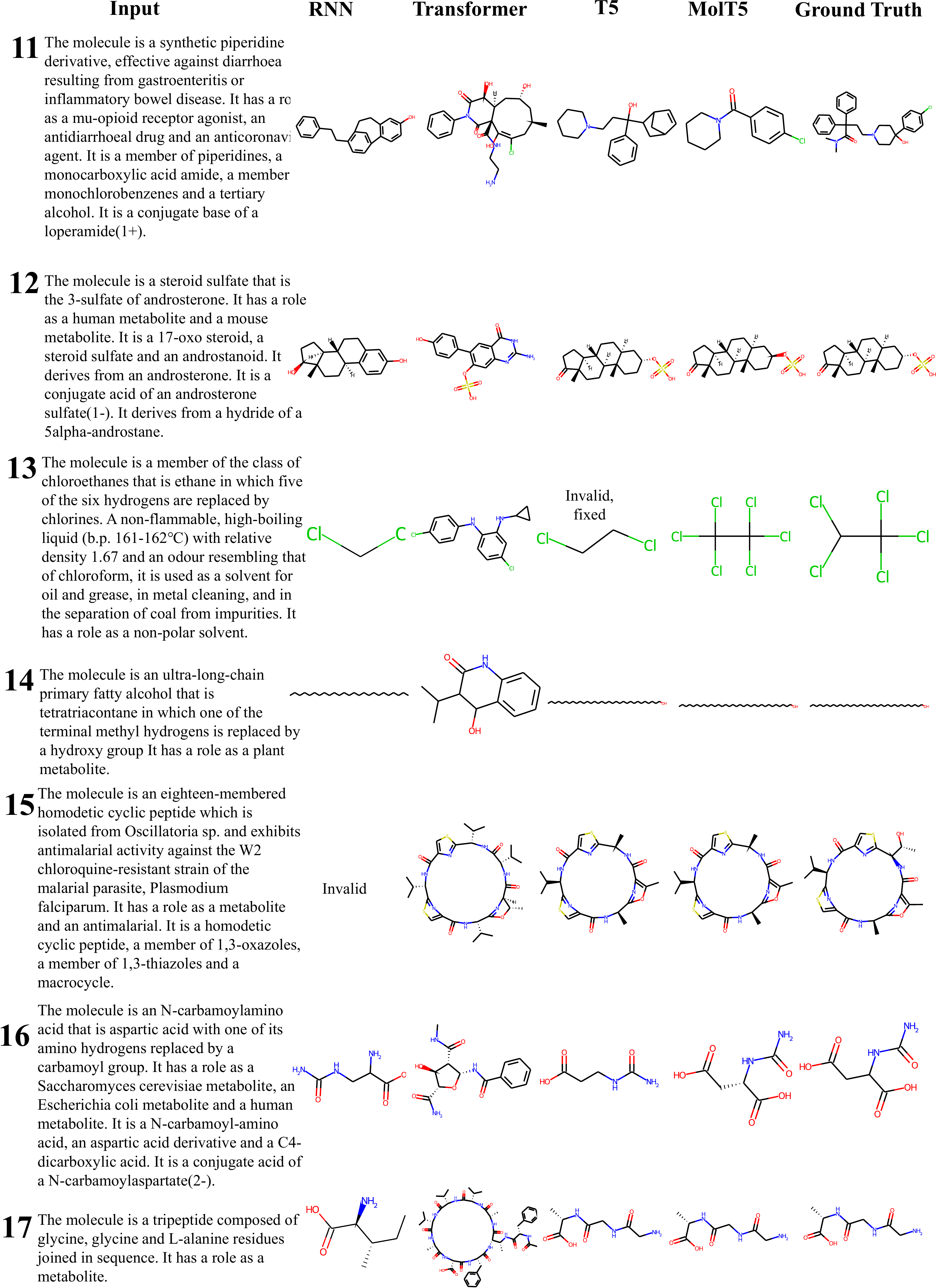}
\caption{More examples of interesting molecules generated by different models.}
\label{fig:qual3}
\end{figure*}

\begin{figure*}
\centering
\includegraphics[width=\textwidth]{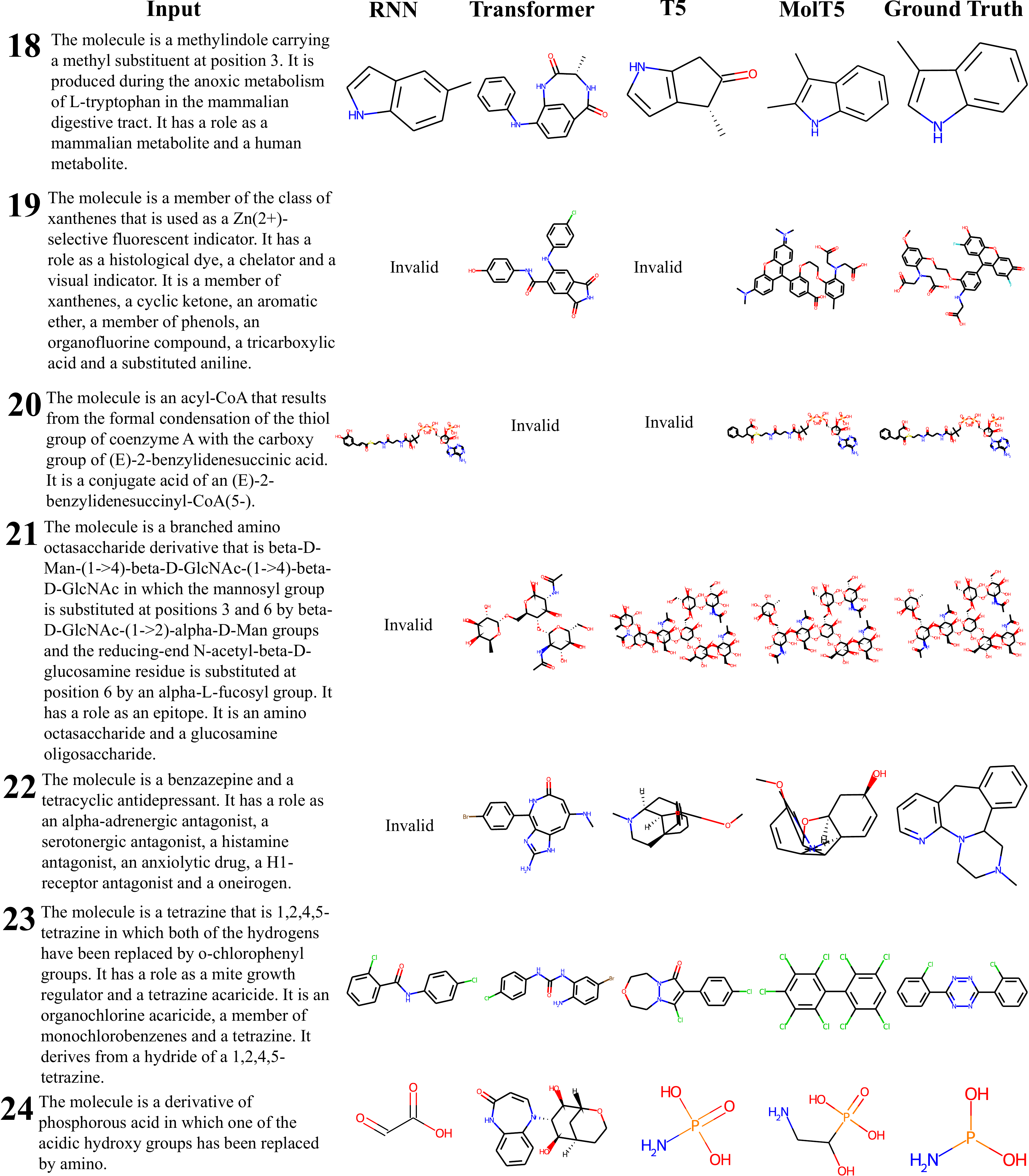}
\caption{More examples of interesting molecules generated by different models.}
\label{fig:qual4}
\end{figure*}

\begin{figure*}
\centering
\includegraphics[width=\textwidth]{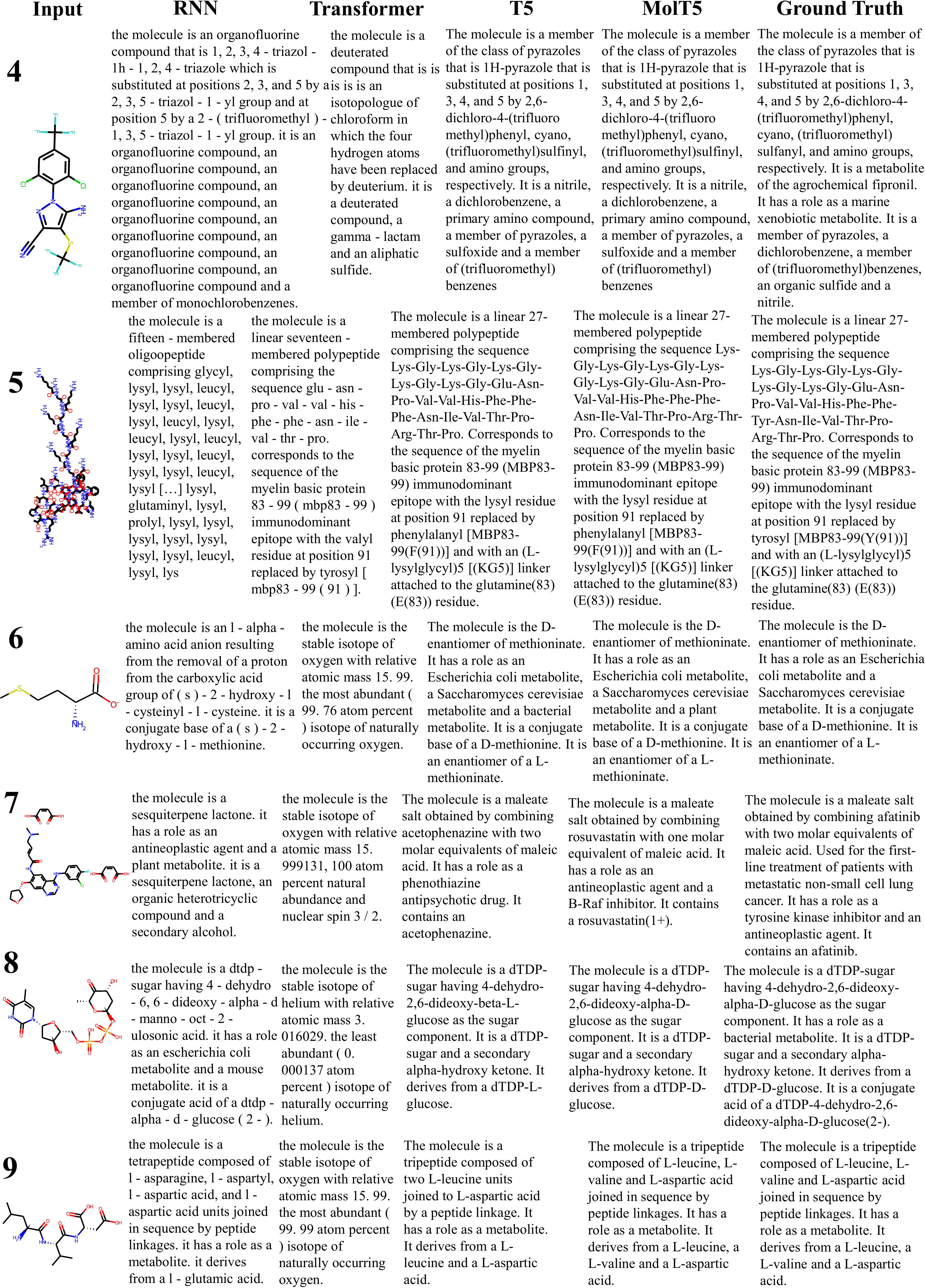}
\caption{More examples of interesting captions generated by different models.}
\label{fig:qual1_cap_appendix}
\end{figure*}

\clearpage

\section{Testing Model Diversity with Retrieval}
\label{appendix:retrieval_diversity}

To test the diversity of generations, we apply a Text2Mol \cite{edwards2021text2mol} cross-modal retrieval model to the entire generated set of molecules or descriptions. In the case of molecules, we first take the molecules generated for our test set. We consider these molecules as our corpus and then use the descriptions (which were used to generate the molecules in the first place) as our queries. So, for each query we look at the rank of its generated molecule (the highest rank is 1). This process tests whether the Text2Mol retrieval model can differentiate between the generated (valid) molecules. Doing so means it can retrieve a specific molecule when given the description used to generate it. If the generative model did not sufficiently take the descriptions into consideration, then the retrieval model won't be able to distinguish between generated molecules and the scores will be very low (such as the transformer model, which frequently generates the same molecule/caption). 

As an example, consider that we have 10 descriptions of molecules.

\begin{table}[h!]
\resizebox{\textwidth}{!}{
\centering
\tiny
\begin{tabular}{ c|c|c|c|c|c|c }

\multicolumn{1}{c}{\textbf{Model}} & \multicolumn{1}{c}{Mean Rank} & \multicolumn{1}{c}{MRR} & \multicolumn{1}{c}{Hits@1} & \multicolumn{1}{c}{Hits@10} & \multicolumn{1}{c}{Hits@100} & \multicolumn{1}{c}{Validity} \\
\thickhline
Ground Truth & 4.9 & 0.735 & 60.4\% & 95.2\% & 99.5\% & 100\% \\\hline
RNN & 106.7 & 0.192 & 10.45\% & 37.0\% & 74.4\% & 54.2\% \\
Transformer & 426.4 & 0.106 & 5.62\% & 19.8\% & 46.2\% & 90.6\% \\\hline
T5-Small & 113.9 & 0.441 & 33.0\% & 64.1\% & 81.1\% & 60.7\% \\
MolT5-Small & 126.2 & 0.413 & 30.1\% & 62.2\% & 80.2\% & 72.1\% \\\hline
T5-Base & 97.8 & 0.467 & 35.5\% & 67.1\% & 84.7\% & 66.0\% \\
MolT5-Base & 113.9 & 0.438 & 32.3\% & 65.2\% & 83.7\% & 77.2\% \\\hline
T5-Large & 84.5 & 0.586 & 46.5\% & 81.2\% & 90.5\% & 90.2\% \\
MolT5-Large & 87.4 & 0.570 & 44.6\% & 80.1\% & 91.0\% & 90.5\% \\
\end{tabular}
}
\captionsetup{width=.8\textwidth,margin={0cm,-8cm}}
\caption{Retrieval of generated molecules on the test split of CheBI-20.}
\label{tab:results_mol_retrieval}
\end{table}

\begin{table}[h!]
\resizebox{\textwidth}{!}{
\centering
\tiny
\begin{tabular}{ c|c|c|c|c|c }

\multicolumn{1}{c}{\textbf{Model}} & \multicolumn{1}{c}{Mean Rank} & \multicolumn{1}{c}{MRR} & \multicolumn{1}{c}{Hits@1} & \multicolumn{1}{c}{Hits@10} & \multicolumn{1}{c}{Hits@100} \\
\thickhline
Ground Truth & 5.6 & 0.703 & 56.4\% & 94.3\% & 99.3\% \\\hline
RNN & 137.0 & 0.160 & 7.45\% & 34.0\% & 73.6\% \\
Transformer & 1750 & 0.007 & 00.4\% & 01.2\% & 03.2\% \\\hline
T5-Small & 60.3 & 0.414 & 28.4\% & 65.5\% & 88.1\% \\
MolT5-Small & 47.7 & 0.460 & 32.6\% & 70.7\% & 91.4\% \\\hline
T5-Base & 69.2 & 0.414 & 28.9\% & 65.0\% & 87.0\% \\
MolT5-Base & 40.2 & 0.465 & 32.5\% & 72.5\% & 92.0\% \\\hline
T5-Large & 29.4 & 0.499 & 35.5\% & 77.6\% & 94.2\% \\
MolT5-Large & 16.1 & 0.558 & 40.4\% & 84.2\% & 96.8\% \\
\end{tabular}
}
\captionsetup{width=.8\textwidth,margin={0cm,-8cm}}
\caption{Retrieval of generated captions on the test split of CheBI-20.}
\label{tab:results_desc_retrieval}
\end{table}

\newpage
\noindent For each description, we use a generative model to generate a molecule. Now, we treat these 10 generated molecules as our corpus. Using our retrieval model, we now consider each description as a query and try to retrieve the molecule that was generated from that description. If the retrieval model performs poorly, that means the molecules which were generated are difficult to distinguish from one another. By using this method with different generative models, we measure the relative diversity of generated molecules along with how well the generated molecules match the description.

Results are reported in Tables \ref{tab:results_mol_retrieval} and \ref{tab:results_desc_retrieval} for retrieving generated molecules from descriptions and for retrieving generated descriptions from molecules, respectively. We use the same Text2Mol model for retrieval here as in the Text2Mol metric. For description of metrics, see \cite{edwards2021text2mol}. Results indicate that MolT5 model generations are sufficiently distinct to be retrievable. In contrast, the outputs of the captioning transformer are essentially indistinguishable for the retrieval model.

\newpage
\section{Statistical Significance} \label{appendix:stats_sign}

To strengthen the quantitative results, we conducted statistical tests between T5-Large and MolT5-Large. For molecule captioning, we carried out paired t-tests. The computed p-values and test statistics are:

\begin{itemize}[nosep]
    \item For ROUGE-1, the p-value is 1.53e-22. The test statistic is -9.841.
    \item For ROUGE-2, the p-value is 3.27e-26. The test statistic is -10.683.
    \item For ROUGE-L, the p-value is 3.58e-21. The test statistic is -9.509.
    \item For METEOR, the p-value is 2.02e-21. The test statistic is -9.57.
    \item For Text2Mol, the p-value is 1.053e-29. The test statistic is -11.431.
\end{itemize}
Note that for every metric above, the higher the score, the better the performance. Since all the test statistics are negative and the p-values are extremely small, MolT5-Large produces significant improvements over T5-Large on the task of molecule captioning.
\\~\\
For molecule generation, we conducted independent t-tests to compare between T5-Large and MolT5-Large:
\begin{itemize}[nosep]
    \item For MACCS FTS, the p-value is 0.008. The test statistic is -2.652.
    \item For RDK FTS, the p-value is 0.0092. The test statistic is -2.604.
    \item For Morgan FTS, the p-value is 0.0153. The test statistic is -2.426.
    \item For Levenshtein, the p-value is 0.064. The test statistic is 1.8544704091978725.
    \item For Text2Mol, the p-value is 0.168. The test statistic is -1.376724743237994.
\end{itemize}
Note that for Levenshtein, the lower the score, the better the performance. We see that the test statistics for all metrics except Levenshtein is negative. In addition, while the p-values now are typically larger than the ones computed for  molecule captioning, the p-values for molecule generation are still reasonably small. Therefore, we can still conclude that MolT5-Large also produces significant improvements over T5-Large on the task of molecule generation.

\section{NLP Capabilities of MolT5}
We finetune our MolT5-based models on some GLUE tasks and see similar results for MolT5 and T5. For example, our finetuned MolT5-base model achieved an accuracy score of 95.6\% on SST-2. For comparison, T5-base achieved a score of 95.2\%. Since our self-supervised learning framework uses a large amount of natural language text in addition to SMILES string, it is reasonable that our MolT5-based models still possess ``typical'' NLP capabilities.

\section{Model Probing Tests}
\label{appendix:probing}

To generate a variety of output molecules given a single input, we employ diverse beam search \cite{vijayakumar2016diverse} with a beam width and beam group of 30 and a diversity penalty of 0.5. The goal of these tests (shown in the following figures) is to explore molecule outputs given very specific desired properties. Note that these brief input descriptions are out-of-distribution from the finetuning data. In the following figures, the top 10 valid molecules are shown for each prompt (order: left to right, top to bottom).

\begin{figure*}
\centering
\includegraphics[width=\textwidth]{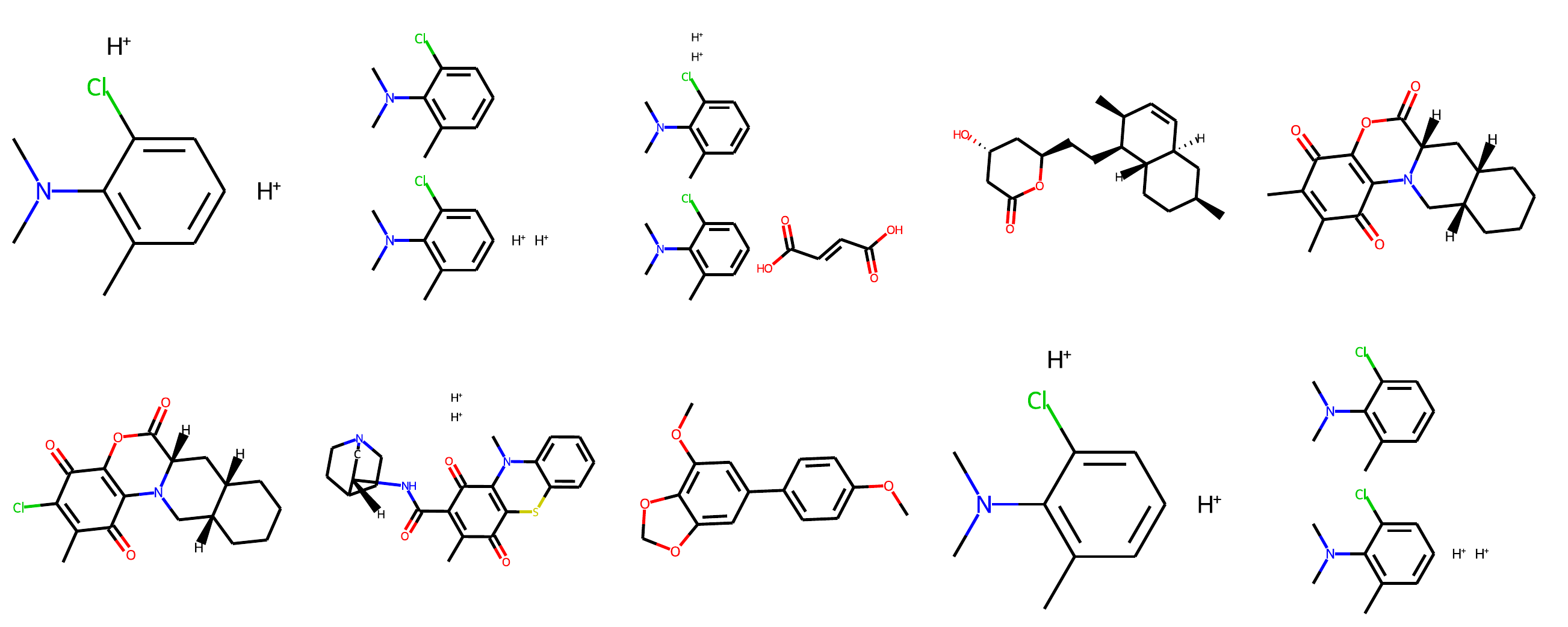}
\caption{Input: The molecule displays antimalarial properties.}
\end{figure*}

\begin{figure*}
\centering
\includegraphics[width=\textwidth]{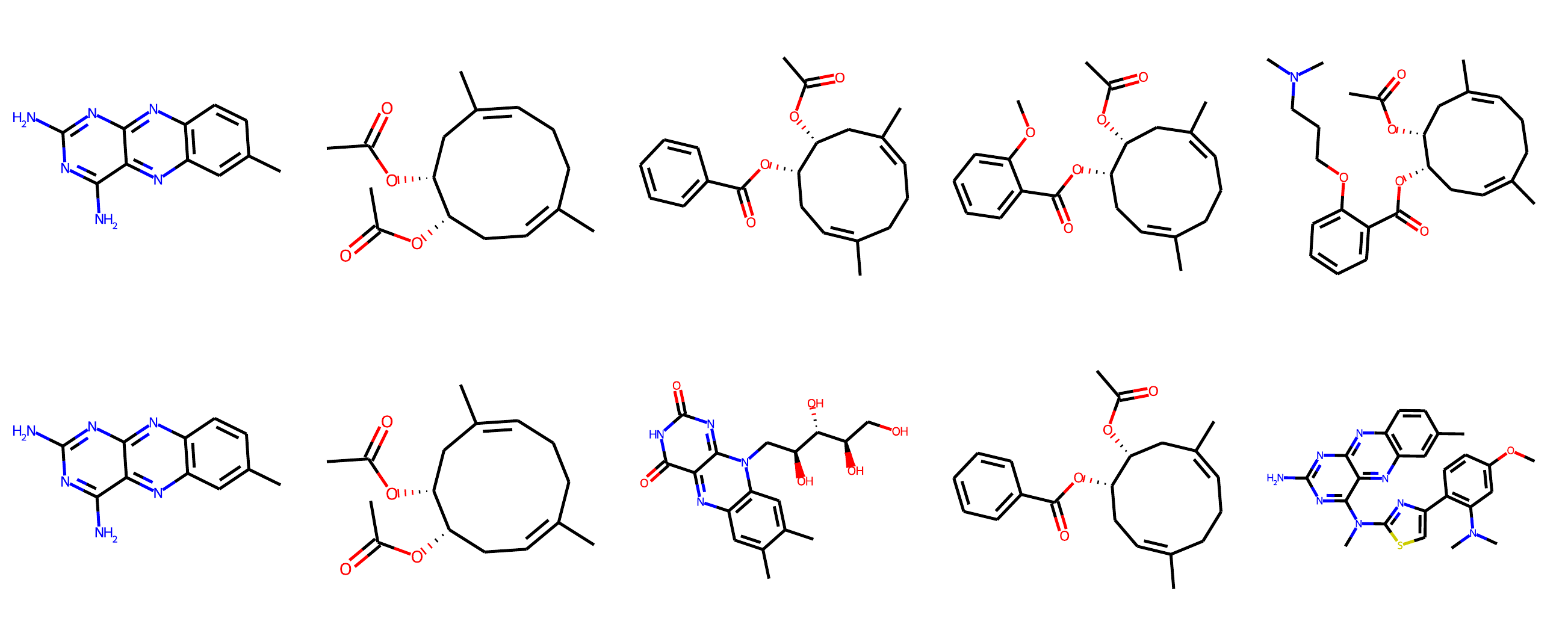}
\caption{Input: The molecule is a apoptosis inducer.}
\end{figure*}

\begin{figure*}
\centering
\includegraphics[width=\textwidth]{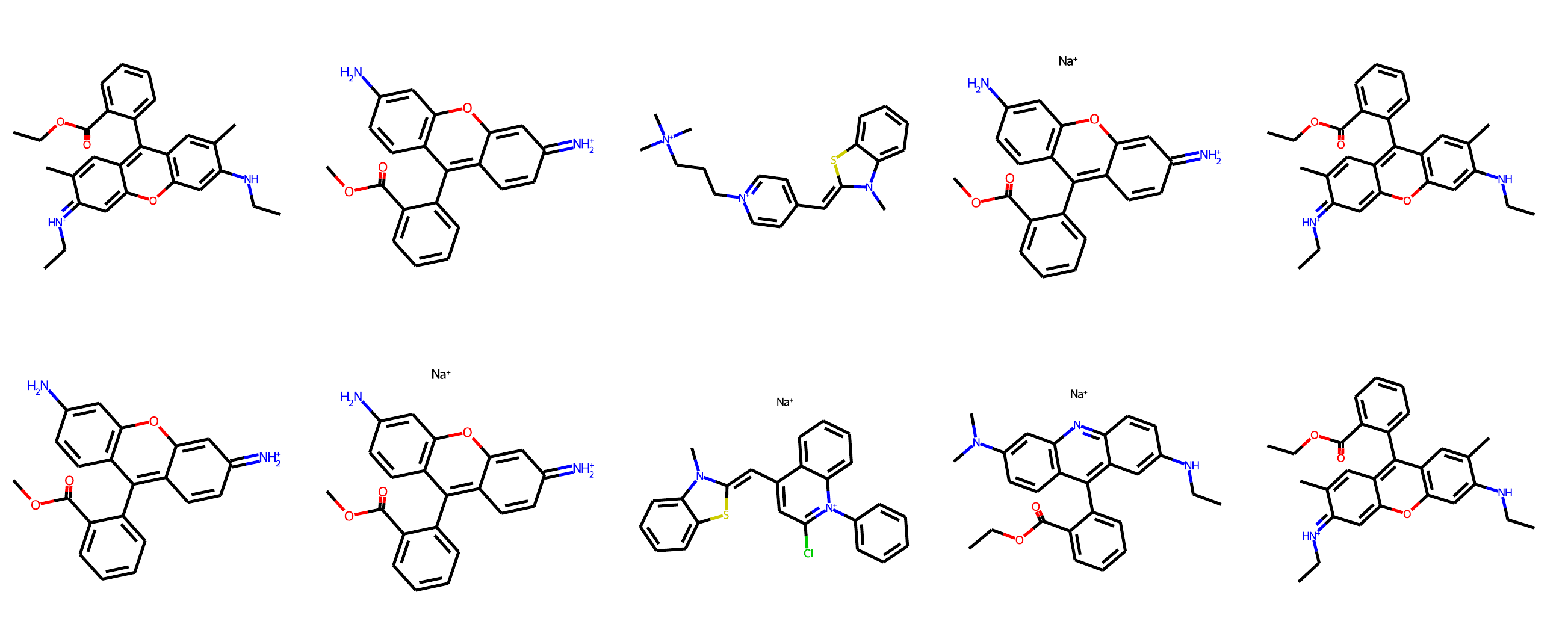}
\caption{Input: The molecule is a blue dye.}
\end{figure*}

\begin{figure*}
\centering
\includegraphics[width=\textwidth]{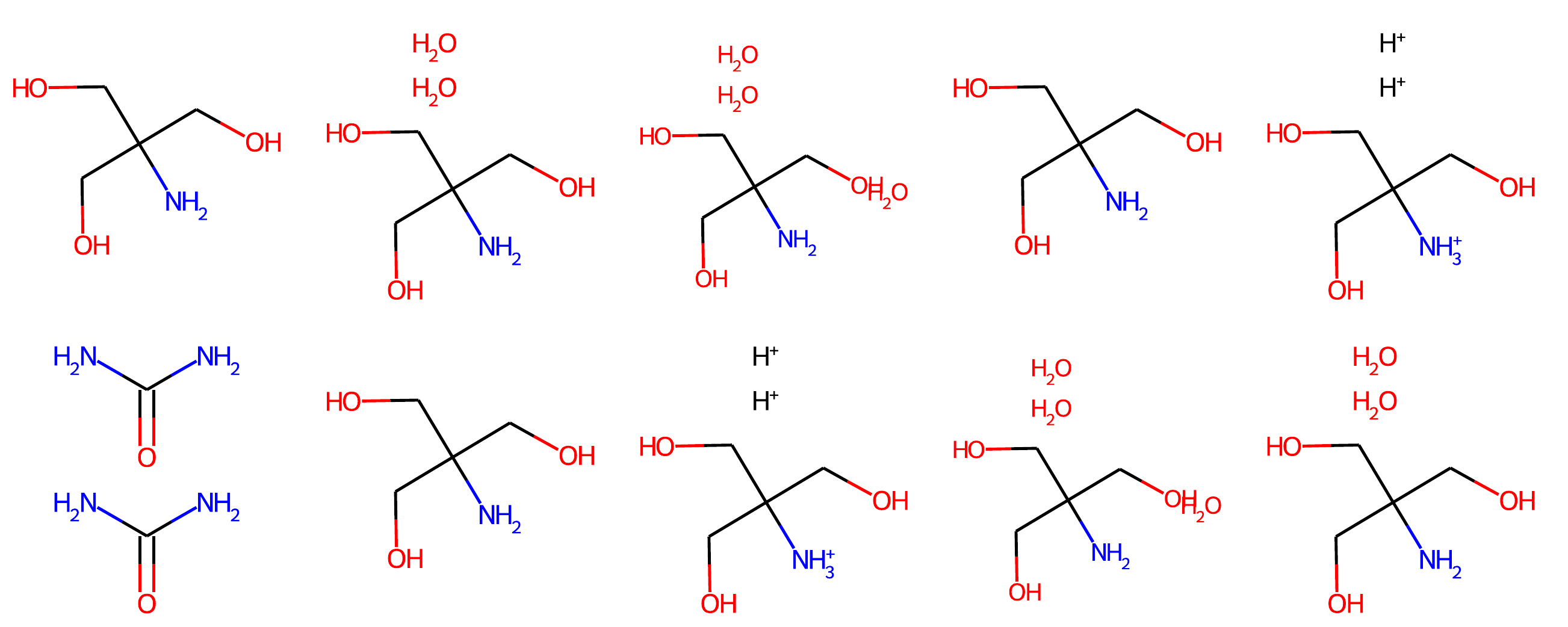}
\caption{Input: The molecule is a coagulent.}
\end{figure*}

\begin{figure*}
\centering
\includegraphics[width=\textwidth]{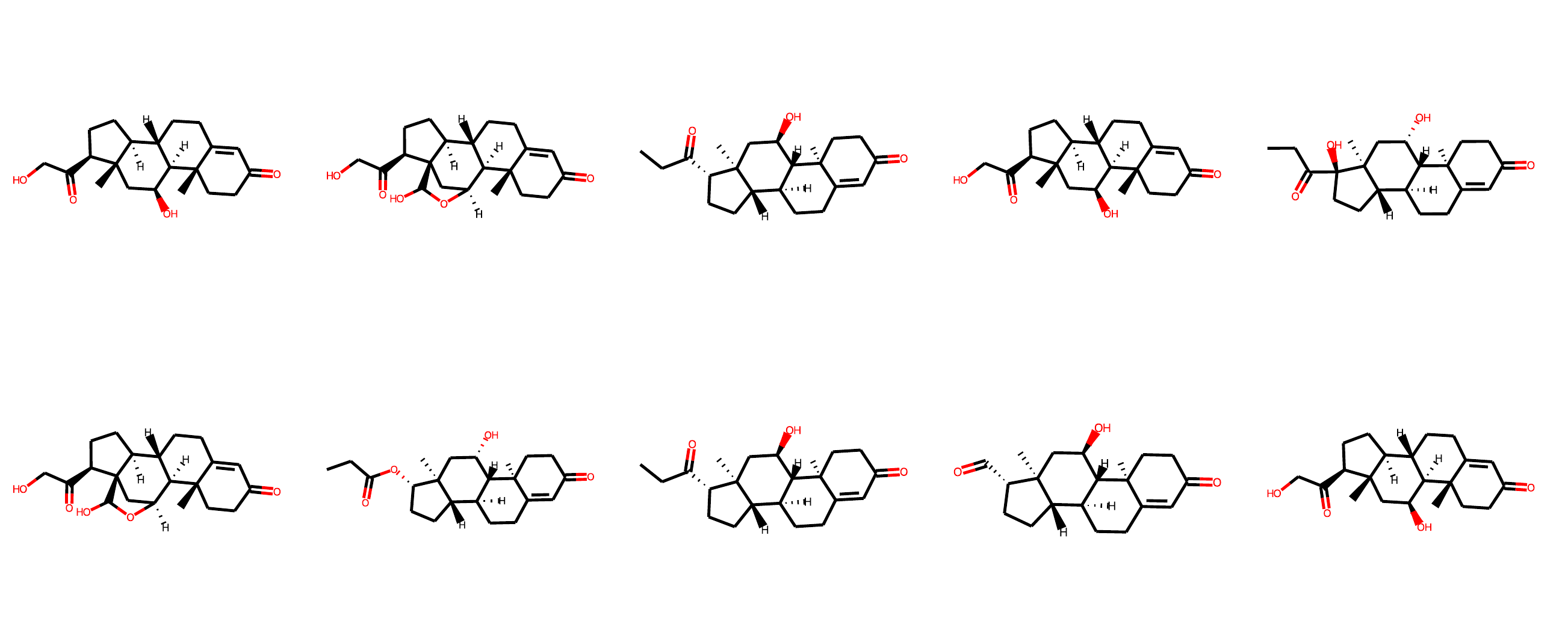}
\caption{Input: The molecule is a corticosteroid.}
\end{figure*}

\begin{figure*}
\centering
\includegraphics[width=\textwidth]{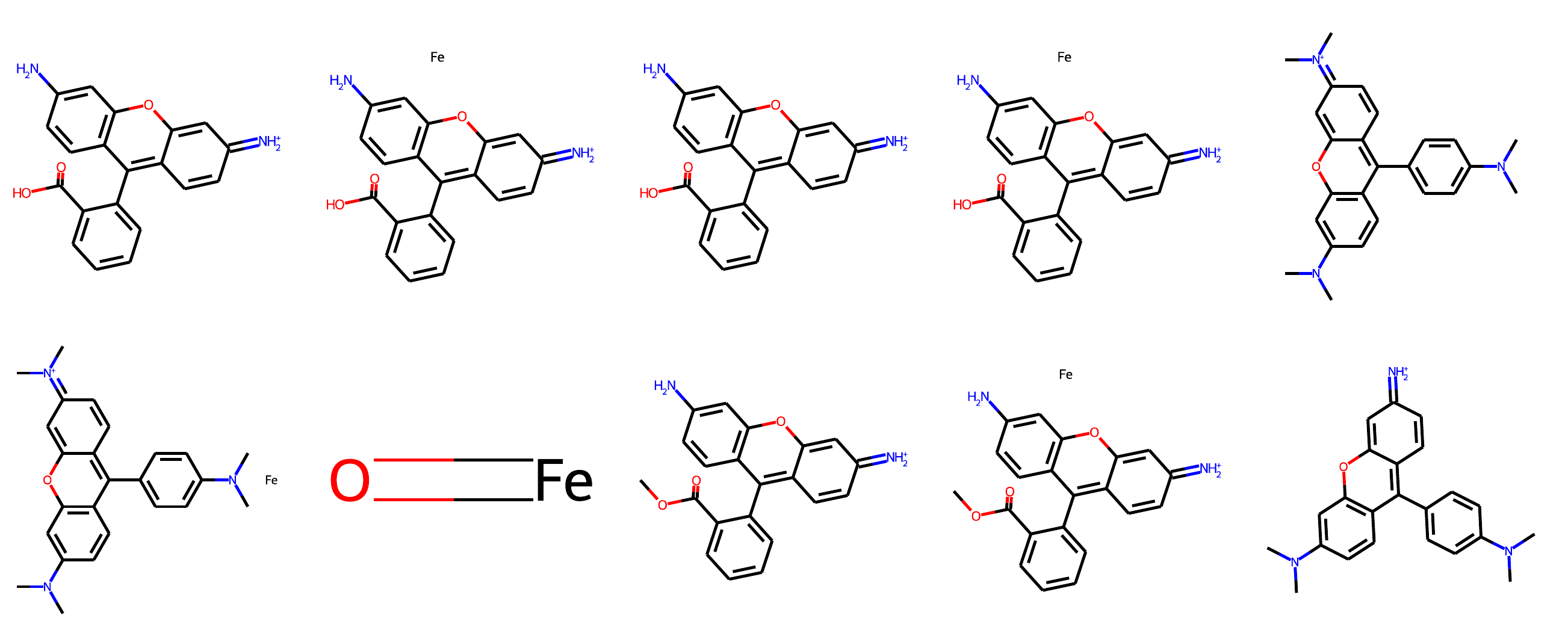}
\caption{Input: The molecule is a fluorochrome.}
\end{figure*}

\begin{figure*}
\centering
\includegraphics[width=\textwidth]{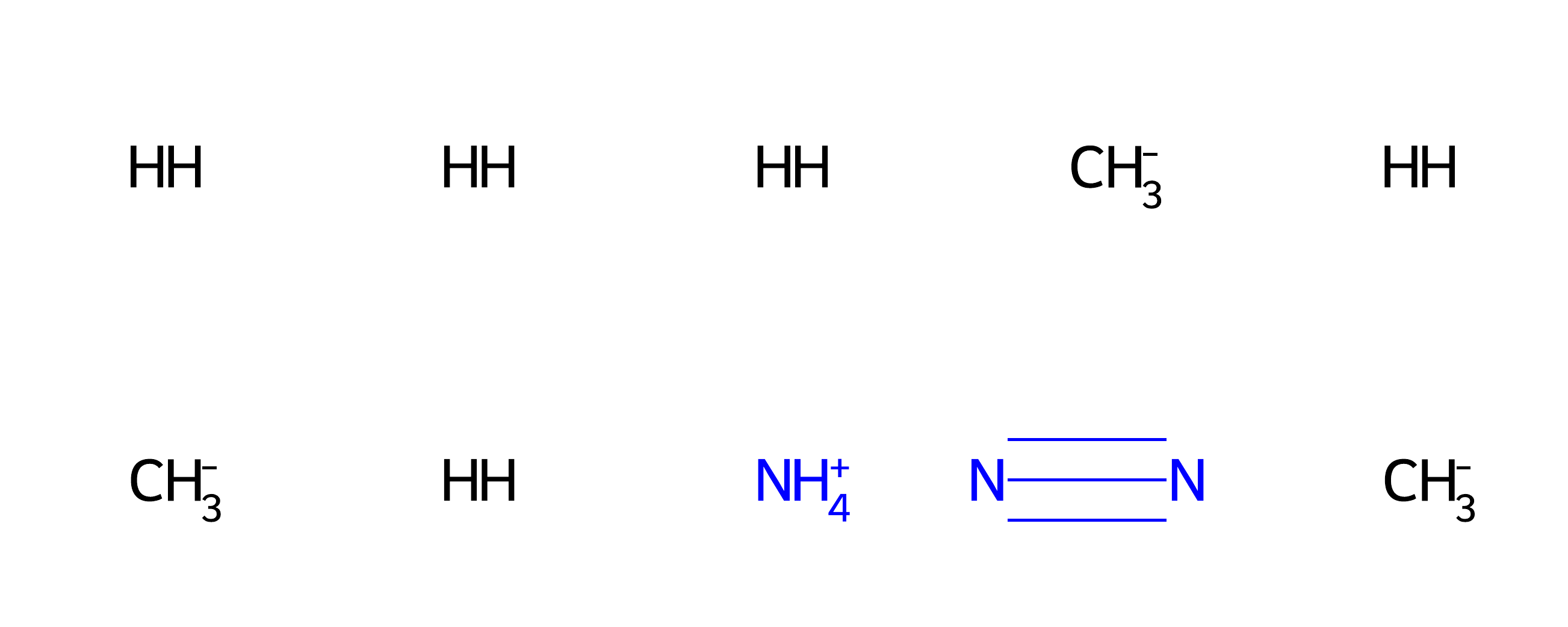}
\caption{Input: The molecule is a gas at room temperature.}
\end{figure*}

\begin{figure*}
\centering
\includegraphics[width=\textwidth]{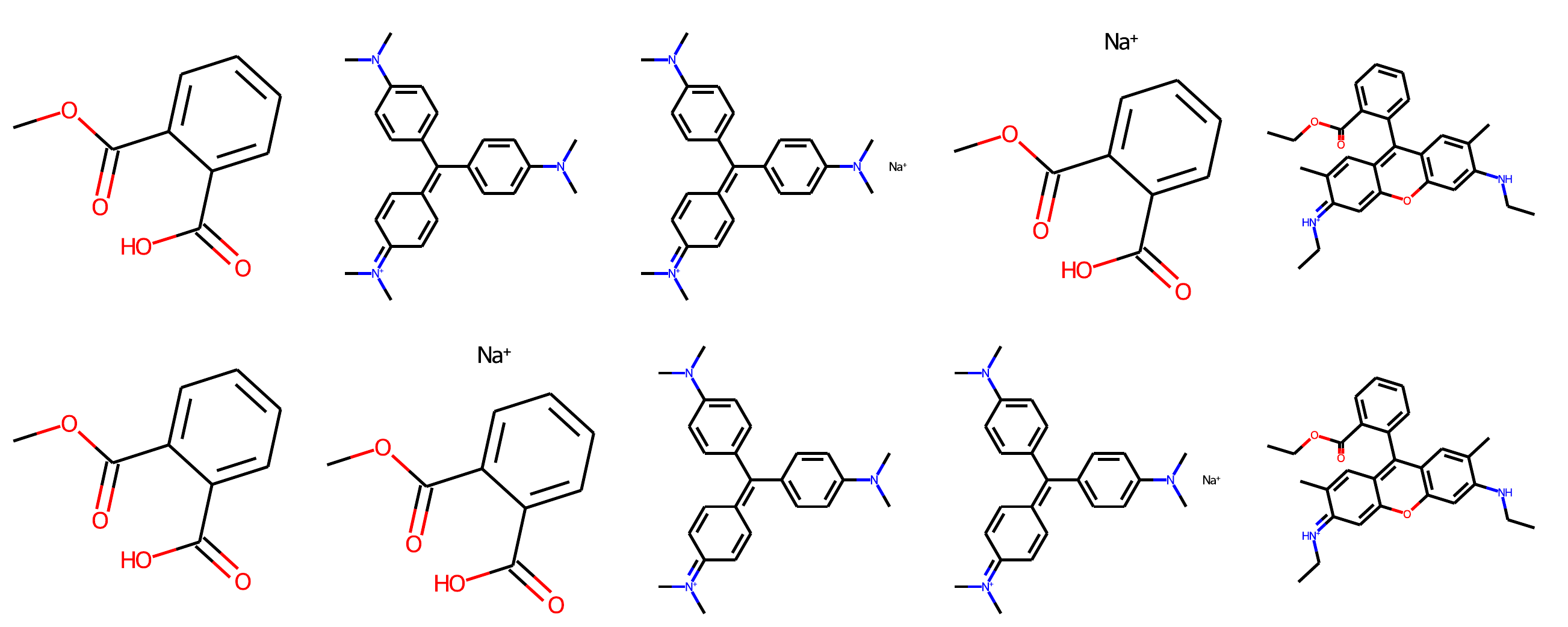}
\caption{Input: The molecule is a green dye.}
\end{figure*}

\begin{figure*}
\centering
\includegraphics[width=\textwidth]{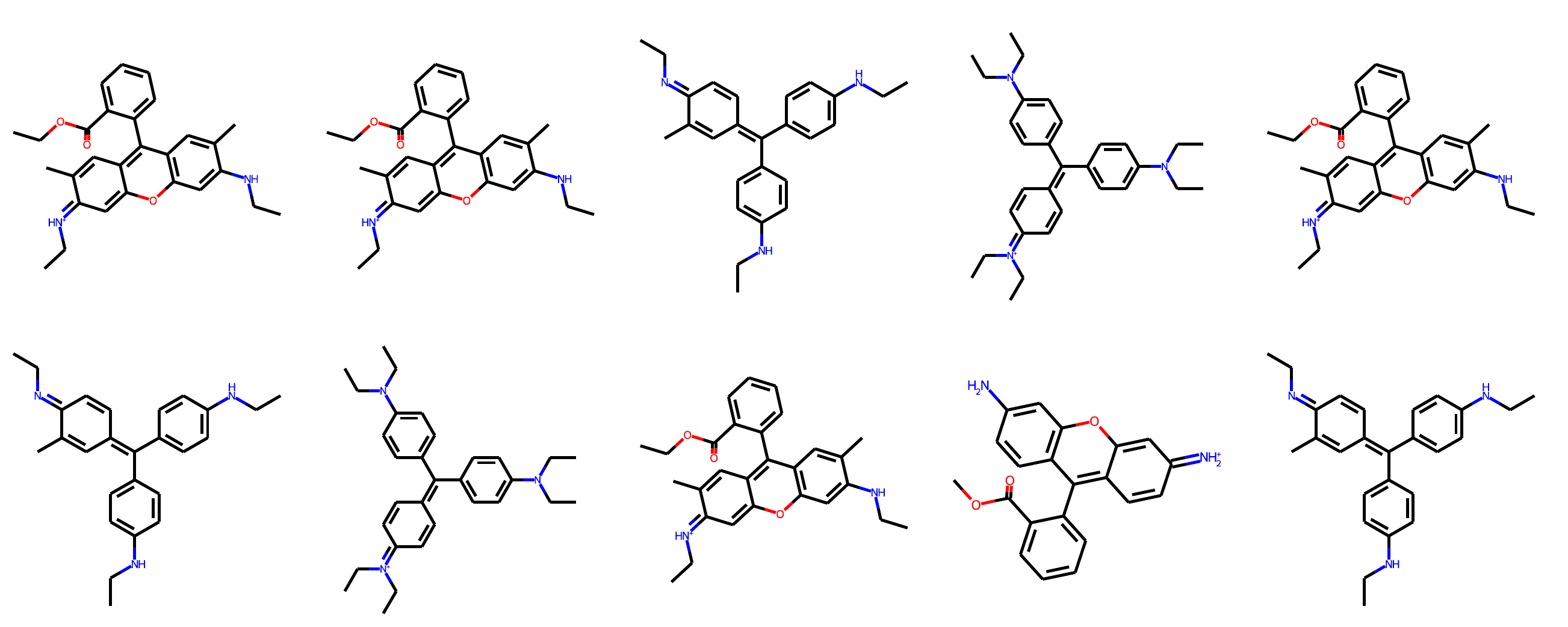}
\caption{Input: The molecule is a histological dye.}
\end{figure*}

\begin{figure*}
\centering
\includegraphics[width=\textwidth]{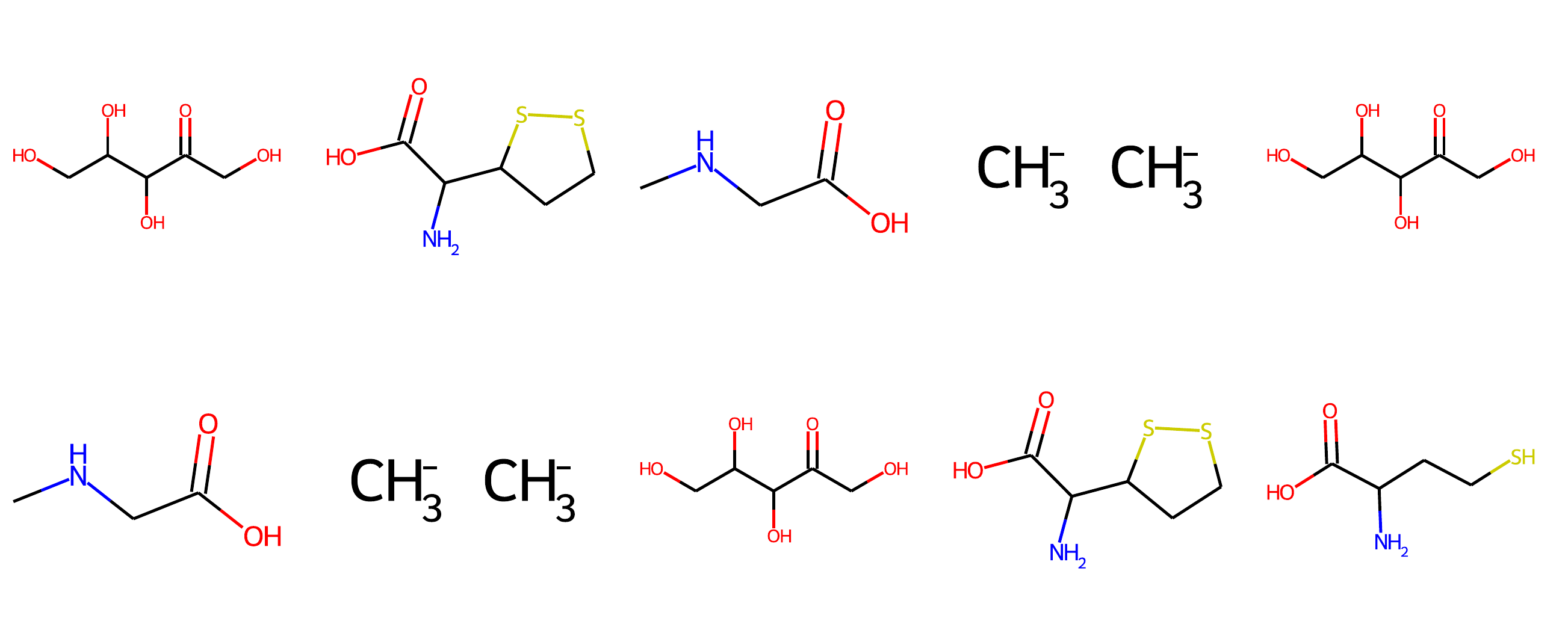}
\caption{Input: The molecule is a human metabolite.}
\end{figure*}

\begin{figure*}
\centering
\includegraphics[width=\textwidth]{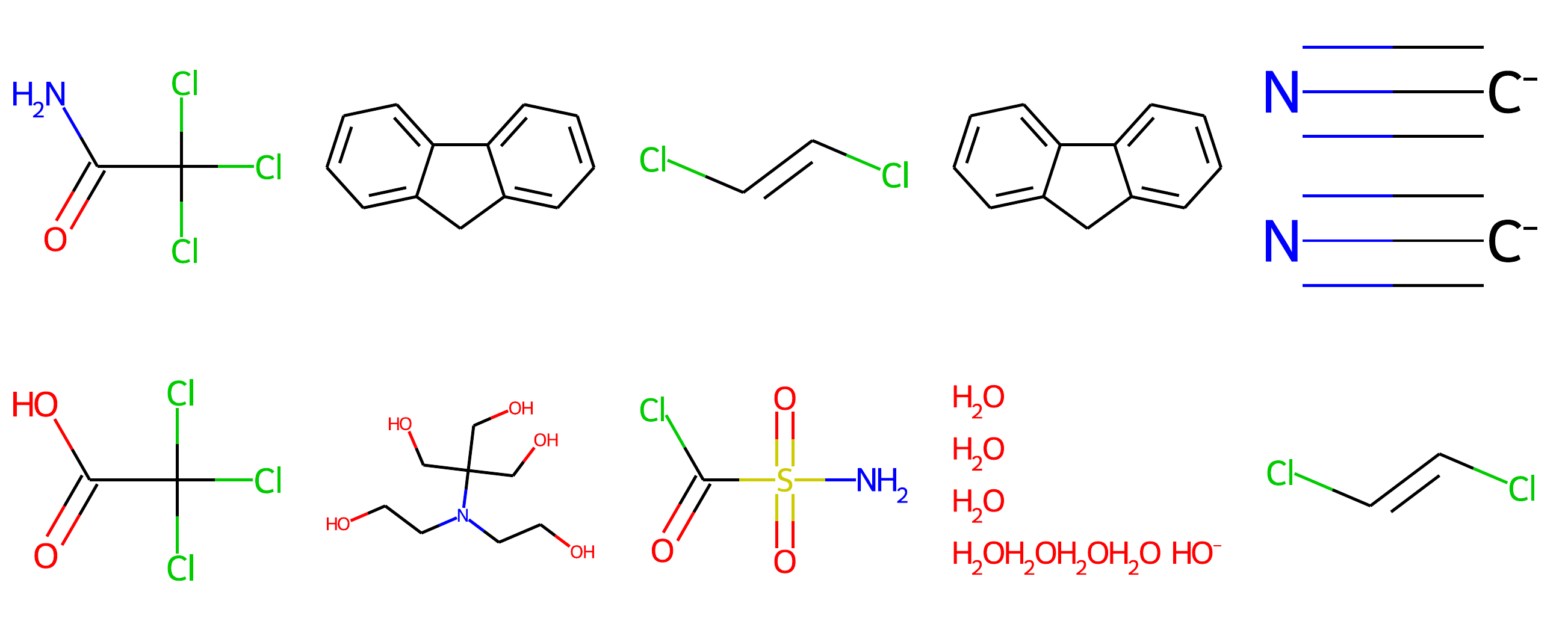}
\caption{Input: The molecule is a hydrocarbon which tastes really cool.}
\end{figure*}

\begin{figure*}
\centering
\includegraphics[width=\textwidth]{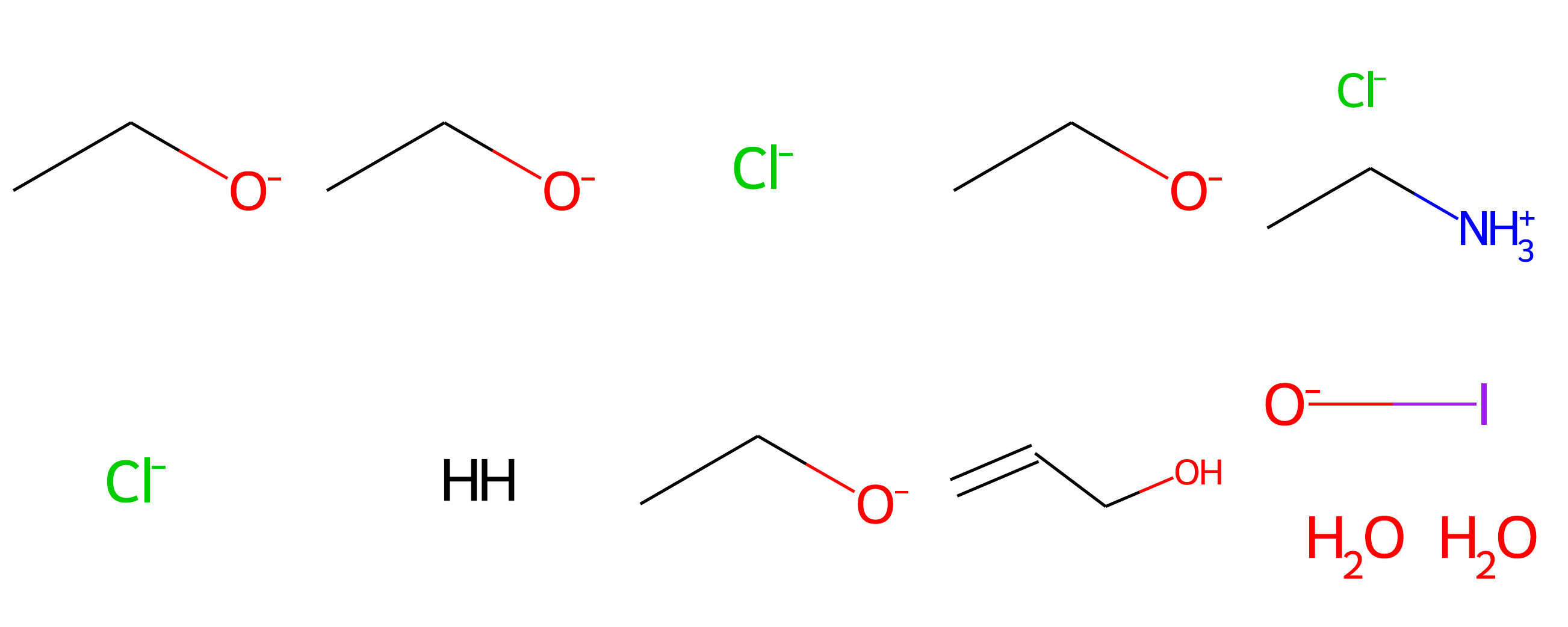}
\caption{Input: The molecule is a liquid at room temperature.}
\end{figure*}

\begin{figure*}
\centering
\includegraphics[width=\textwidth]{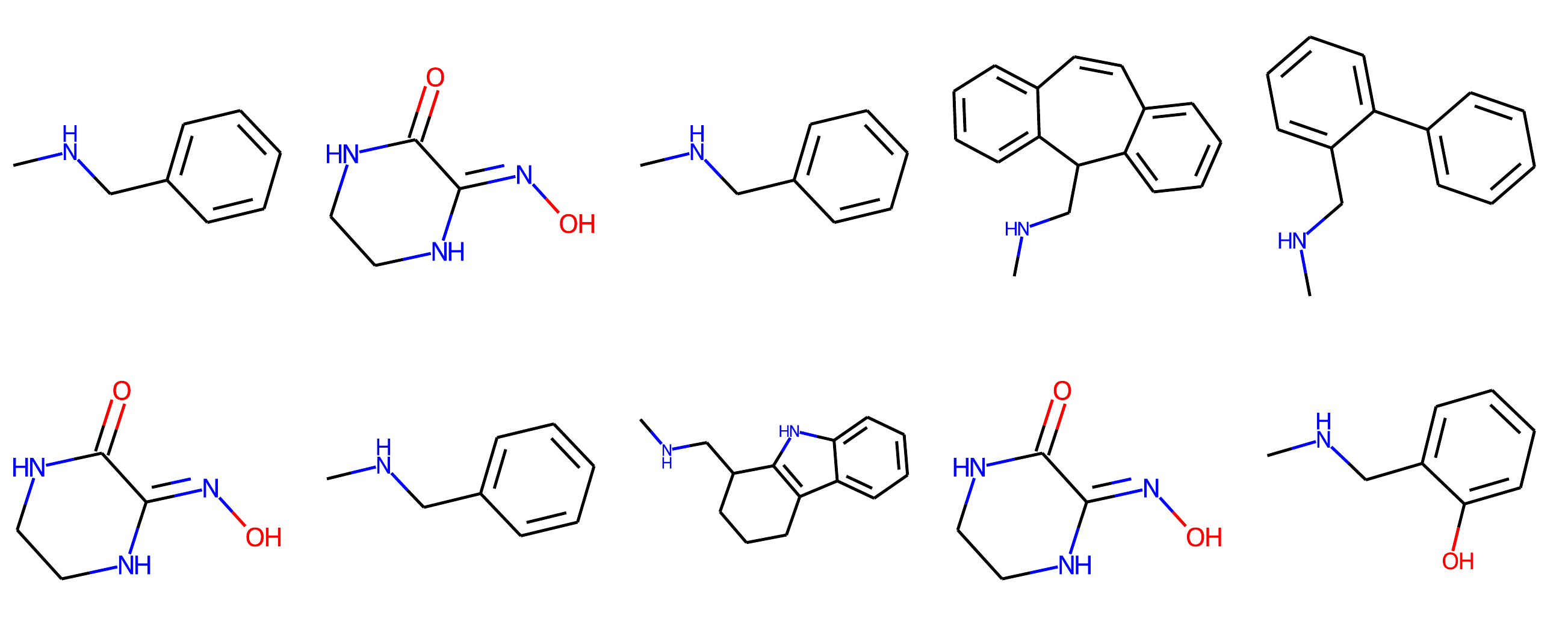}
\caption{Input: The molecule is a macrocycle.}
\end{figure*}

\begin{figure*}
\centering
\includegraphics[width=\textwidth]{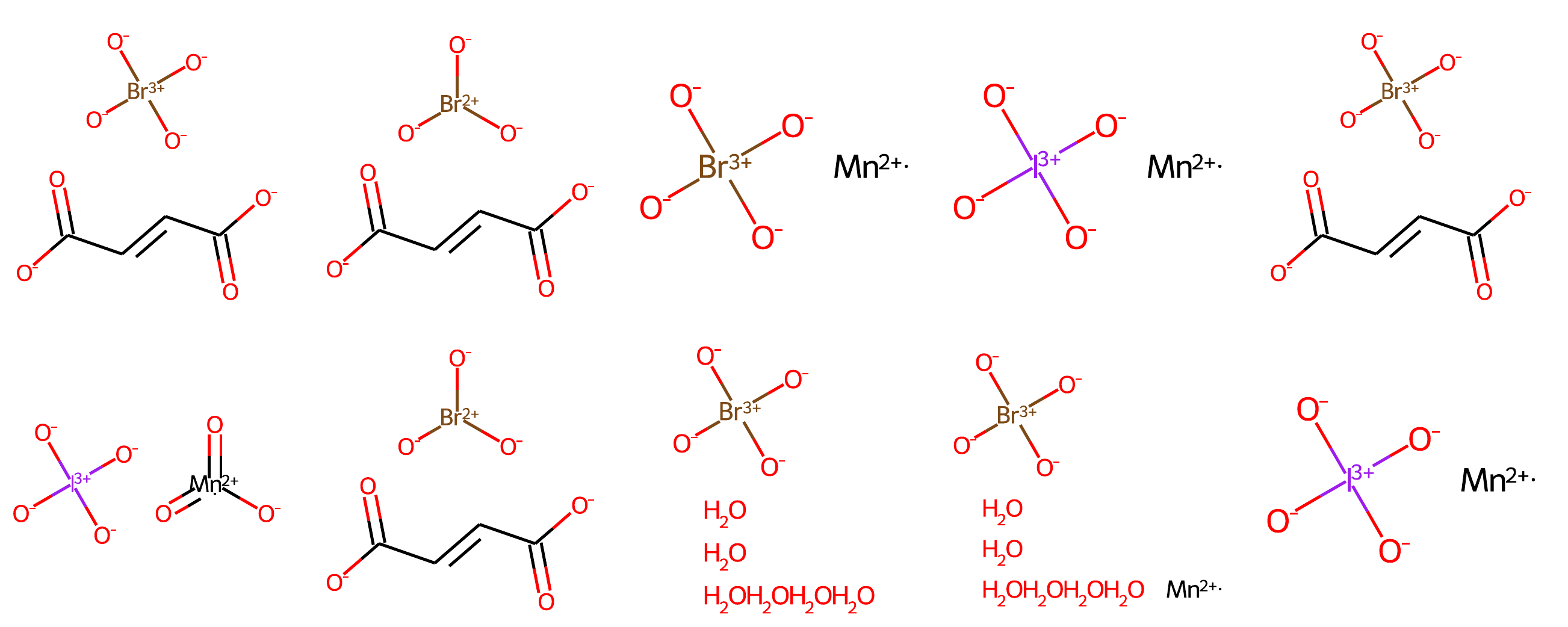}
\caption{Input: The molecule is a maleate salt.}
\end{figure*}

\begin{figure*}
\centering
\includegraphics[width=\textwidth]{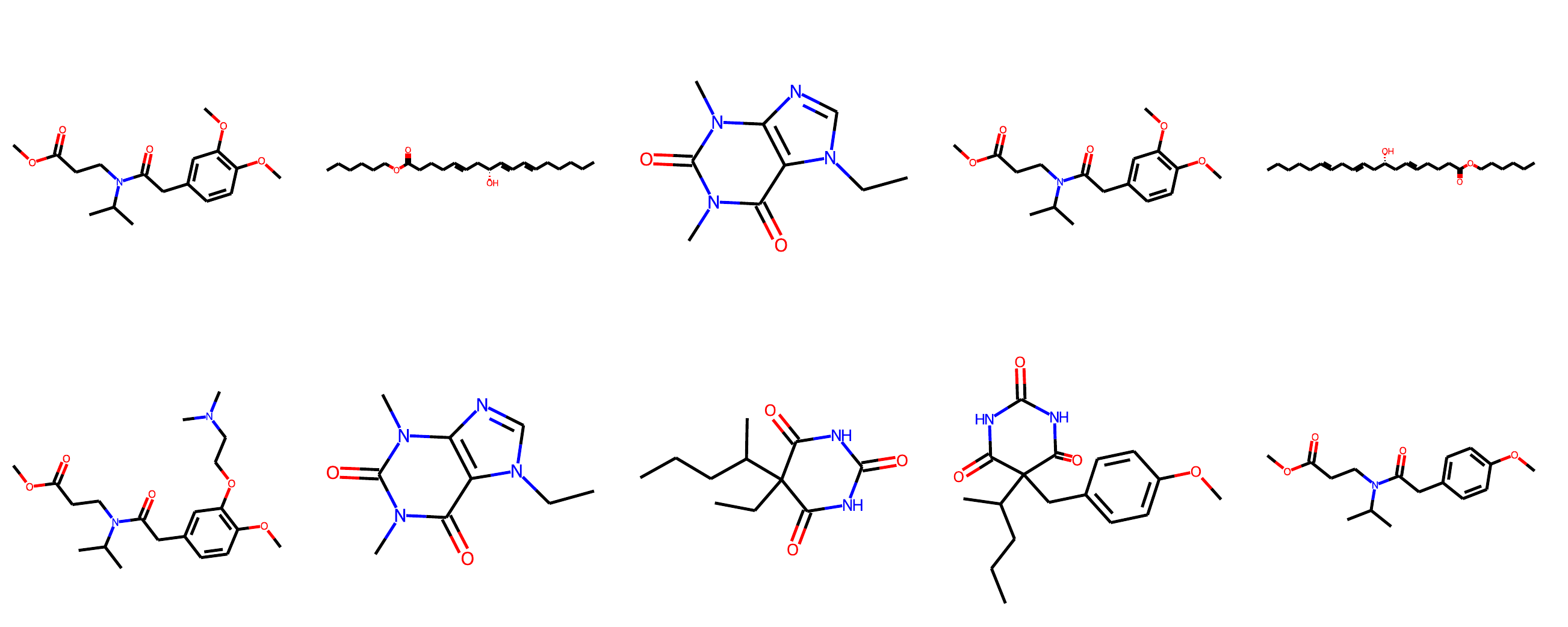}
\caption{Input: The molecule is a neurotransmitter agent.}
\end{figure*}

\begin{figure*}
\centering
\includegraphics[width=\textwidth]{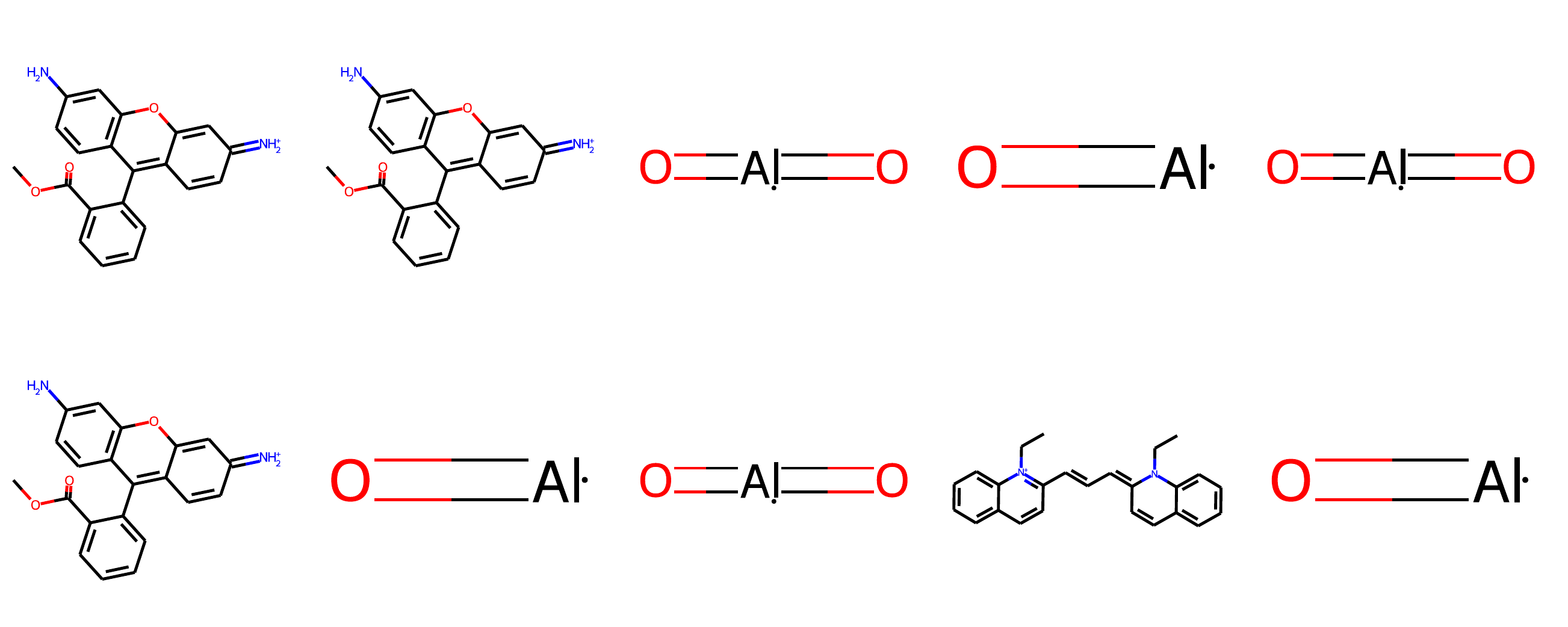}
\caption{Input: The molecule is a orange dye.}
\end{figure*}

\begin{figure*}
\centering
\includegraphics[width=\textwidth]{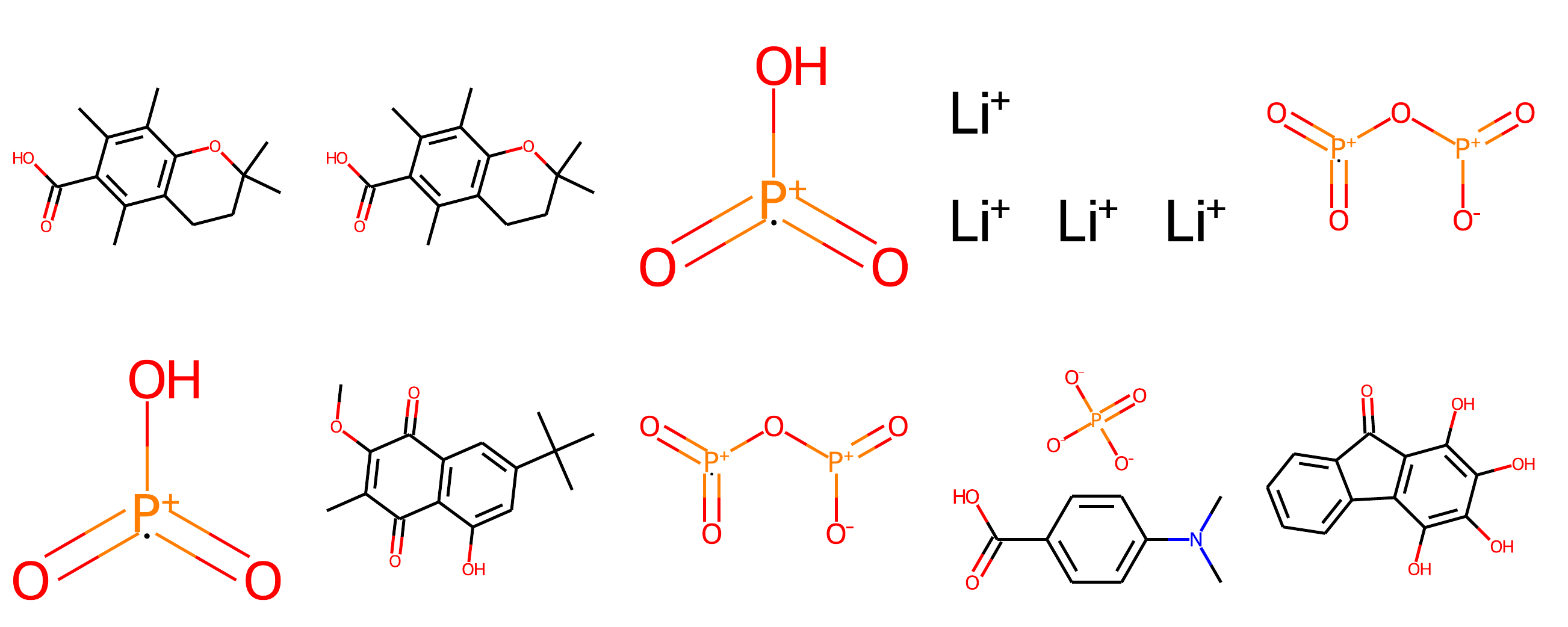}
\caption{Input: The molecule is a photovoltaic.}
\end{figure*}

\begin{figure*}
\centering
\includegraphics[width=\textwidth]{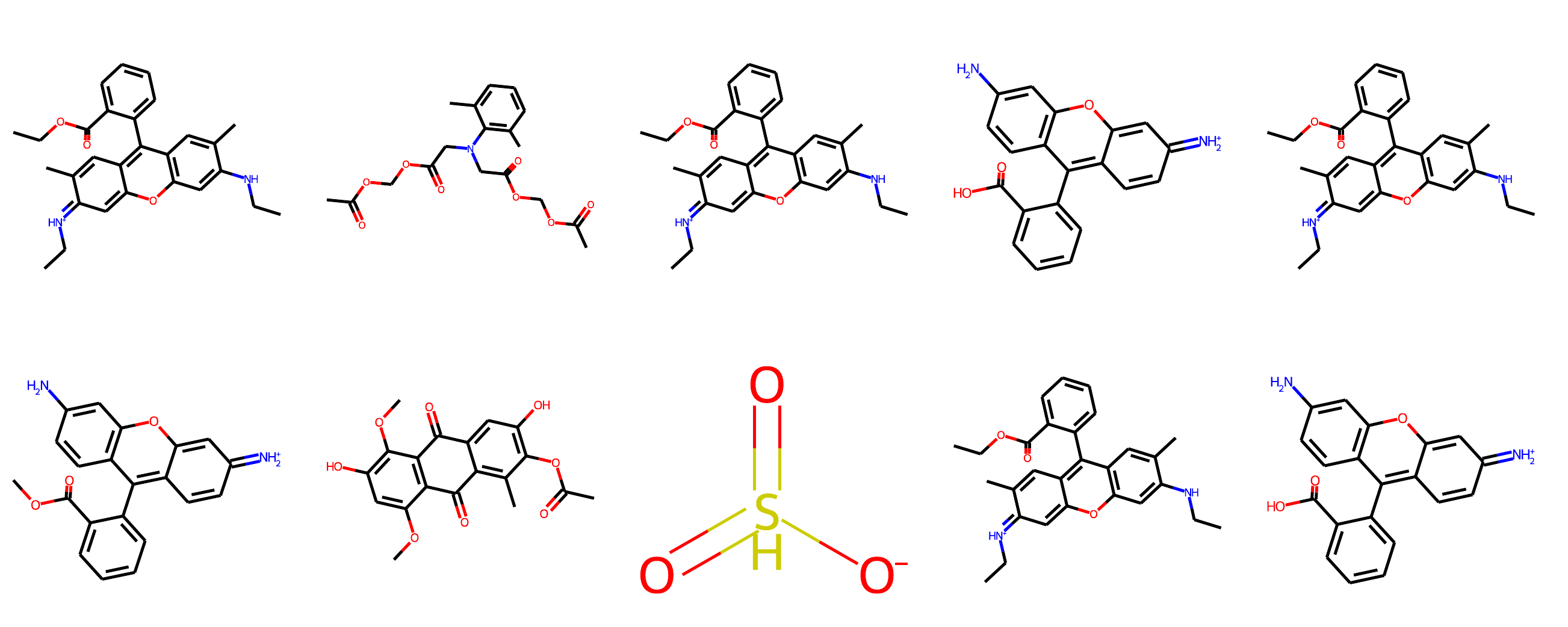}
\caption{Input: The molecule is a pigment which converts sunlight into energy.}
\end{figure*}

\begin{figure*}
\centering
\includegraphics[width=\textwidth]{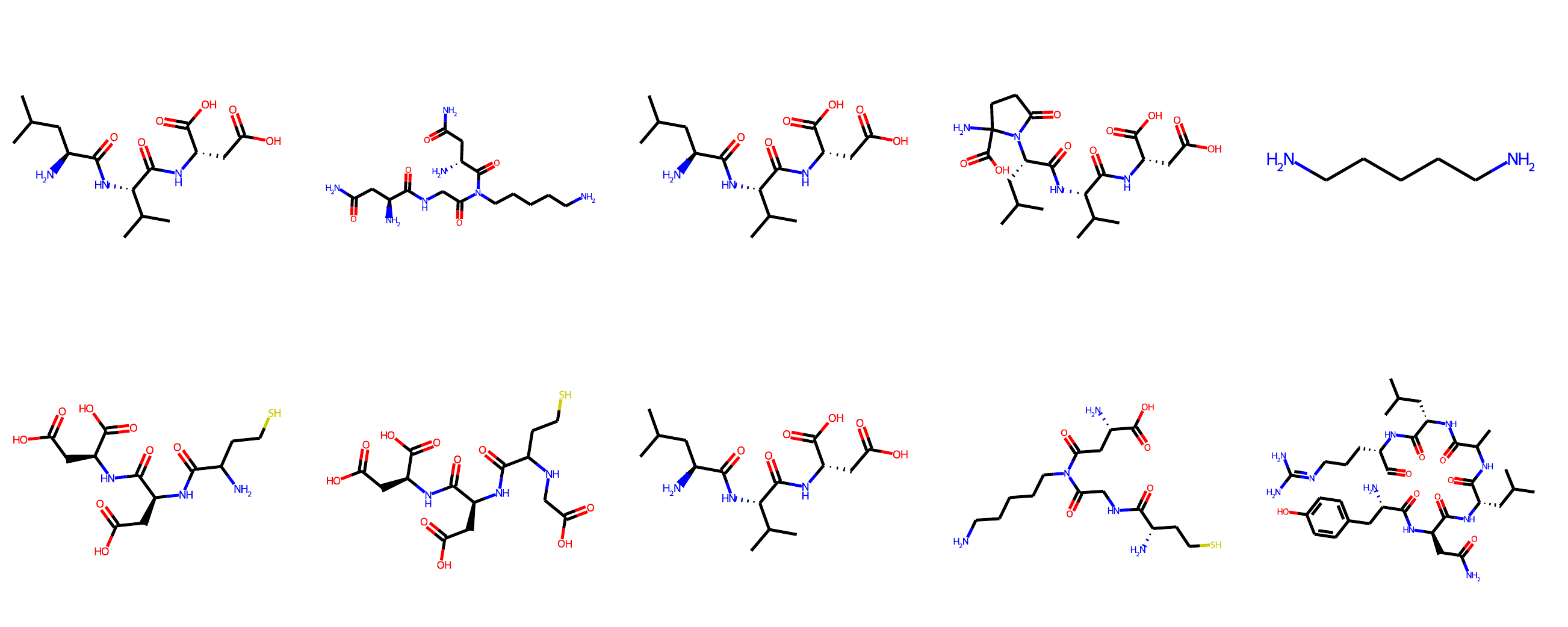}
\caption{Input: The molecule is a polypeptide.}
\end{figure*}

\begin{figure*}
\centering
\includegraphics[width=\textwidth]{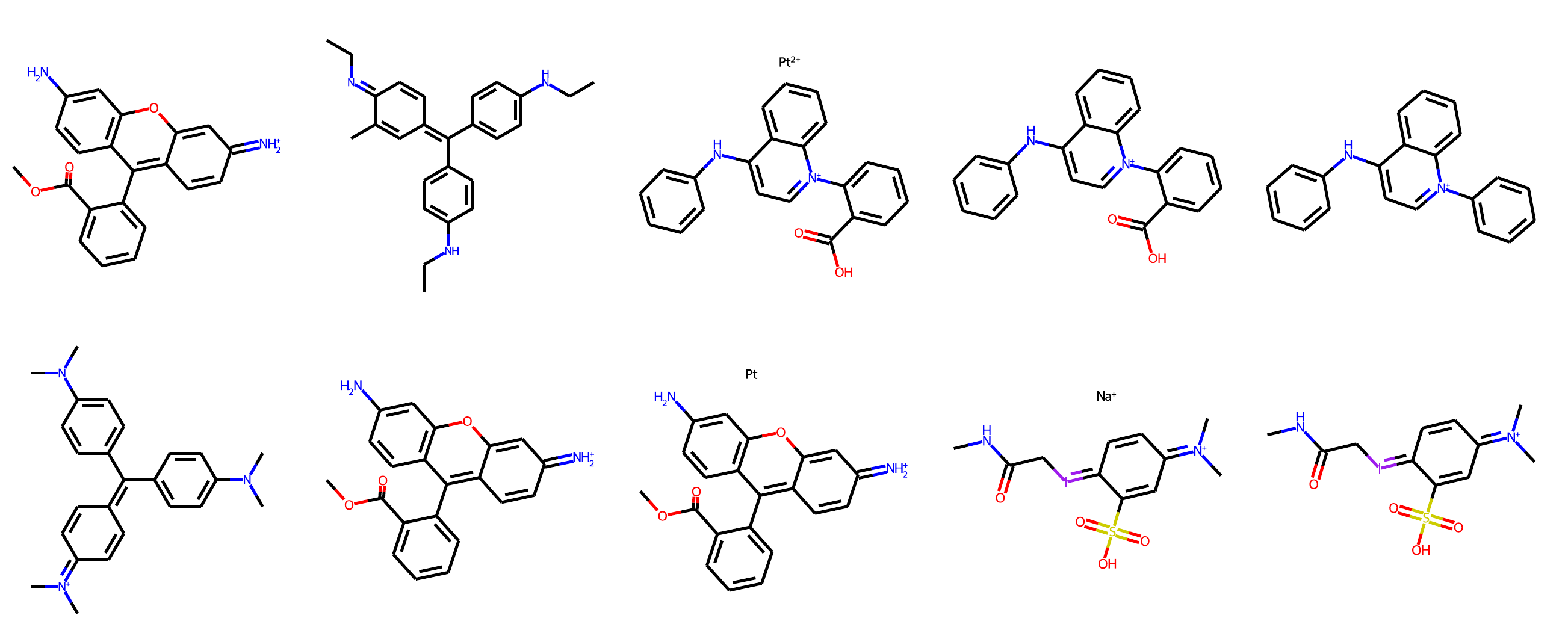}
\caption{Input: The molecule is a purple dye.}
\end{figure*}

\begin{figure*}
\centering
\includegraphics[width=\textwidth]{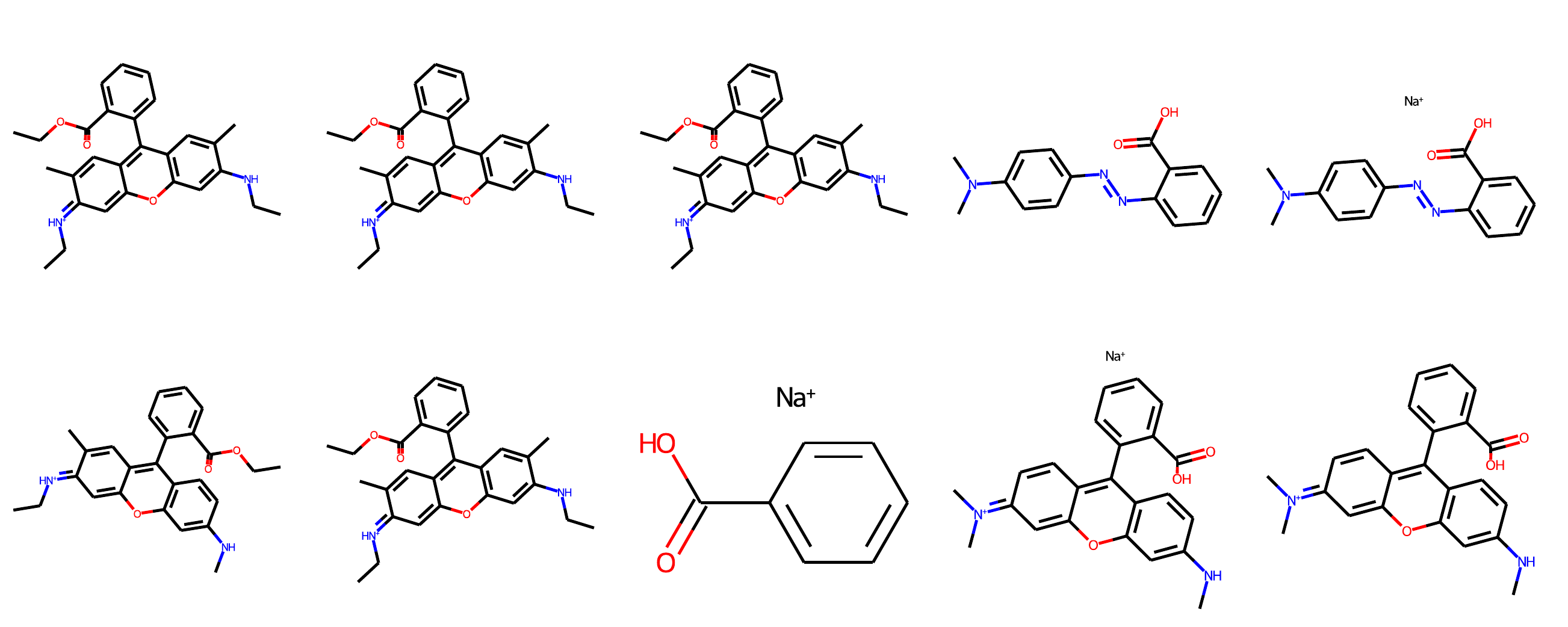}
\caption{Input: The molecule is a red dye.}
\end{figure*}

\begin{figure*}
\centering
\includegraphics[width=\textwidth]{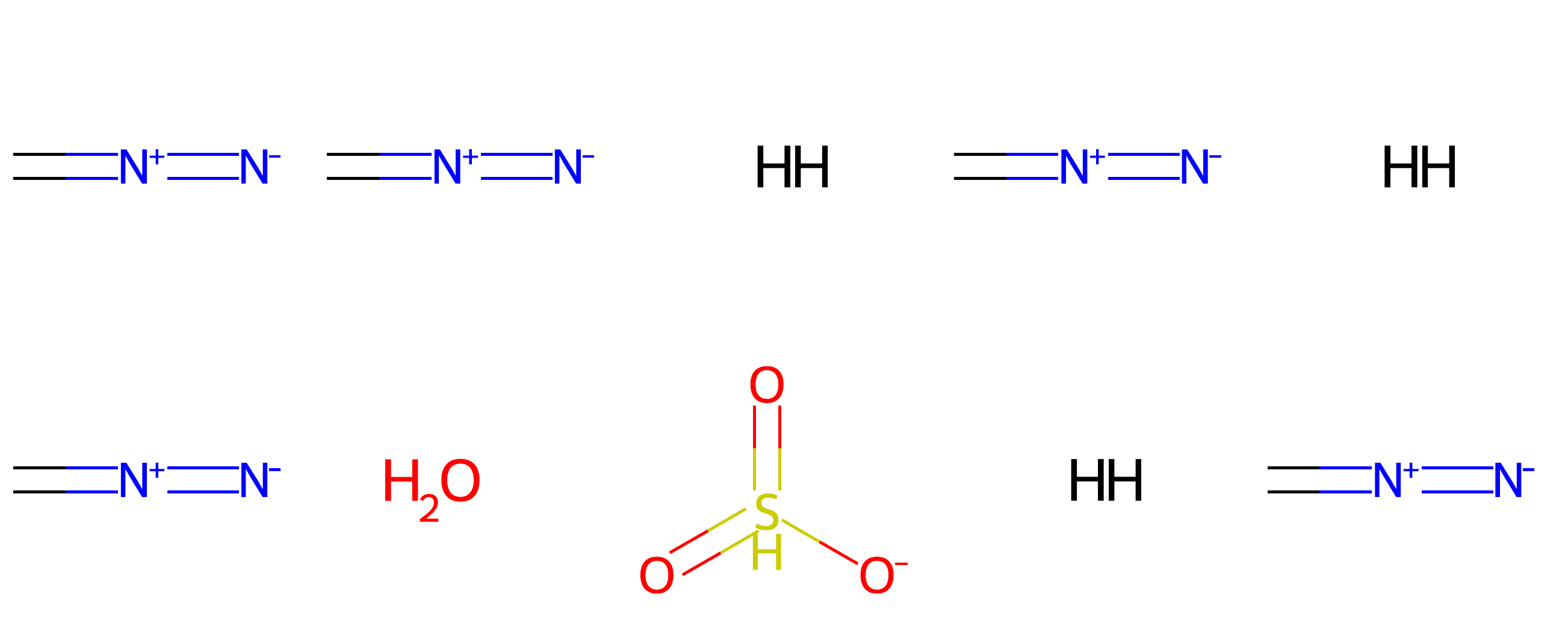}
\caption{Input: The molecule is a solid at room temperature.}
\end{figure*}

\begin{figure*}
\centering
\includegraphics[width=\textwidth]{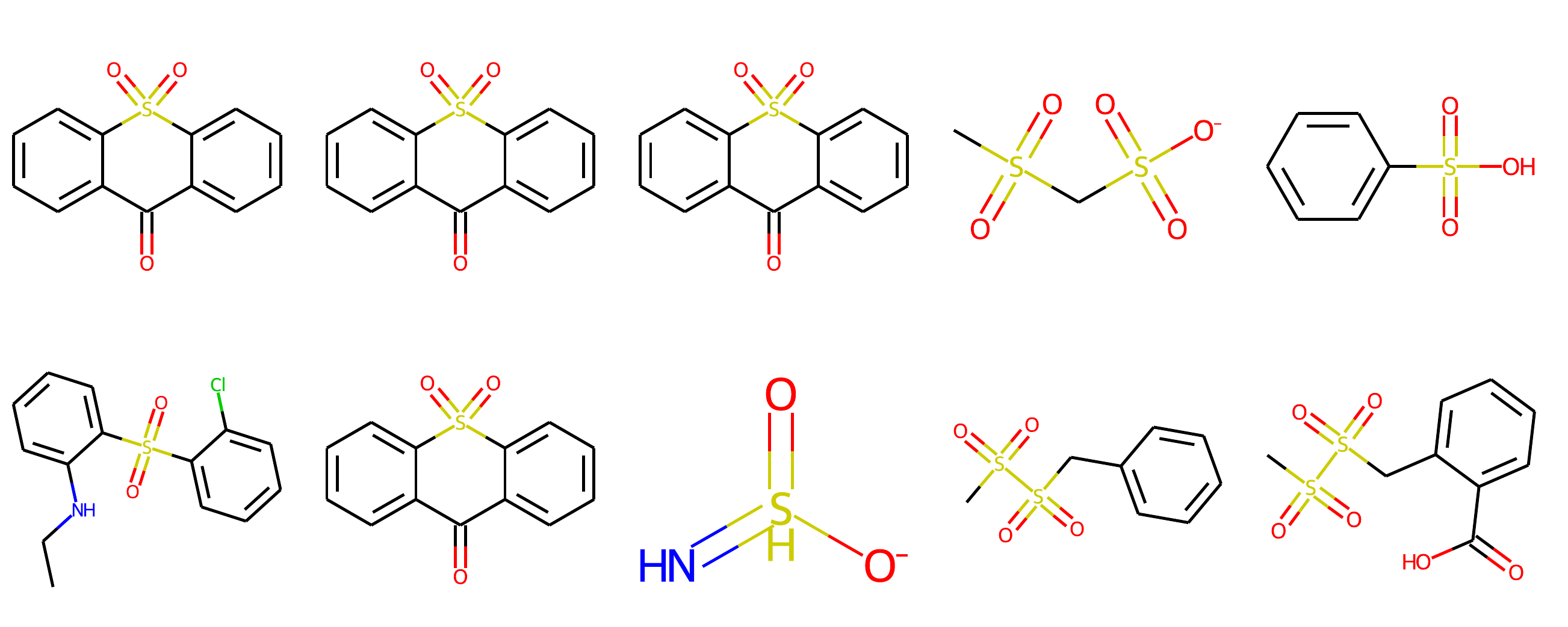}
\caption{Input: The molecule is a sulfonated xanthene.}
\end{figure*}

\begin{figure*}
\centering
\includegraphics[width=\textwidth]{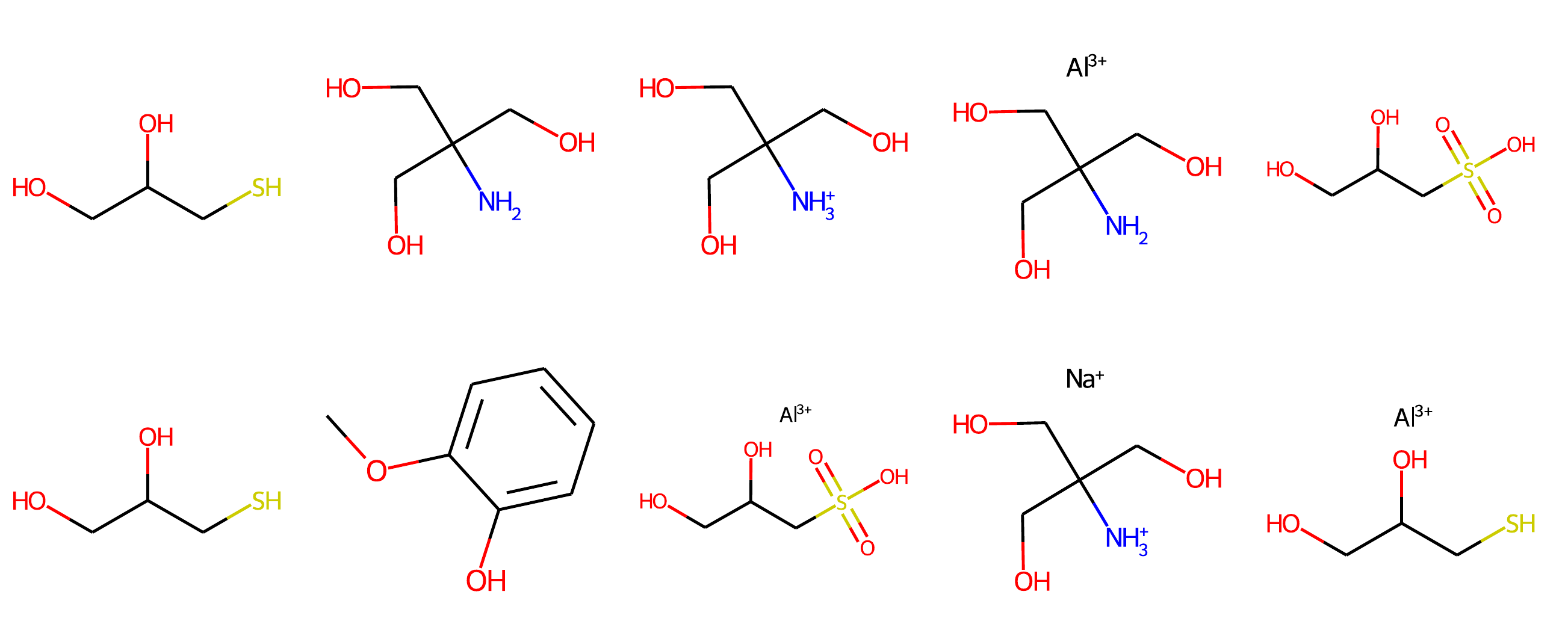}
\caption{Input: The molecule is a sweet tasting sugar additive.}
\end{figure*}

\begin{figure*}
\centering
\includegraphics[width=\textwidth]{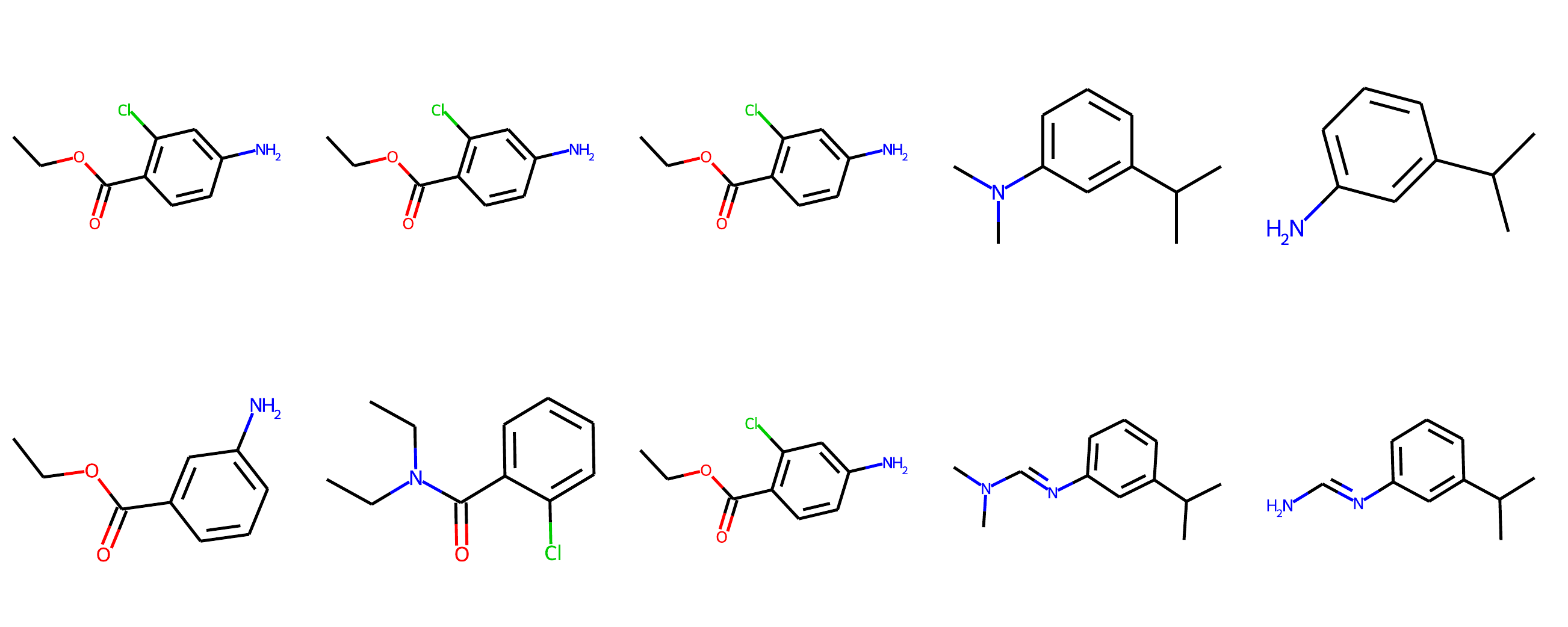}
\caption{Input: The molecule is a topical anaesthetic.}
\end{figure*}

\begin{figure*}
\centering
\includegraphics[width=\textwidth]{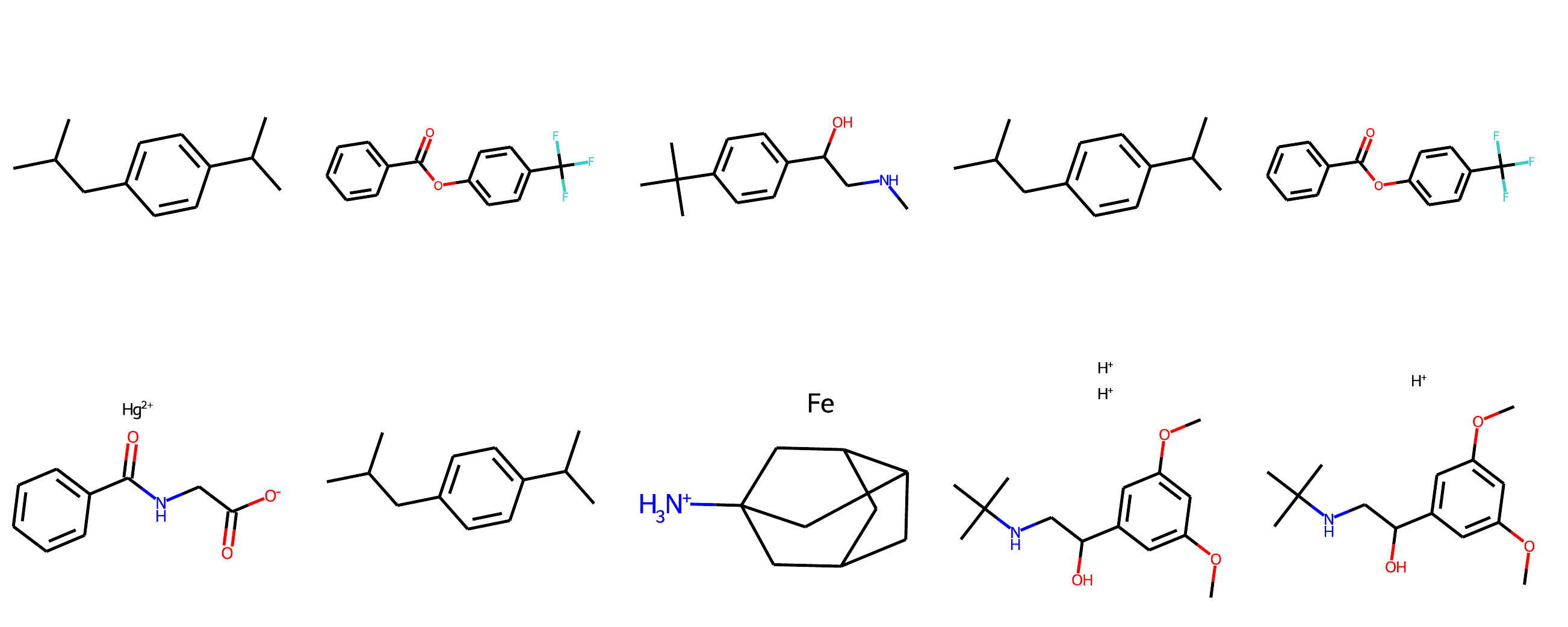}
\caption{Input: The molecule is able to lower blood pressure.}
\end{figure*}

\begin{figure*}
\centering
\includegraphics[width=\textwidth]{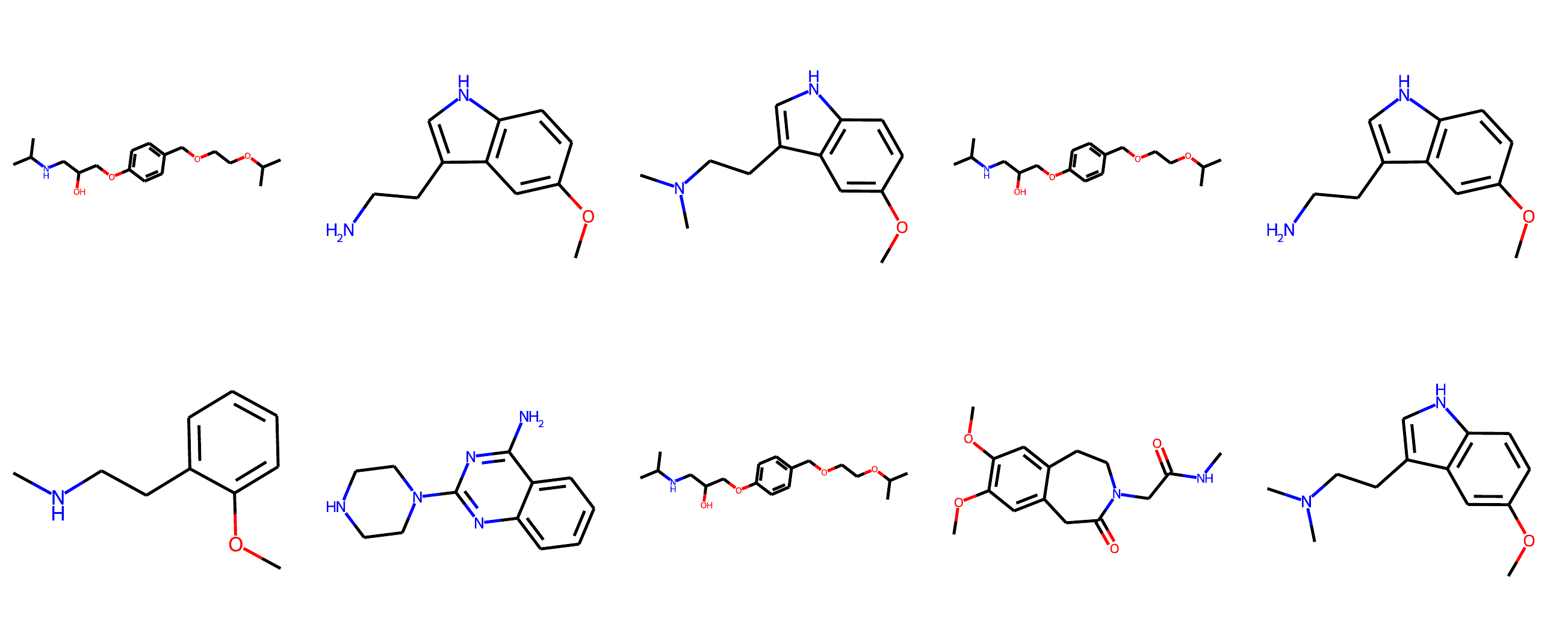}
\caption{Input: The molecule is an adrenergic uptake inhibitor.}
\end{figure*}

\begin{figure*}
\centering
\includegraphics[width=\textwidth]{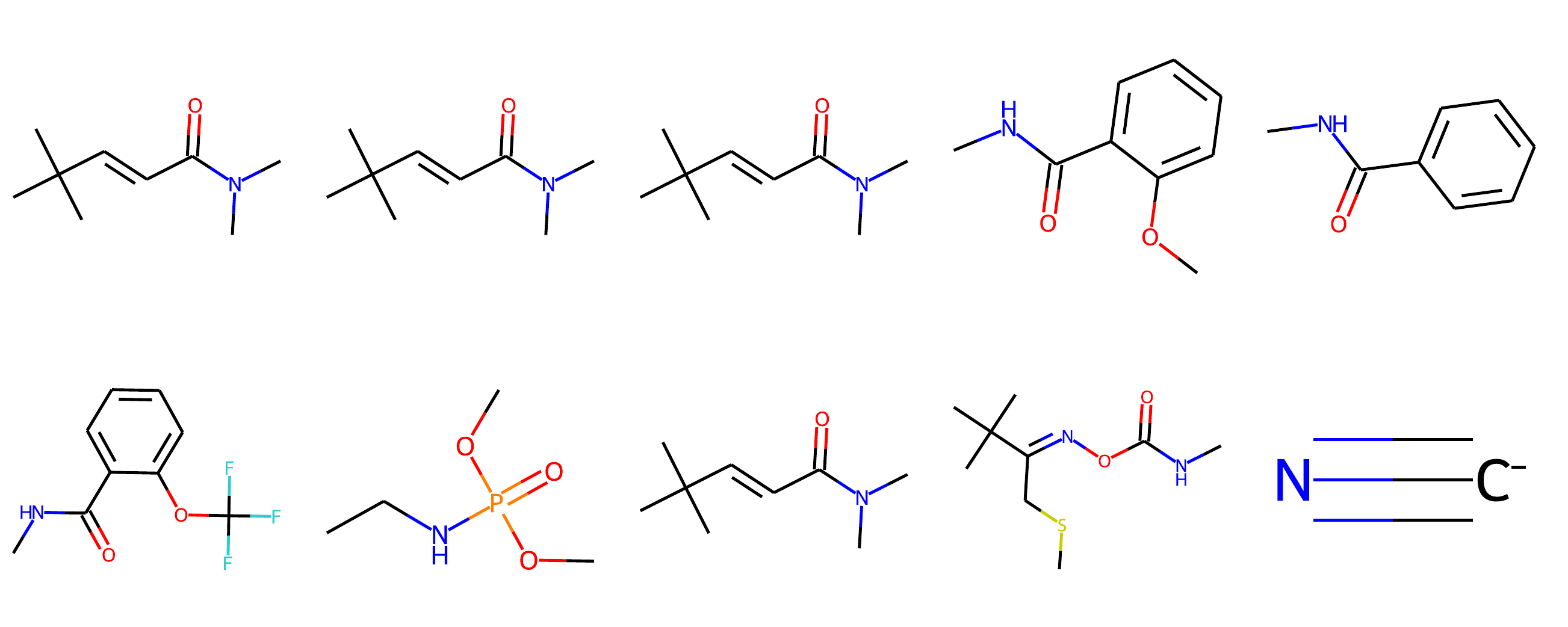}
\caption{Input: The molecule is an agrochemical.}
\end{figure*}

\begin{figure*}
\centering
\includegraphics[width=\textwidth]{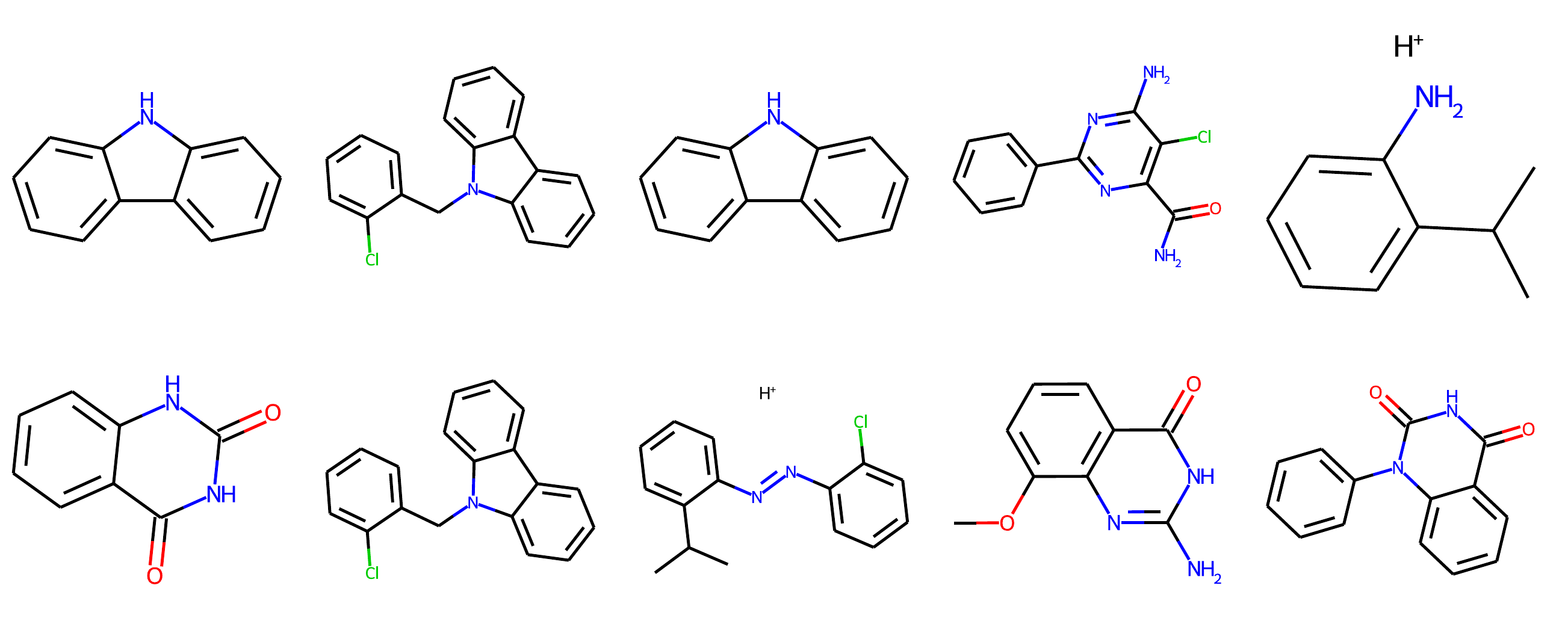}
\caption{Input: The molecule is an anabolic agent.}
\end{figure*}

\begin{figure*}
\centering
\includegraphics[width=\textwidth]{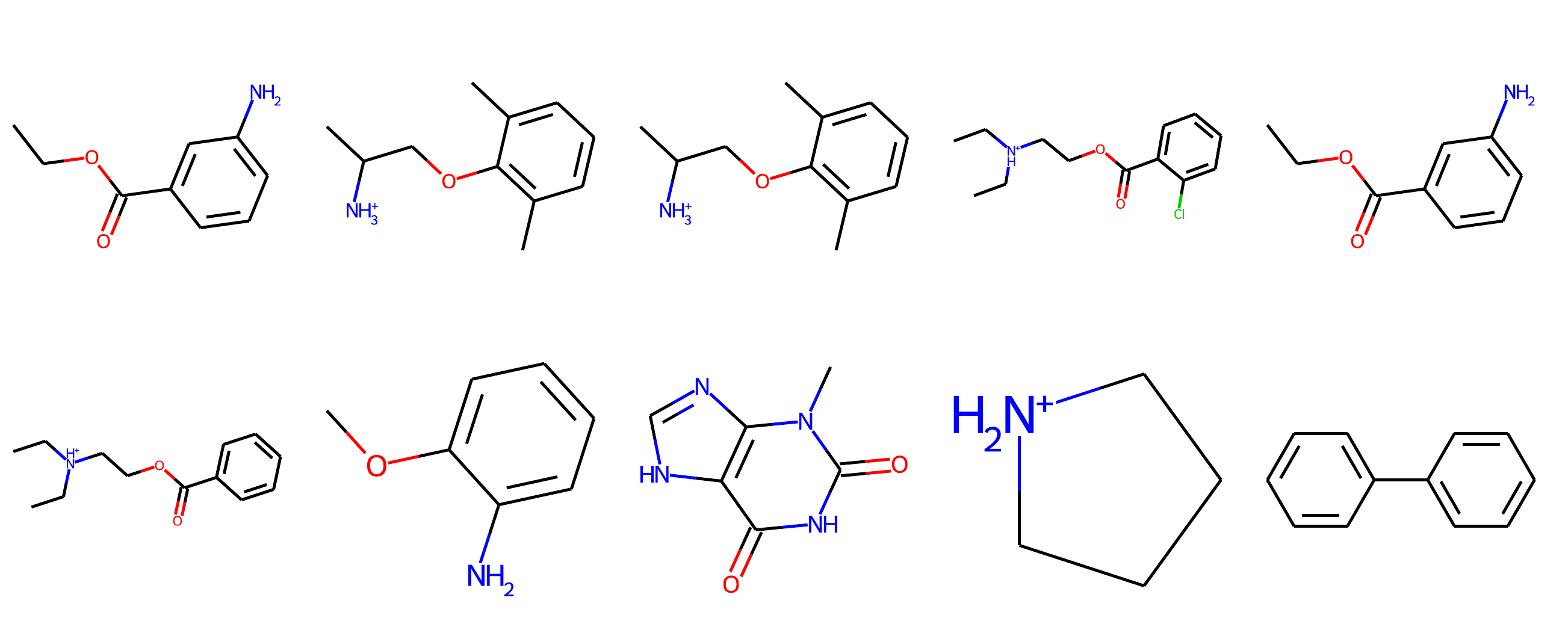}
\caption{Input: The molecule is an analgesic.}
\end{figure*}

\begin{figure*}
\centering
\includegraphics[width=\textwidth]{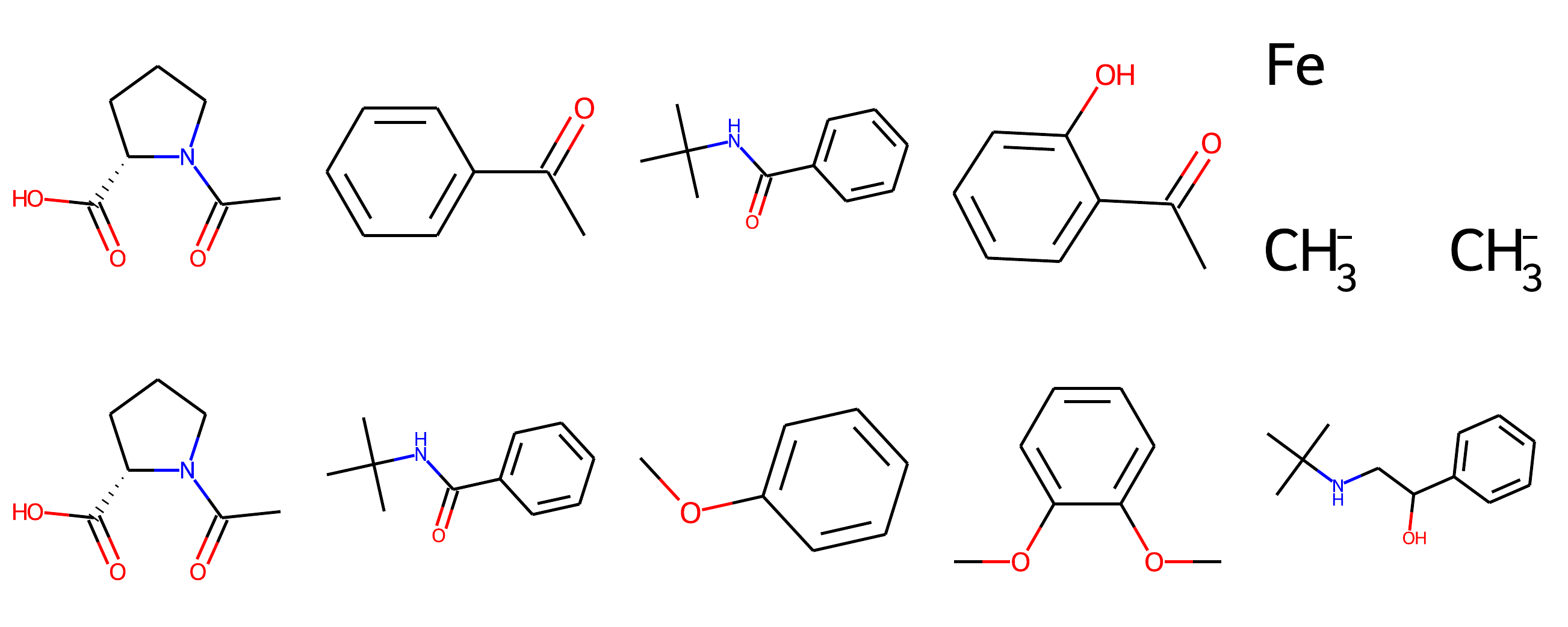}
\caption{Input: The molecule is an angry man.}
\end{figure*}

\begin{figure*}
\centering
\includegraphics[width=\textwidth]{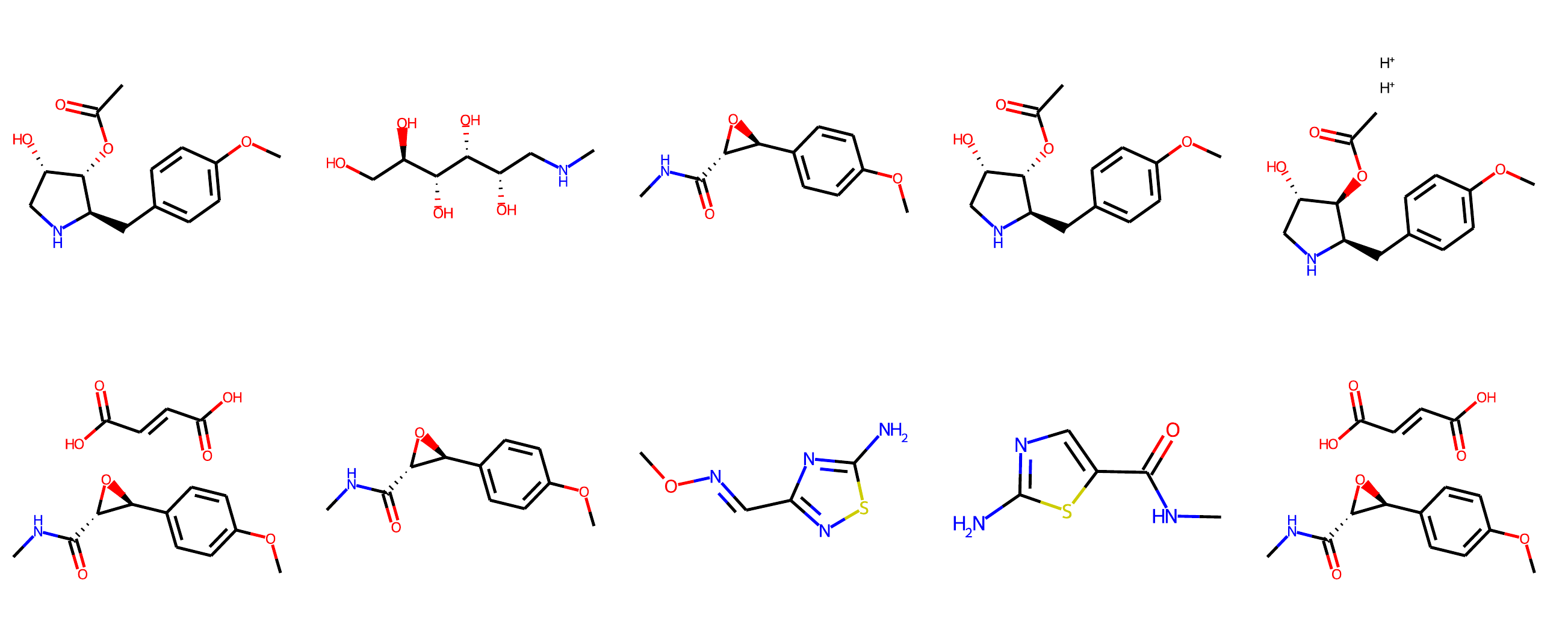}
\caption{Input: The molecule is an antibiotic.}
\end{figure*}

\begin{figure*}
\centering
\includegraphics[width=\textwidth]{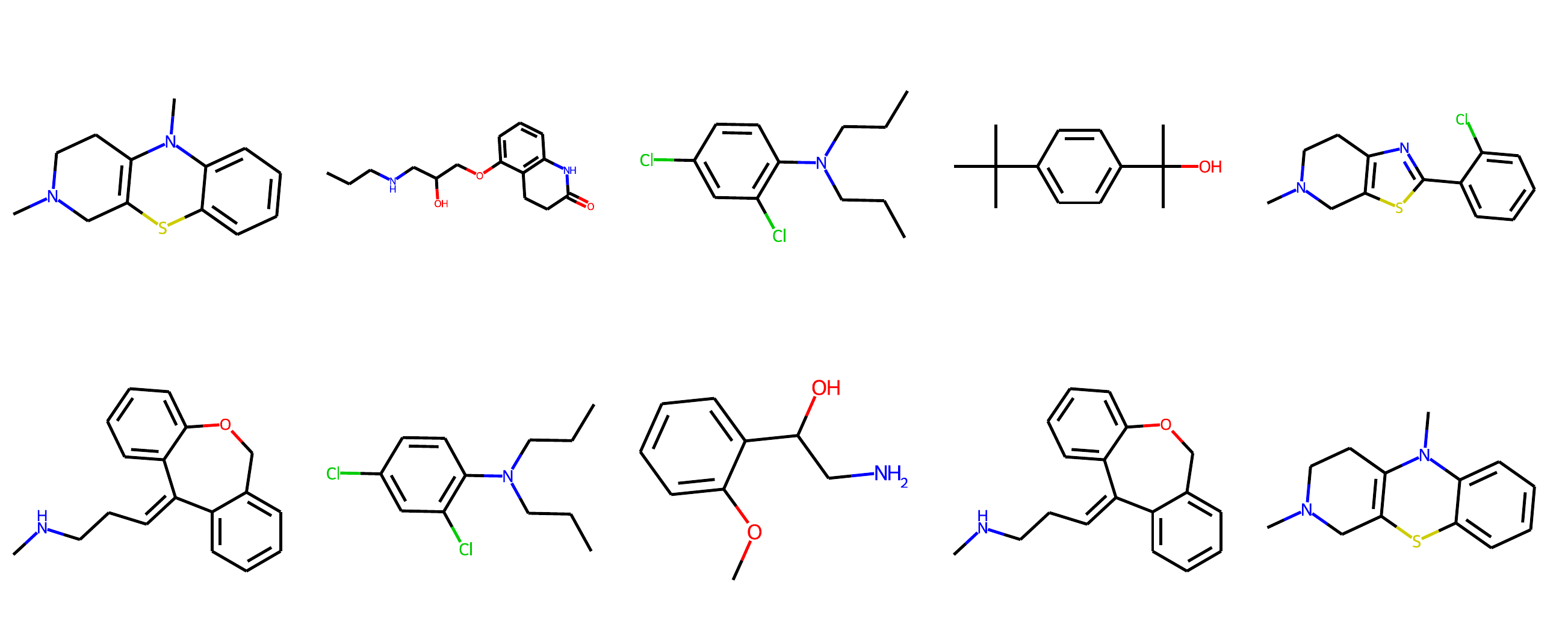}
\caption{Input: The molecule is an antidepressant.}
\end{figure*}

\begin{figure*}
\centering
\includegraphics[width=\textwidth]{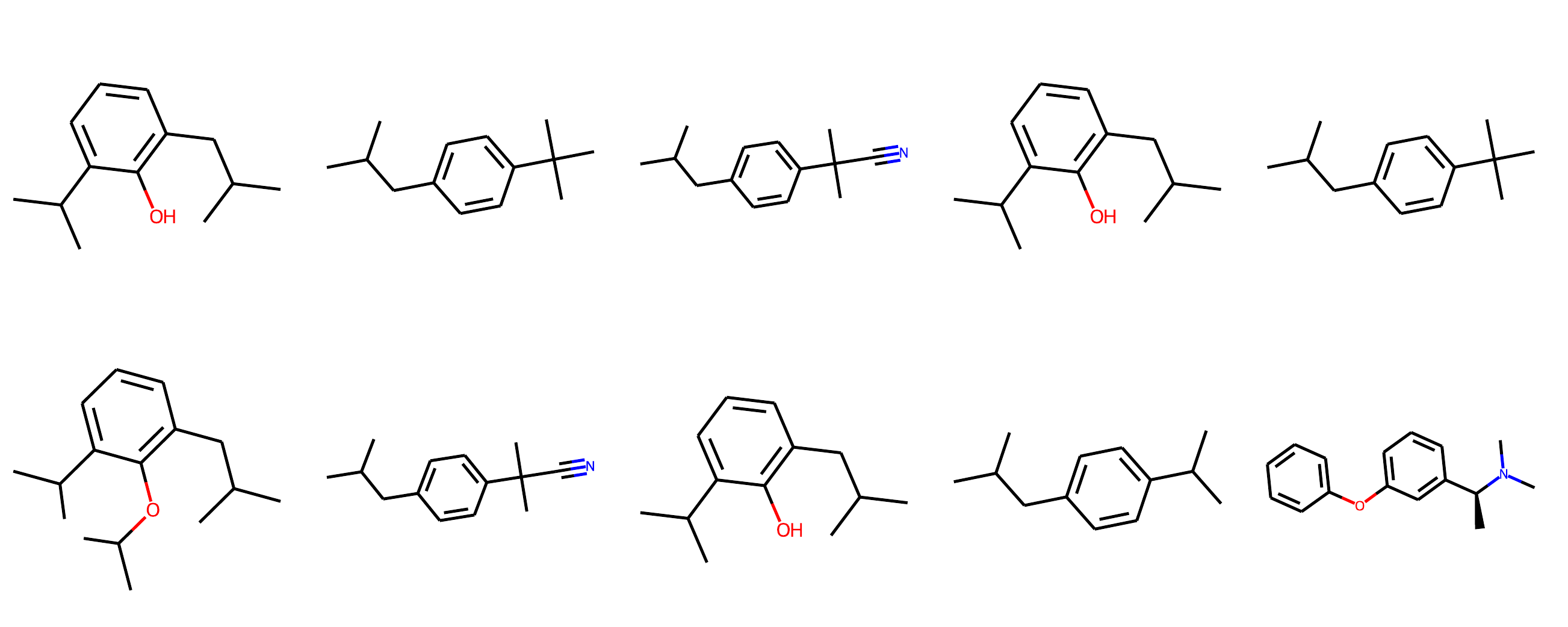}
\caption{Input: The molecule is an anti-inflammatory agent.}
\end{figure*}

\begin{figure*}
\centering
\includegraphics[width=\textwidth]{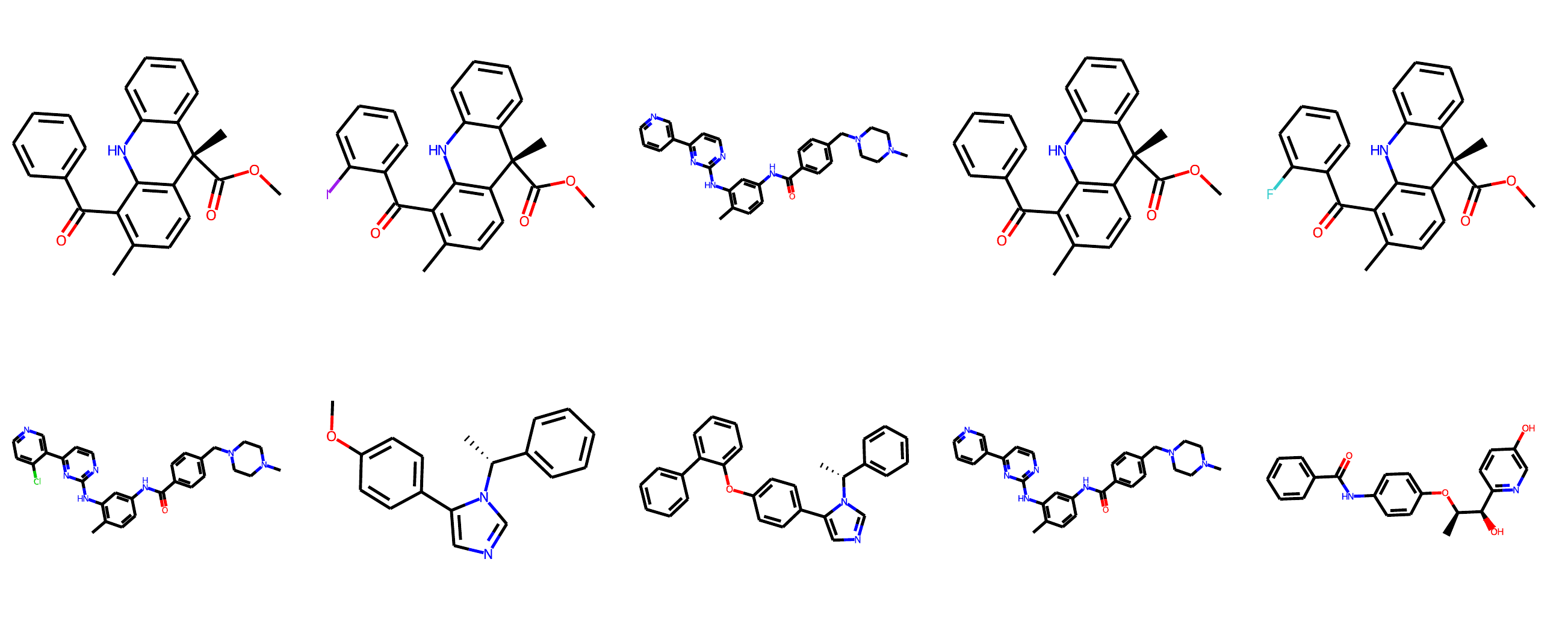}
\caption{Input: The molecule is an antineoplastic agent.}
\end{figure*}

\begin{figure*}
\centering
\includegraphics[width=\textwidth]{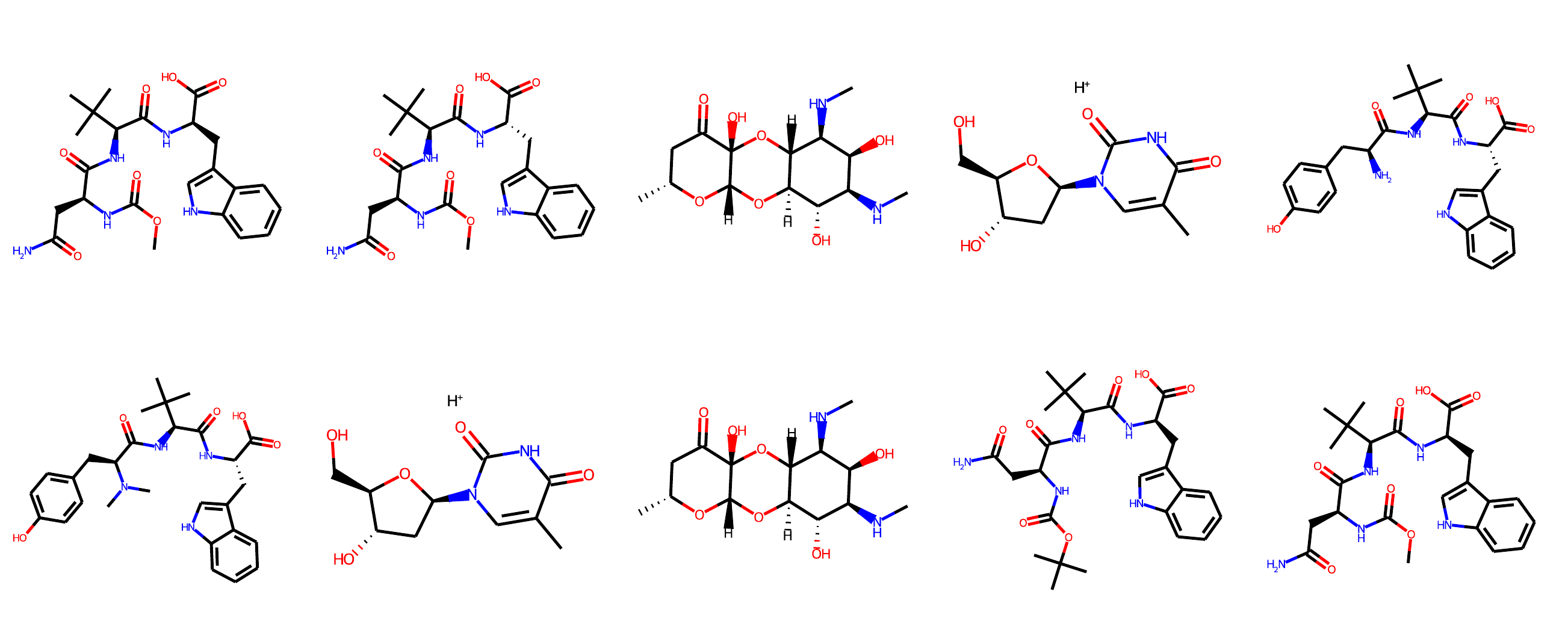}
\caption{Input: The molecule is an antiplasmodial drug.}
\end{figure*}

\begin{figure*}
\centering
\includegraphics[width=\textwidth]{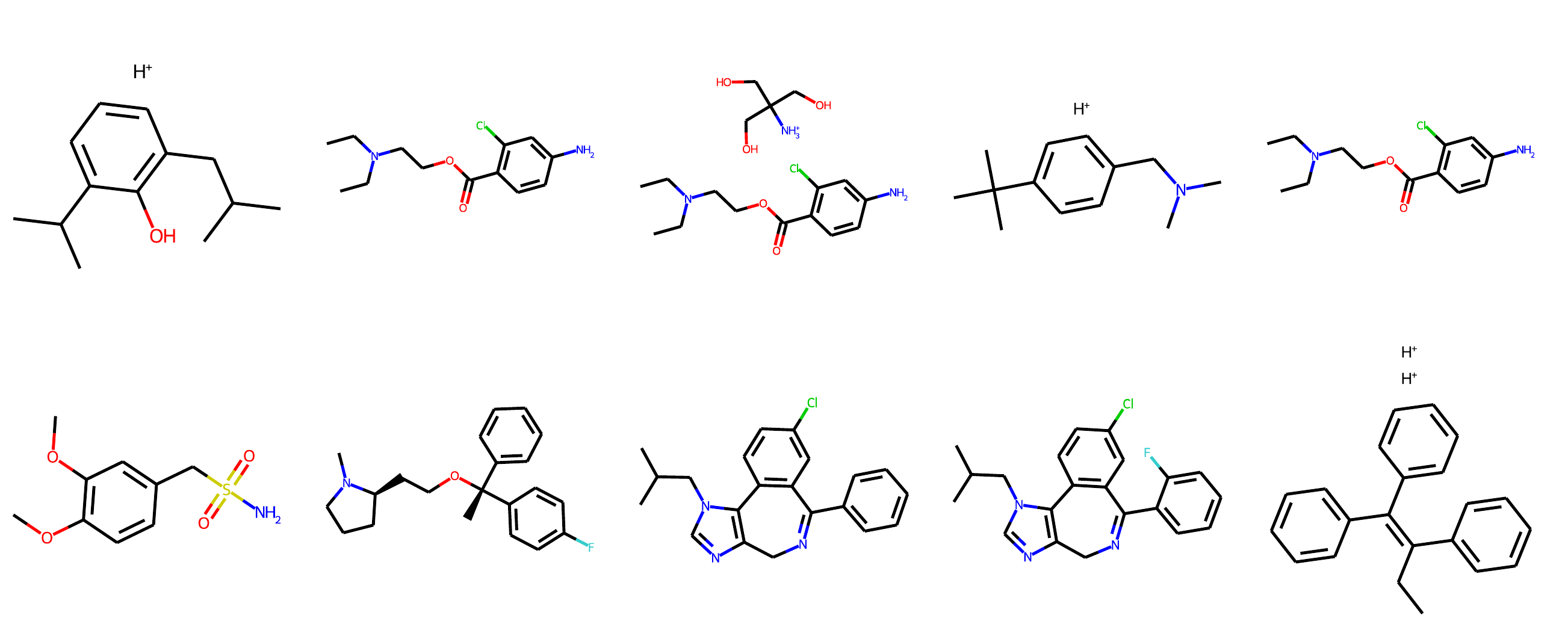}
\caption{Input: The molecule is an antipruritic drug.}
\end{figure*}

\begin{figure*}
\centering
\includegraphics[width=\textwidth]{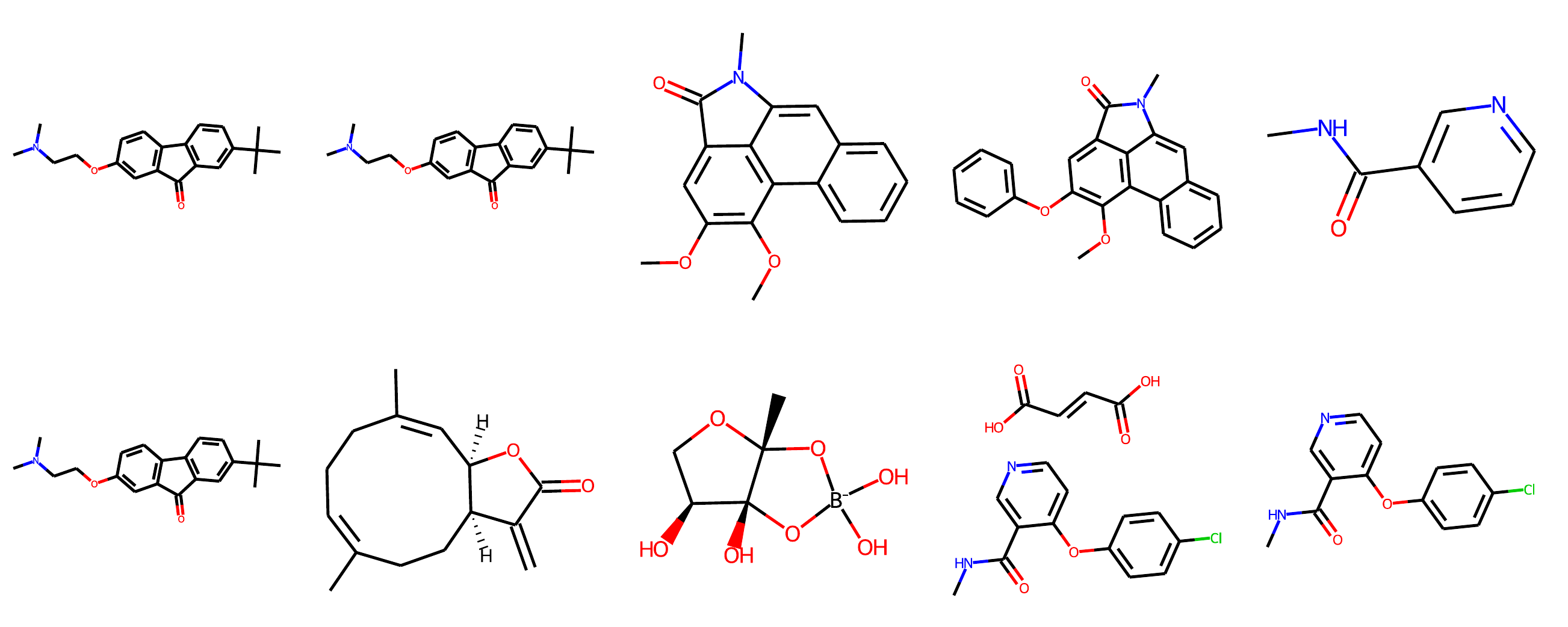}
\caption{Input: The molecule is an antitubercular agent.}
\end{figure*}

\begin{figure*}
\centering
\includegraphics[width=\textwidth]{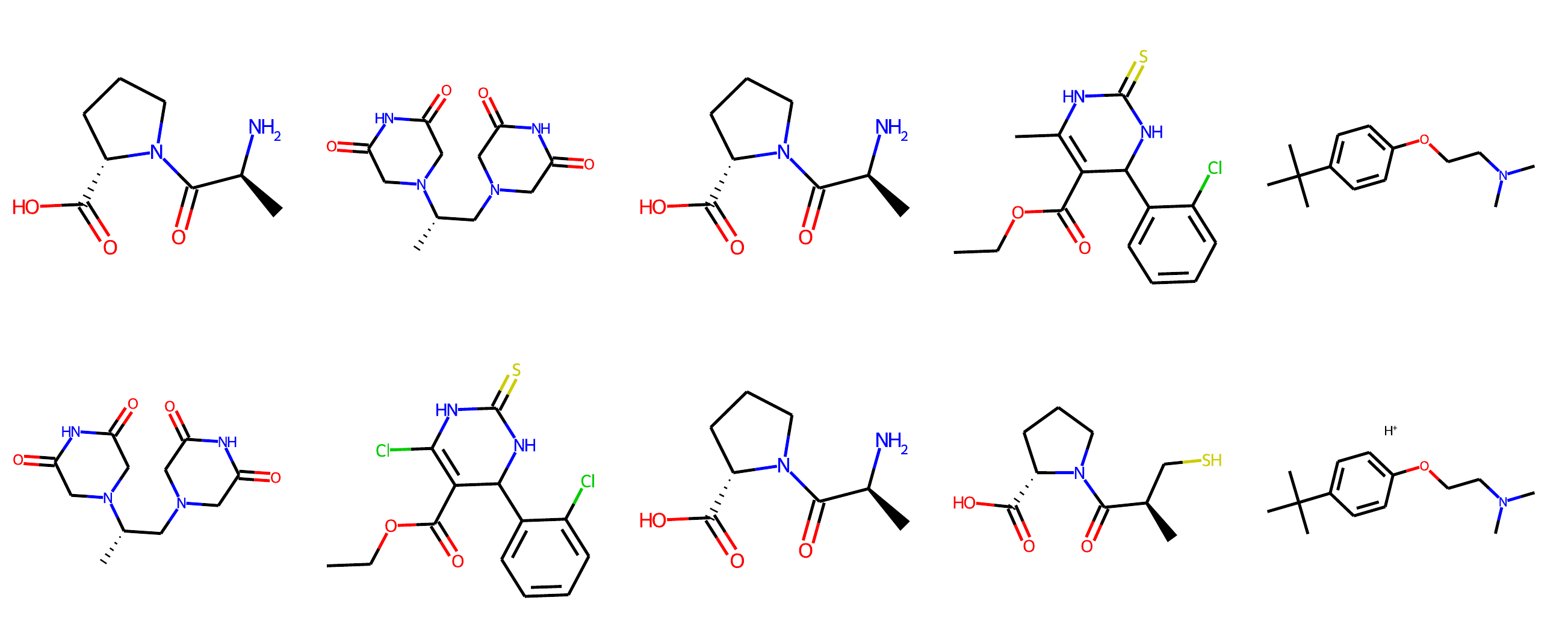}
\caption{Input: The molecule is an anti-ulcer drug.}
\end{figure*}

\begin{figure*}
\centering
\includegraphics[width=\textwidth]{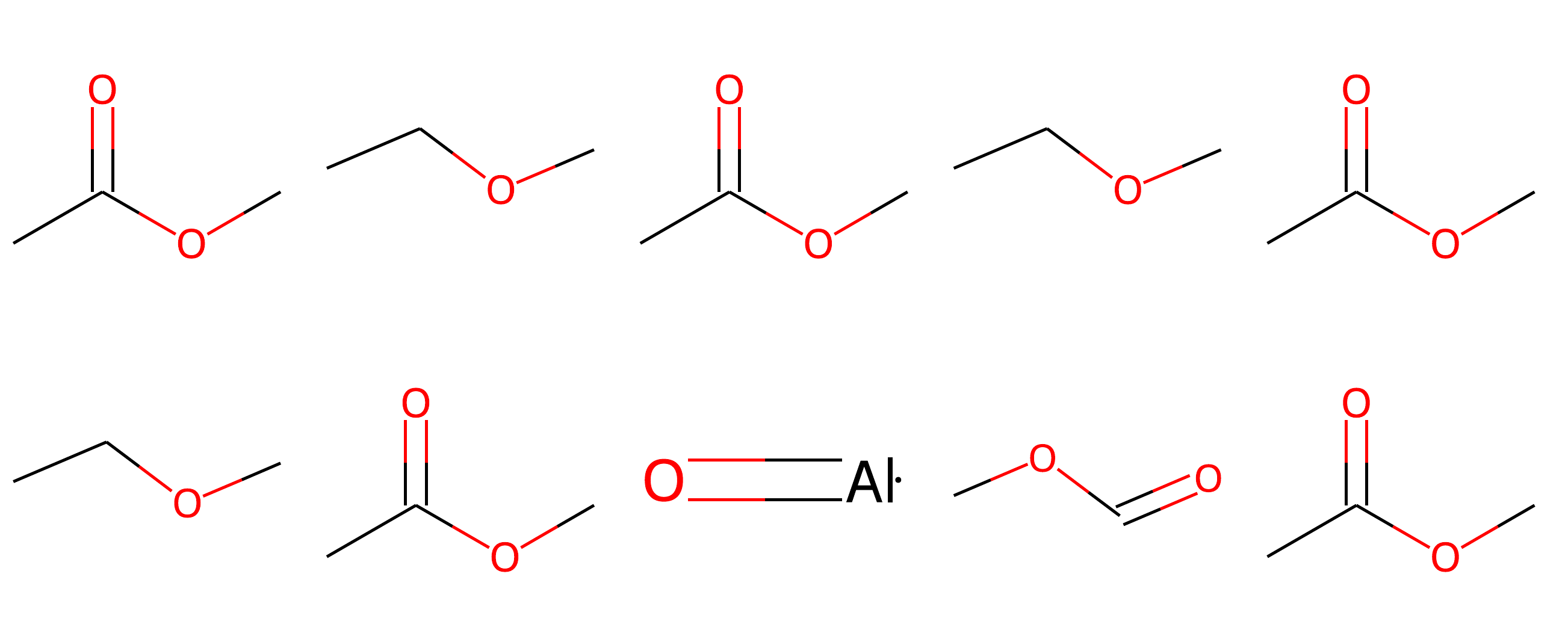}
\caption{Input: The molecule is an aromatic ether.}
\end{figure*}

\begin{figure*}
\centering
\includegraphics[width=\textwidth]{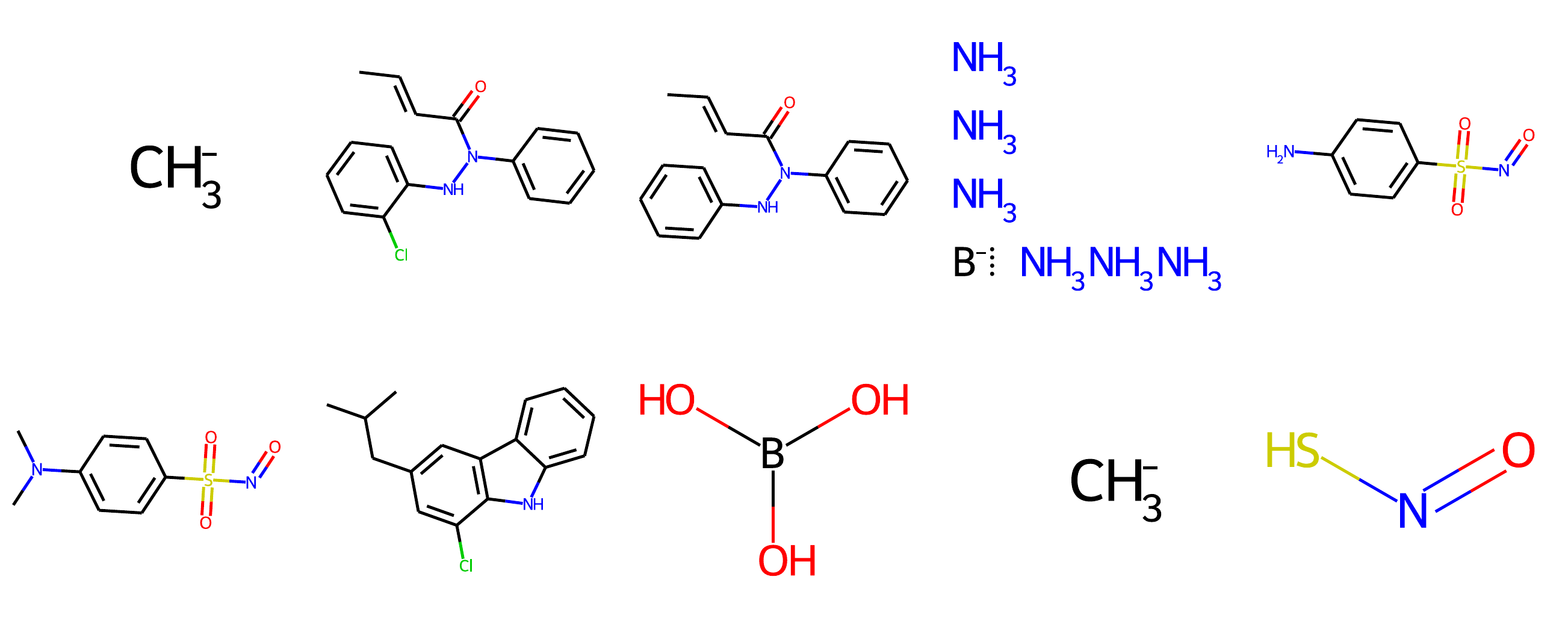}
\caption{Input: The molecule is a catabolic agent.}
\end{figure*}

\begin{figure*}
\centering
\includegraphics[width=\textwidth]{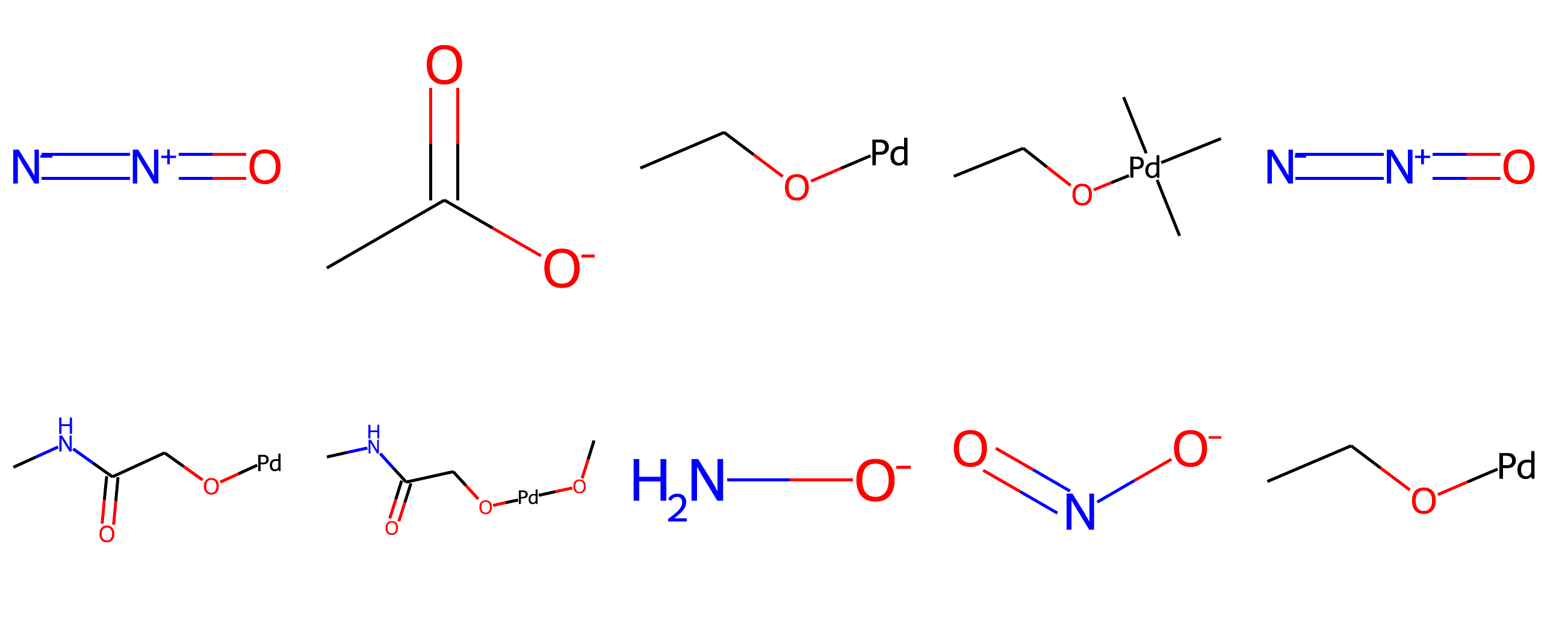}
\caption{Input: The molecule is an explosive.}
\end{figure*}

\begin{figure*}
\centering
\includegraphics[width=\textwidth]{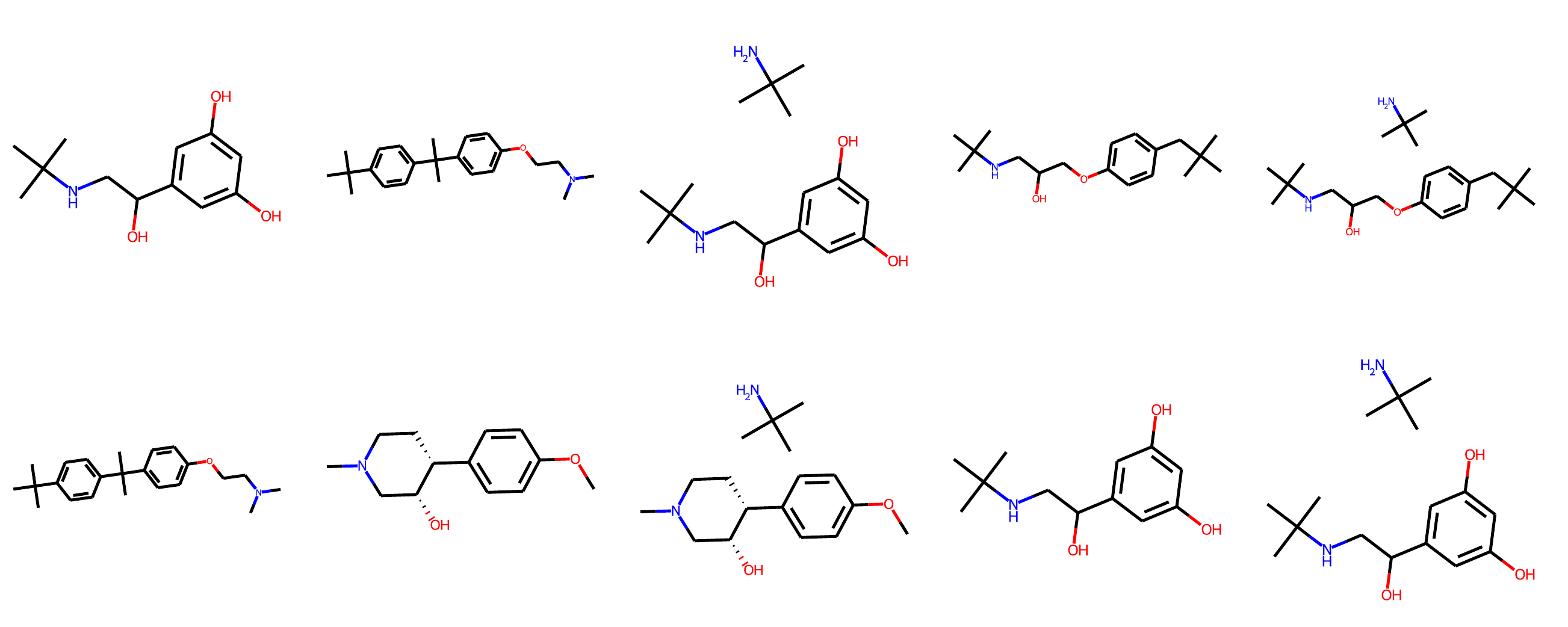}
\caption{Input: The molecule is an inhibitor of the Parkinson's disease.}
\end{figure*}

\begin{figure*}
\centering
\includegraphics[width=\textwidth]{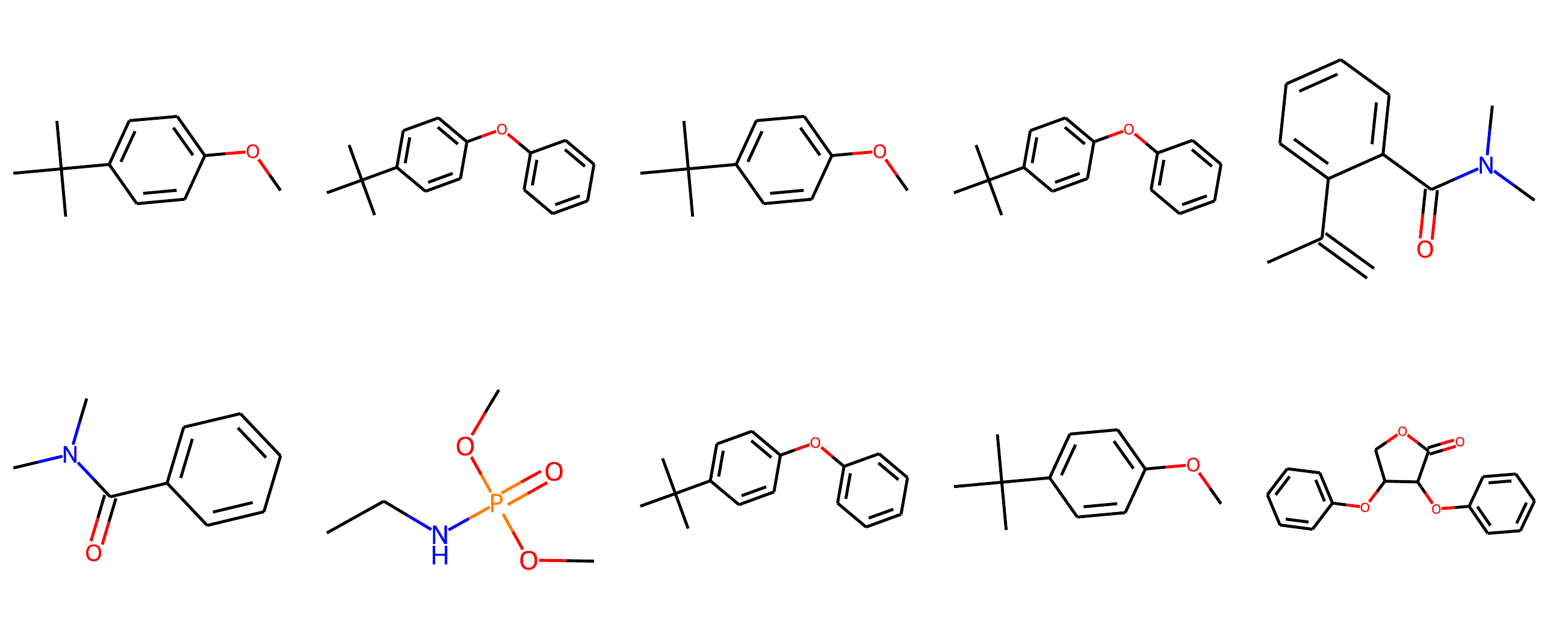}
\caption{Input: The molecule is an insect attractant.}
\end{figure*}

\begin{figure*}
\centering
\includegraphics[width=\textwidth]{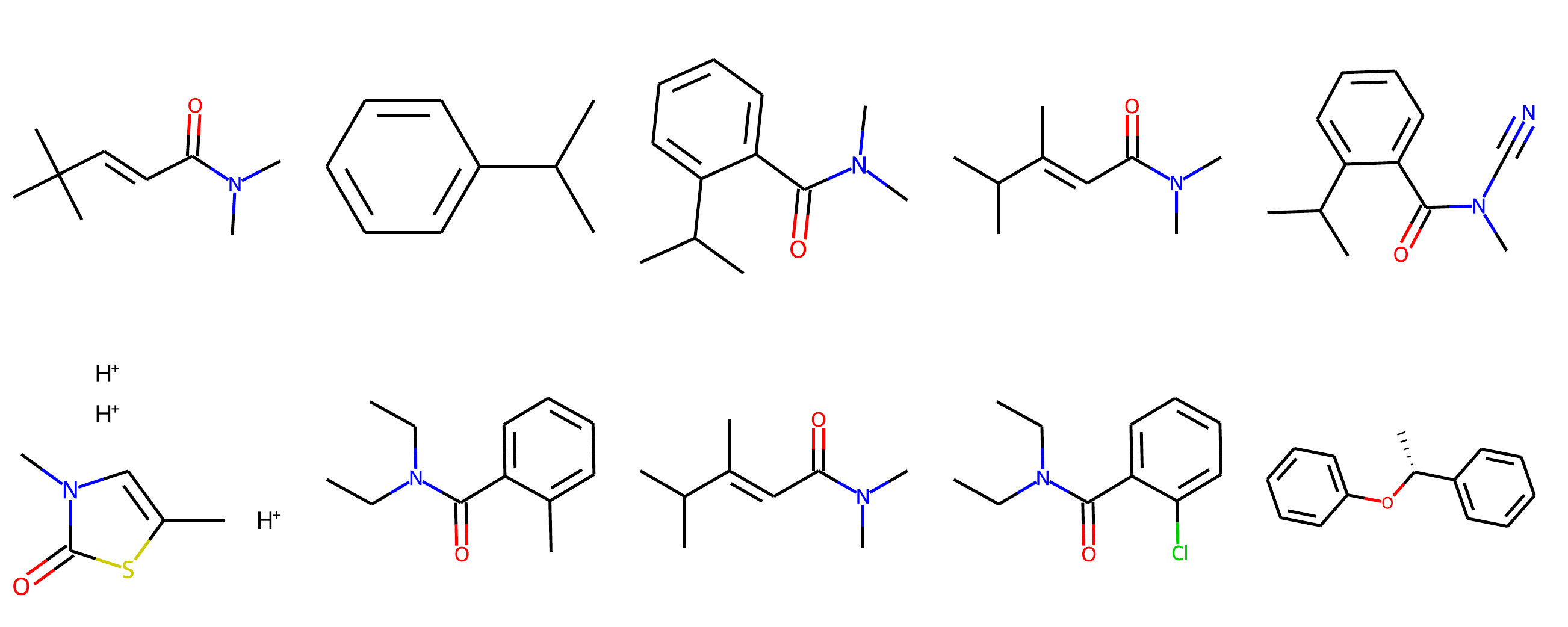}
\caption{Input: The molecule is an insecticide.}
\end{figure*}

\begin{figure*}
\centering
\includegraphics[width=\textwidth]{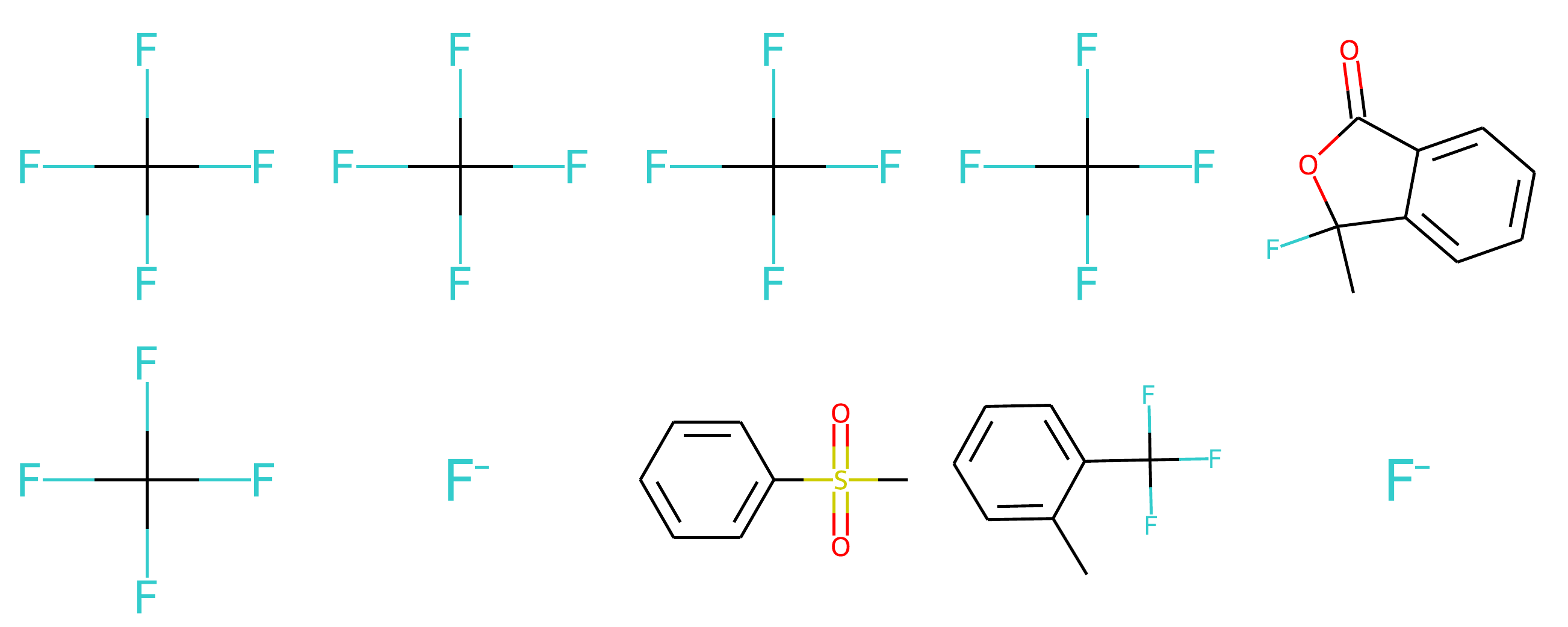}
\caption{Input: The molecule is an organofluorine compound.}
\end{figure*}

\begin{figure*}
\centering
\includegraphics[width=\textwidth]{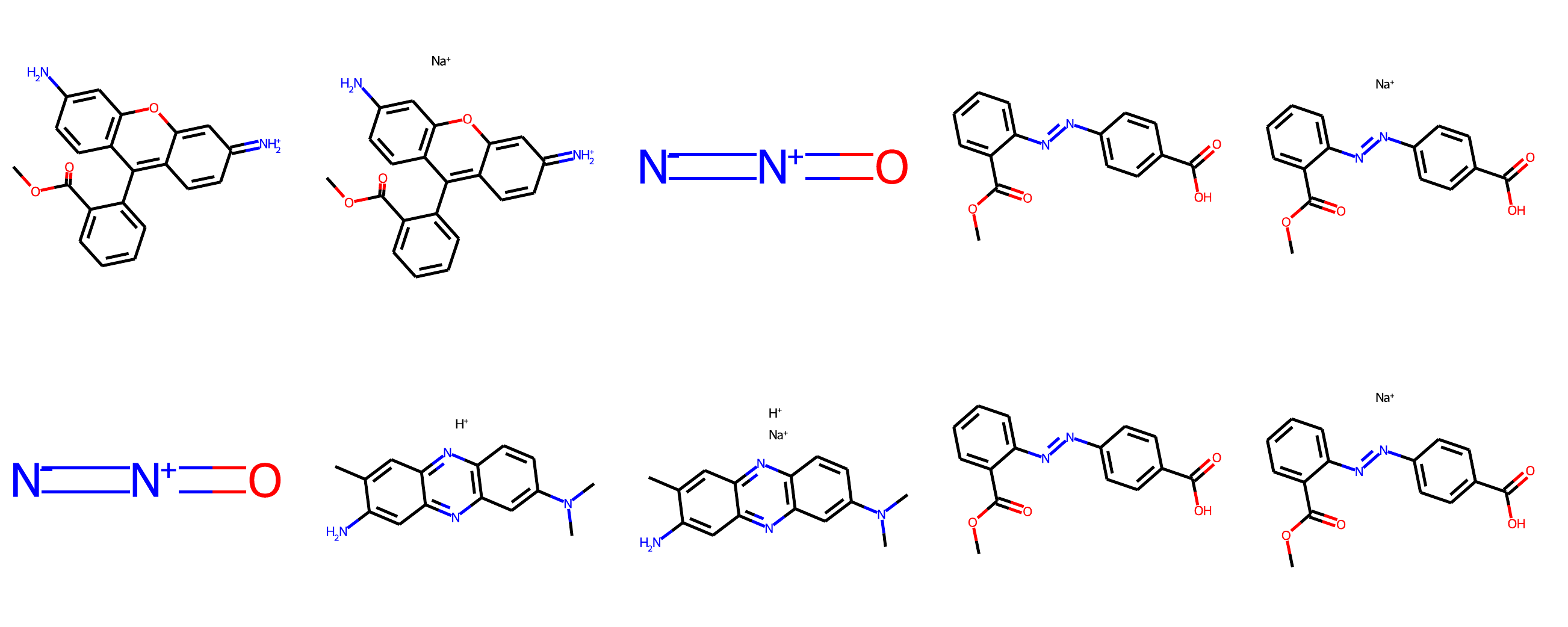}
\caption{Input: The molecule is blue.}
\end{figure*}

\clearpage
\begin{figure*}
\centering
\includegraphics[width=\textwidth]{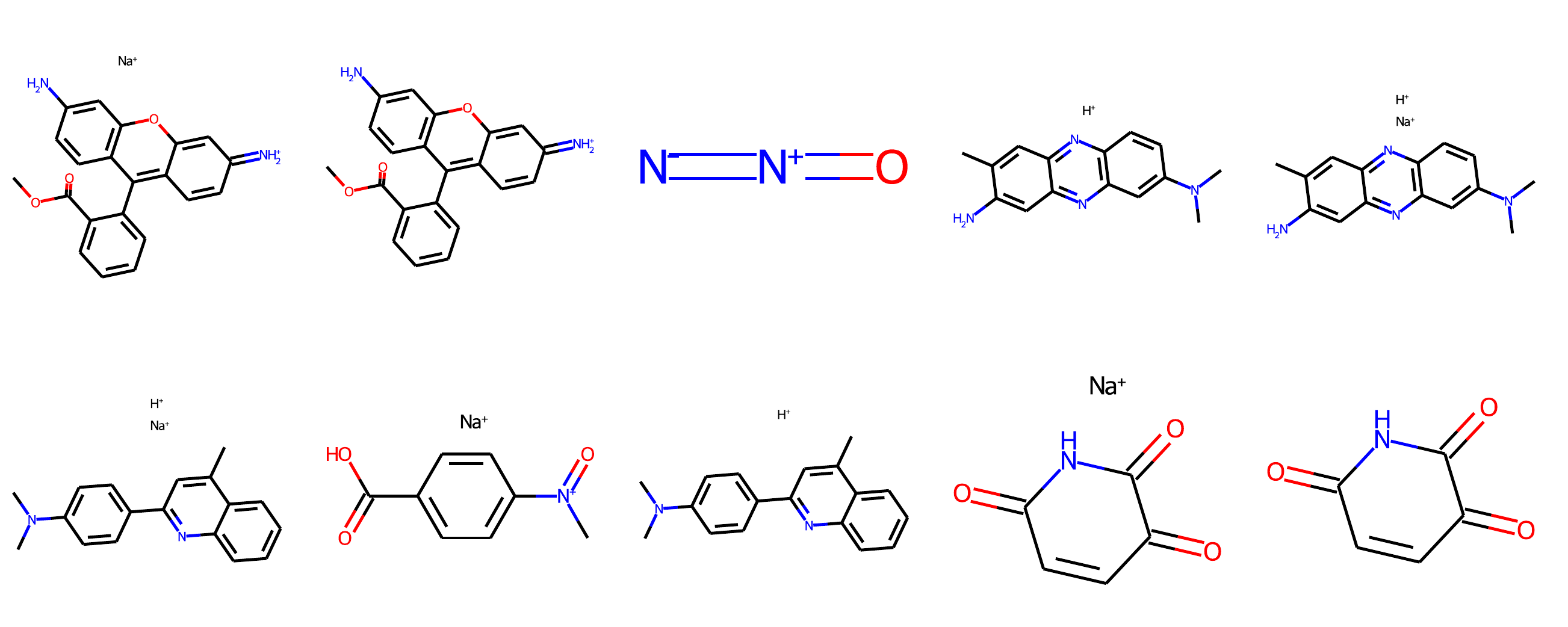}
\caption{Input: The molecule is blue blue.}
\end{figure*}

\begin{figure*}
\centering
\includegraphics[width=\textwidth]{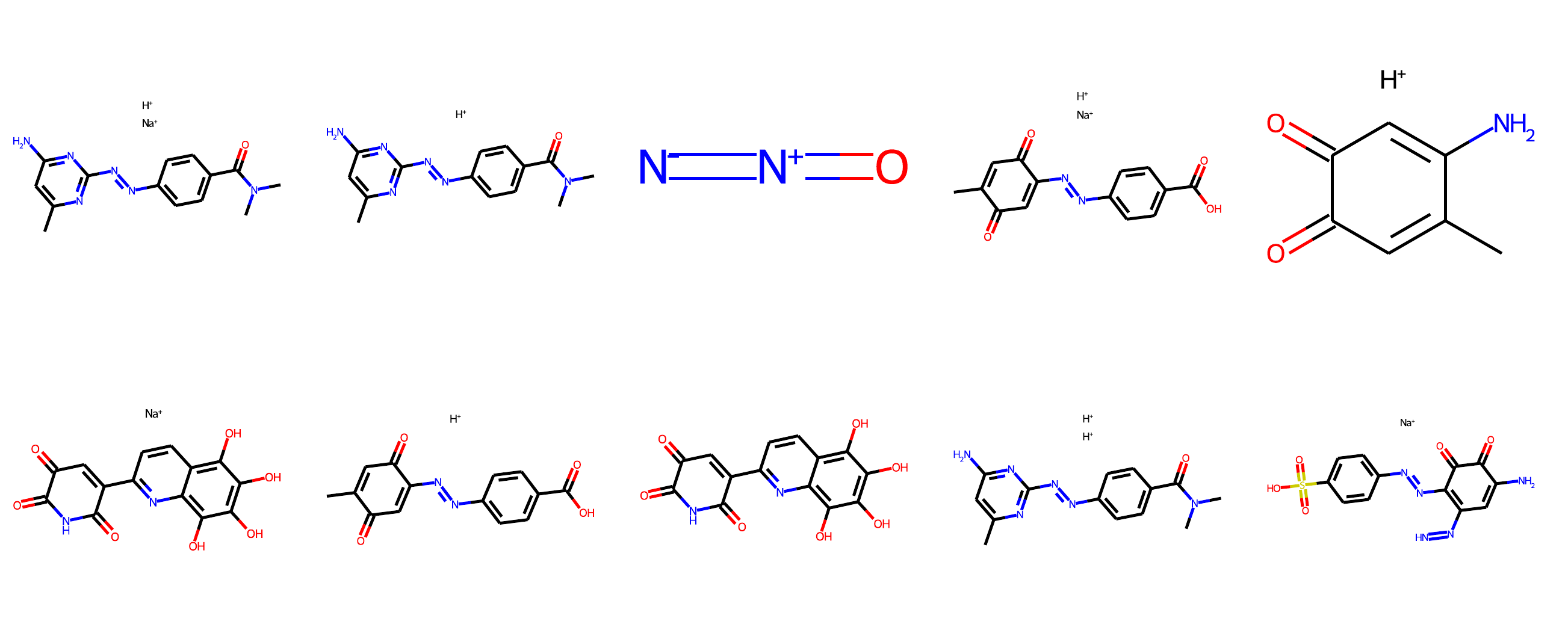}e
\caption{Input: The molecule is blue blue blue.}
\end{figure*}

\begin{figure*}
\centering
\includegraphics[width=\textwidth]{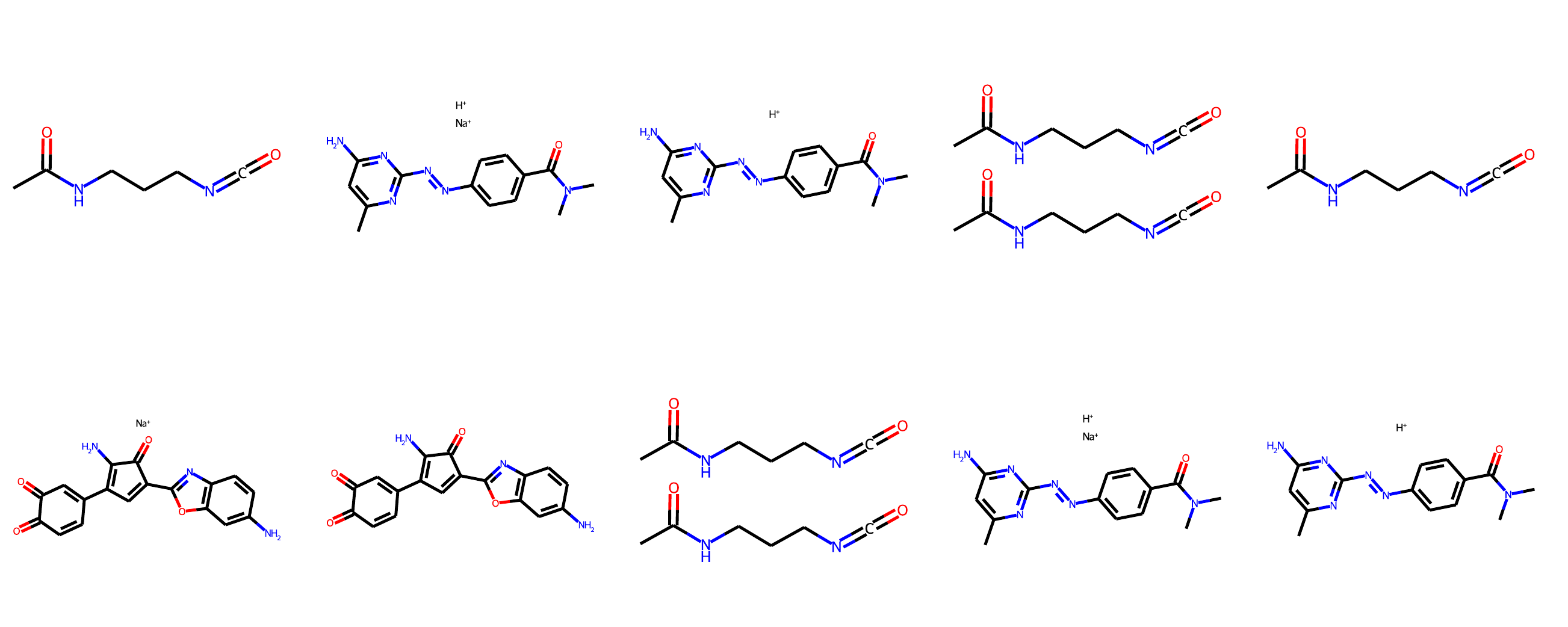}
\caption{Input: The molecule is blue blue blue blue.}
\end{figure*}

\begin{figure*}
\centering
\includegraphics[width=\textwidth]{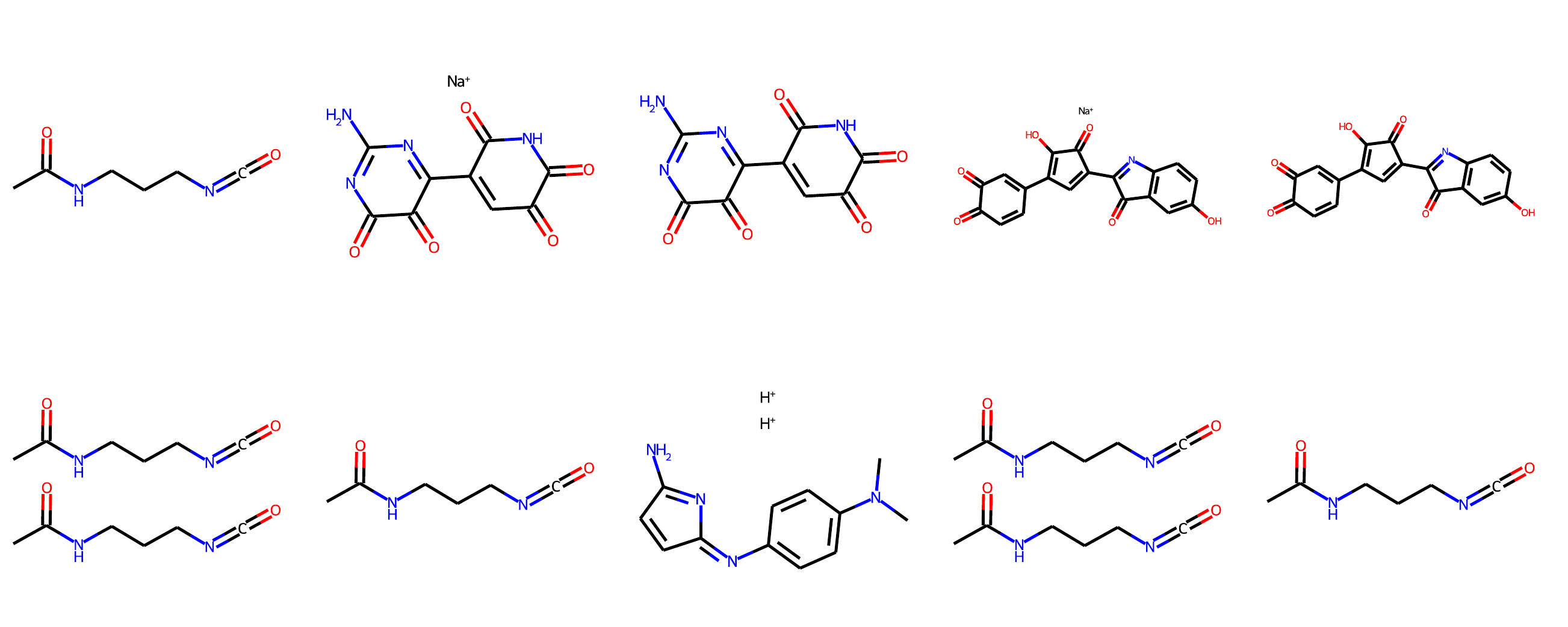}
\caption{Input: The molecule is blue blue blue blue blue.}
\end{figure*}

\begin{figure*}
\centering
\includegraphics[width=\textwidth]{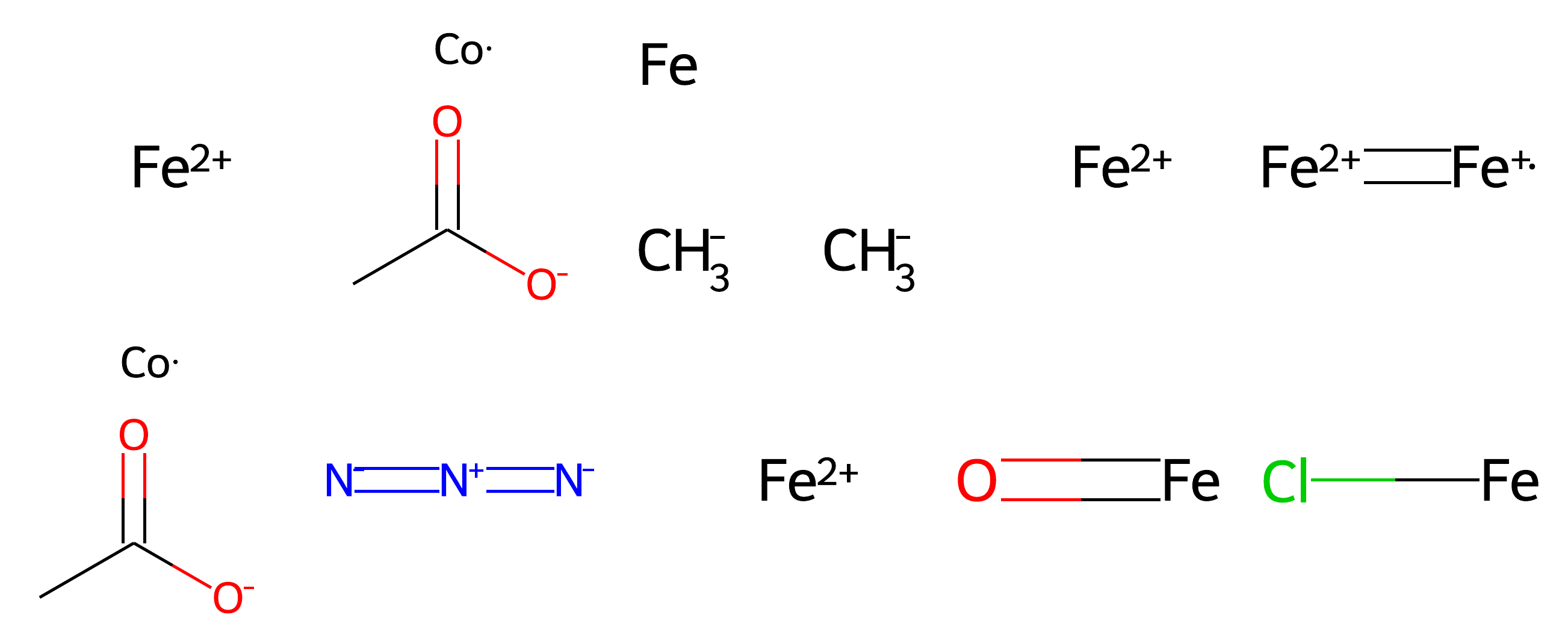}
\caption{Input: The molecule is electrically conductive.}
\end{figure*}

\end{document}